*Review*

# Embodied AI with Foundation Models for Mobile Service Robots: A Systematic Review

**Matthew Lisondra[1*], Beno Benhabib[1], and Goldie Nejat [1,2*]**

[1] Autonomous Systems and Biomechatronics Laboratory (ASBLab), Department of Mechanical and Industrial Engineering, University of Toronto, Toronto, ON M5S 3G8, Canada
[2] KITE, Toronto Rehabilitation Institute43., University Health Network (UHN), Toronto, ON M5G 2A2, Canada
[*] Authors to whom correspondence should be addressed: lisondra@mie.utoronto.ca, nejat@mie.utoronto.ca

**Abstract**

Rapid advancements in foundation models, including Large Language Models, Vision-Language Models, Multimodal Large Language Models, and Vision-Language-Action Models, have opened new avenues for embodied AI in mobile service robotics. By combining foundation models with the principles of embodied AI, where intelligent systems perceive, reason, and act through physical interaction, mobile service robots can achieve more flexible understanding, adaptive behavior, and robust task execution in dynamic real-world environments. Despite this progress, embodied AI for mobile service robots continues to face fundamental challenges related to the translation of natural language instructions into executable robot actions, multimodal perception in human-centered environments, uncertainty estimation for safe decision-making, and computational constraints for real-time onboard deployment. In this paper, we present the first systematic review focused specifically on the integration of foundation models in mobile service robotics. We analyze how recent advances in foundation models address these core challenges through language-conditioned control, multimodal sensor fusion, uncertainty-aware reasoning, and efficient model scaling. We further examine real-world applications in domestic assistance, healthcare, and service automation, highlighting how foundation models enable context-aware, socially responsive, and generalizable robot behaviors. Beyond technical considerations, we discuss ethical, societal, and human-interaction implications associated with deploying foundation model-enabled service robots in human environments. Finally, we outline future research directions emphasizing reliability and lifelong adaptation, privacy-aware and resource-constrained deployment, and governance and human-in-the-loop frameworks required for safe, scalable, and trustworthy mobile service robotics.

**Keywords:** Large Language Models, Mobile Service Robots, Foundation Models, AI-Enabled Robotics

## 1. Introduction

In recent years, there have been rapid advancements in foundation models including Large Language Models (LLMs) [1]-[6], Vision Language Models (VLMs) [7]-[9], Multimodal Large Language Models (MLLMs) [10], [11] and Vision Language Action Models (VLAs) [12]-[14]. These models are significantly changing the overall field of robotics and its numerous applications. In particular, there have been notable improvements to personalized task management and context-aware smart assistant capabilities for domestic assistance [15]-[17], healthcare [18]-[21], and service automation [22], [23]. Foundation models are trained on vast datasets to understand and generate human-like text and speech, as well as interpret visual information and reason. For example, language-guided mobile service robots utilizing foundation models can carry-out complex instructions [24]-[26], perform fetch and carry tasks [27]-[30] and/or socially interact [31]-[33] with human users.

As robots are deployed in real-world environments to increasingly interact with humans, the integration of language-based intelligence has become crucial for intuitive and effective communications. In particular, mobile service robots can provide adaptable, context-aware assistance in dynamic and unpredictable environments to achieve various goals, from navigating person-centered environments to object or person detection and identification. Their ability to autonomously interpret human instructions, reason about tasks, and execute actions has enabled them to be deployed in numerous environments, from warehouses [34], [35] to hospitals [12], [36] and personal homes [26], [37], [38].

By leveraging the advancements in foundation models, service robots can bridge the gap between AI-driven perception and human-like decision-making, leading to natural human-robot interactions (HRI) and collaborations in everyday scenarios. For mobile service robots, success also depends on dexterous, compliant physical interactions with people and objects, adherence to HRI constraints (e.g., personal space, idiosyncrasies, behavior predictability), and long-term autonomy in human-centered environments. However, to fully integrate the promising capabilities of foundation models, the open challenges in the development and deployment of embodied AI mobile service robots must be addressed. These open challenges include:

*1) Translation of Natural Language Instructions into Executable Robot Actions*: mobile service robots need to handle ambiguous, incomplete, or colloquial instructions ("*bring me that from the other room*") while grounding these instructions in physical navigation and manipulation tasks [39]-[43]. This can be challenging in domestic and healthcare settings, where instructions may be highly contextualized by the environment or a user's state (e.g., a patient who is resting, in pain, or under medication may give shortened or unclear commands that require contextual interpretation by the robot). They also require long-horizon, context-aware planning across multiple rooms, dynamic layouts, and human activities [44]-[46].

*2) Multi-modal Perception*: mobile service robots must integrate multimodal inputs such as vision, speech, and tactile, while operating in human-centered environments such as homes and hospitals. As they navigate from region to region, they need to be able to adapt to: 1) lighting variations, 2) crowds, 3) occlusions by dynamic people, furniture, and/or medical equipment, and 4) noisy environments with background conversations or sounds including alarms [47]-[51].

*3) Uncertainty Estimation*: as mobile service robots operate in safety-critical, human-facing environments, decisions must be made under partial observability, unpredictable human behaviors, and socially acceptable norms [52]-[54]. Namely, they need to estimate both aleatoric (sensor/environmental) and epistemic (model) uncertainty while ensuring their actions remain socially acceptable. However, current embodied AI methods struggle with uncertainty estimation, leading to overconfident predictions that can risk user safety. For example, a hospital delivery robot might proceed through a crowded hallway despite incomplete LiDAR or vision data, misjudging the proximity of patients or equipment and causing potential collisions [55]-[58].

*4) Computational Capabilities*: Mobile service robots are constrained by onboard hardware, as continuous reliance on remote servers or cloud computation is often infeasible due to latency, connectivity instability, and data privacy concerns in hospitals and homes [59], [60]. Sensing, perception, and decision-making need to be executed locally on embedded GPUs, NPUs, or other onboard accelerators close to the data source. The challenge lies within the tight compute, energy, and latency budgets of embedded platforms while ensuring safe operation around humans.

In general, foundation models can address the above challenges. They can bridge the gap between high-level human instruction, language understanding and low-level robotic control to enable more intelligent, adaptable, and interactive robots for diverse real-world applications, as shown in Figure 1.

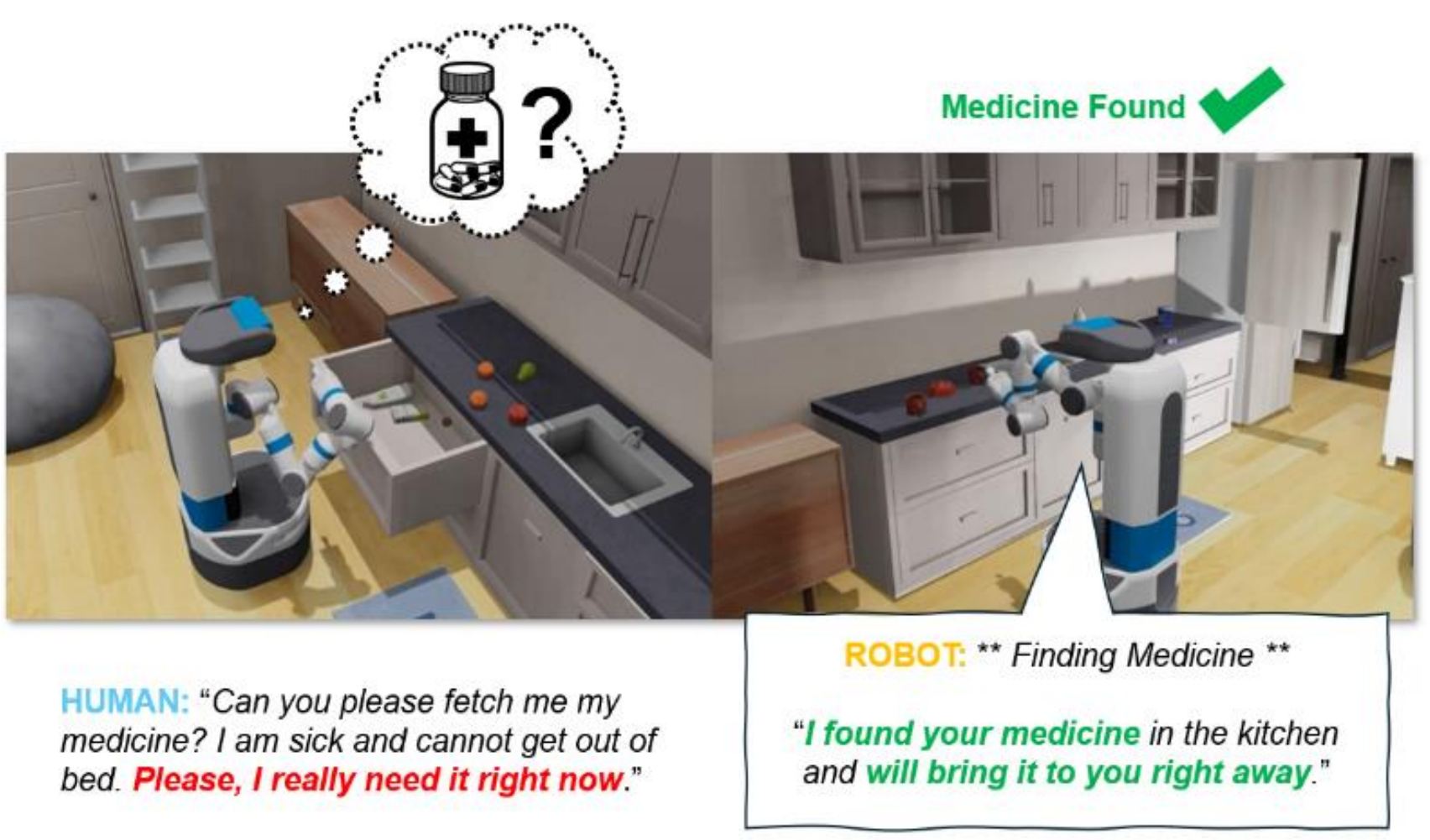

**Figure 1.** A VLM-MLLM pipeline enables language-guided medicine retrieval in a home. Images obtained from the Habitat 1.0 Simulator [61].

To-date, existing surveys have primarily examined either broad, non-task-specific applications in general-purpose robotics [62], [63] or language-conditioned manipulation tasks for stationary robotic arms [64]. These have not yet investigated the role of mobility in enabling robots to assist with human-centered tasks. In this paper, we present the first systematic review and analysis on the integration of foundation models in mobile service robotics, examining their role in advancing embodied AI. We introduce a unified architecture view in Figure 2 that illustrates how foundation models can integrate perception, planning, and control within mobile robots operating in human environments. Building on this integrated perspective, we identify research challenges, and discuss how foundation models can enable task generalization, dynamic scene understanding, and social compliance. In particular, we focus on the promising applications of domestic assistance, healthcare and service automation for mobile service robots.

## 2. Open Challenges of Embodied AI for Mobile Service Robots

In this section, we identify four challenges that need to be addressed to leverage the potential of foundation models in service robotics.

To quantify how research effort in mobile service robotics has been distributed to-date, we conducted a publication analysis using OpenAlex [65], a large-scale scientific indexing platform. We queried and filtered works related to mobile service robots integrating language, perception, planning, and control, and grouped the resulting literature by recurring technical themes that consistently emerge across decades of research. A total of 7,506 relevant works published between 1968 and 2025 were retrieved after excluding works related to autonomous driving and aerial robotics.

Table 1 presents a ranking of the challenges based on the number of works associated with each open challenge. Language-to-Action Mapping ranked the highest (29.18%), representing an emphasis on grounding instructions into executable behavior through planning, navigation, and manipulation. Multimodal Perception was the second-highest ranked theme (28.83%), highlighting the challenge of integrating heterogeneous sensory inputs for varying environments Uncertainty Estimation was ranked third (26.71%) due to the concerns related to robustness, safety, and decision-making under ambiguity in human-centered settings. Computational Capabilities, while ranked fourth (15.28%), still remains a limitation to real-time execution and embedded deployment. Below we discuss each of the four challenges in more details, while defining specific subchallenges for each.

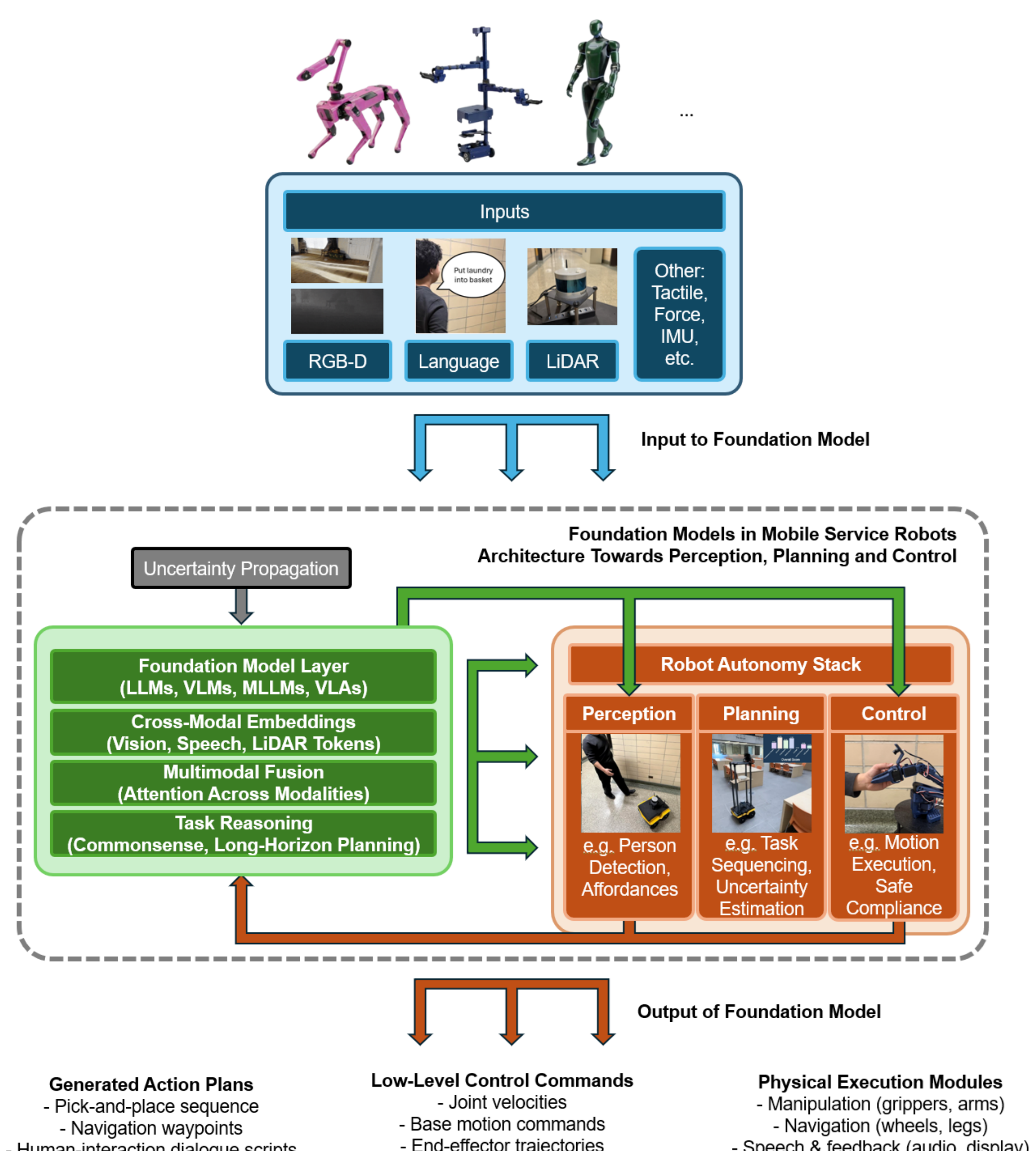

**Figure 2.** Foundation models fuse multimodal inputs to drive perception, planning, and control. Images for the top row robots were generated using the Nano Banana Pro AI Image Generator. RGB-D pair of images obtained from the Habitat 1.0 Simulator [61]. All other images are from the authors.

**TABLE 1**
RANKING OF THE FOUR CORE CHALLENGES IN MOBILE SERVICE ROBOTICS

| Rank | Challenge | Paper Count N | Percent of Papers (N / 7,506 × 100%) | Conclusions |
|---|---|---|---|---|
| #1 | Challenge #1 Language-to-Action Mapping | 2,190 | 29% | Largest research focus: semantic grounding, task sequencing, navigation & manipulation logic |
| #2 | Challenge #2 Multimodal Perception | 2,164 | 29% | Strong support for cross-modal sensing and perception alignment |
| #3 | Challenge #3 Uncertainty Estimation | 2,005 | 27% | Strong emphasis on safety, robustness, and decision-making under uncertainty |
| #4 | Challenge #4 Computational Capabilities | 1,147 | 15% | Focus on real-time execution, embedded efficiency, and resource constraints |

*2.1. Challenge #1: Translation of Natural Language Instructions into Executable Robot Actions*

Enabling language-guided mobile service robots to follow high-level natural language commands provided by humans remains a challenge. Key limitations include:

*1) Symbolic-to-Embodied Mapping and Instruction Ambiguity*: Natural language instructions are typically provided by non-expert human users. These commands are often expressed at a high level ("*clean the room*," "*get my walking cane*") and may include colloquial, ambiguous, or underspecified phrasing, lacking precise object identifiers, spatial context, or temporal constraints. While classical planners such as STRIPS [39] and SHOP2 [40], [41], or logic-based systems (e.g., PDDL) [42], [43], attempt to decompose commands into discrete subgoals, they rely on hand-engineered operators that are domain-specific and brittle [66]-[69]. This challenge is particularly critical for mobile service robots engaging in HRI with people in different roles who may provide incomplete requests ("*bring me that*") or might use gestures and speech simultaneously, requiring robots to adapt ambiguous instructions into executable actions.

*2) Lack of Embodied Commonsense and Physical Constraints Awareness*: Mobile service robots must not only interpret commands but also reason about constraints in human-centered environments. Many symbolic planners, grounding models, and PDDL-based models lack awareness of robot embodiment and environmental variability, [70], [71]. While language-only systems or semantic parsers trained on demonstrations can capture basic kinematics or affordances, however, they are not able to generalize across diverse service contexts [44]-[46]. For example, a patient may ask a robot to "*place the food tray on the bed*," but without understanding that the bed is already occupied, the robot could attempt an unsafe or socially inappropriate action. Similarly, commands like "*open the door for my grandson*" may be physically infeasible if the robot lacks the dexterity to open door. In these cases, failing to account for physical constraints can result not only in task failure but also in loss of user trust.

*3) Failure in Long-Horizon Task Planning*: Many mobile service tasks, such as medication delivery, multi-room navigation, or guided patient assistance unfold over extended time horizons, requiring robots to reason across multiple steps, maintain working memory, and adapt to evolving conditions [72]. Classical planners such as Simple Task Network (STN) Planning [73] or Fast Forward (FF) [74] perform well in static domains, but they are not able to handle dynamic, human-centered environments where unexpected subgoals may emerge due to users' needs or environmental changes [75], [76]. Learning-based models such as Neural Task Graphs (NTG) [77] and Option-Critic [78] offer flexibility in sequencing actions, yet they can suffer from policy drift, compounding errors, and limited memory retention when deployed in real-world service contexts [79]-[81]. As a result, mobile service robots are unable to maintain task continuity, recover from errors, or adapt their plans in response to user interruptions or environmental changes, leading to cascading task failures during long-horizon service execution.

*2.2. Challenge #2: Multimodal Perception*

Multimodal perception is essential for mobile service robots to interpret complex environments. These modalities often differ in spatial resolution, sampling frequency, and noise characteristics, making real-time sensor fusion and alignment difficult in human-centered environments. We discuss four core limitations below.

*1) Cross-Modal Representation:* Sensory data from various robotic sensors need to be fused into a shared latent space to support real-time reasoning in mobile service robots. Fusion architectures are commonly classified into [82]: (*i*) early fusion, where raw inputs are concatenated or merged at the pixel level, (*ii*) late fusion, which aggregates decisions from different modalities, and/or (*iii*) intermediate fusion, where features from each modality are encoded and combined using attention mechanisms, joint embeddings, or cross-modal transformers. However, for mobile service robots, each fusion type presents unique challenges. Early fusion is sensitive to cluttered environments, background conversations and variations in lighting [83] since it combines raw, low-level sensory inputs (e.g., pixel intensities or waveform amplitudes) before feature extraction. This means that any noise or occlusion in one modality, such as poor lighting in images or overlapping human speech in audio, can potentially corrupt the entire fused representation, leading to degraded perception accuracy and unstable behavior in mobile service robots. Late fusion can present delayed or inconsistent results in the fused perception output and downstream decision-making when different modalities provide conflicting cues, such as audio indicating a person's presence while vision temporarily loses track of the person due to occlusions in crowded human spaces, leading to uncertainty for situational awareness [84], [85]. Intermediate fusion, which combines encoded feature representations from each modality using attention or joint embeddings, often leads to temporal gaps when aligning robot data streams with different sampling rates, resolutions, and noise characteristics. This can cause failures, for example, in gesture-speech association during HRI [48], [49].

The aforementioned challenges are compounded by spatial misalignment and temporal desynchronization when robots: (*i)* follow or guide users across different rooms, or (*ii)* interact with multiple people simultaneously. When a caregiver both speaks ("*pass me that medication*") and simultaneously points toward a medicine cart, the robot must associate the speech command with the pointing gesture. If the audio (speech) and visual (gesture) signals are not synchronized, the robot may misinterpret which object the caregiver is referring to, resulting in ineffective assistance. In practice, fusion inconsistencies can cause object association errors, scene parsing failures, and HRI breakdowns. Such failures can directly undermine fine-grained context interpretation, and a service robot's ability to disambiguate human intent and environmental cues to provide task assistance.

*2) Latency Issues*: Mobile service robots often rely on heterogeneous sensor streams that operate at different sampling rates, leading to temporal desynchronization [86]-[88]. These robots need to maintain situational awareness in dynamic, human-centered environments where people can have unpredictable motion and tasks require real-time social responsiveness. Unlike fusion errors, latency stems from delays in updating perception and control loops. For example, a domestic assistant robot may take a unnecessarily long detour as obstacle information were updated too late during navigation [86]. These delays reduce interaction responsiveness and fluidity, making robots appear slow or hesitant in environments where people expect robots to adapt in real-time.

*3) Uncertainty Propagation Across Modalities*: Sensor quality may degrade due to lighting changes, occlusions due to people, cluttered rooms, and environmental noise [89]-[91]. Uncertainty propagation occurs when sensor noise or degradation in one modality affects the reliability of fused estimates across multiple modalities. For example, in joint estimation pipelines, this problem is magnified for service robots as it can cause cascading perception errors that directly affect safe and effective HRI. Yet, many fusion methods assume stationary sensor noise models and fixed modality trust, which can result in overconfident predictions or action outputs, [92], [93]. Overconfidence can lead to unsafe behaviors for mobile service robots. For example, a robot might continue to navigate confidently through a dim hallway, despite losing its visual information. Adaptive confidence weighting, which assigns modality-specific weights based on online reliability metrics [94], [95], remains underdeveloped for real-world deployment of mobile service robots due to such key limitations as: (*i*) high computational overhead incompatible with embedded edge hardware, (*ii*) assuming static environments, and (*iii*) lacking robustness to abrupt sensor failures and scene changes.

*4) Domain Adaptation and Transferability Issues*: Early, late, and intermediate fusion models are often trained in static, well-lit indoor datasets and degrade in deployment across the cluttered households, dim hospital corridors, or dynamic outdoor and public venues which mobile service robots work in [96], [97]. This domain gap makes it difficult for models to generalize once deployed in cluttered dynamic environments with multiple surface types (hospitals with glass doors and walls). Domain adaptation methods such as domain randomization [98], self-supervised online learning [99], and adversarial style transfer [100] face limitations in mobile service robot deployments. Transferability is further hindered when models are overfit to specific training conditions. Mobile service robots cannot retrain models in real-time during operation, making them vulnerable to perceptual drift. Thus, robots can lose accuracy when exposed to novel environment set-ups; different layouts in long-term care homes, outdoor seasonal or weather-related lighting variations, or different human movement patterns.

*2.4. Challenge #3: Uncertainty Estimation*

Uncertainty is inherent in language-guided mobile service robotics, arising from sensor noise, human ambiguity, partial observability, and/or dynamic environments [55]-[58], [101], [102]. Uncertainty estimation in mobile service robots has four core limitations:

*1) Lack of Explicit Uncertainty Quantification*: Mobile service robots often lack the ability to estimate confidence in their decisions. While probabilistic methods consider modeling confidence [55], [101], they are computationally intensive for real-time robotic applications. As a result, many service-oriented models rely on rule-based pipelines [56] or standard deep learning architectures [57], [58] that do not provide calibrated measures of uncertainty [103]. This limitation can cause implementation failures.

*2) Failure in Long-Horizon Uncertainty Estimation*: Reasoning about how uncertainty accumulates over extended time horizons is critical for service tasks. Traditional state estimation and planning methods, such as Kalman Filters [104], [105], belief-space planners and sampling-based approaches [106], do not model how uncertainty accumulates [107]-[109], which may lead to underestimating uncertainty as prediction horizons increase [110], [111]. Belief-space motion planners [112] and sampling-based approaches such as Partially Observable Sparse Samplers [106] can become computationally intractable as the planning horizon increases [79]-[81]. For mobile service robots, this limitation directly impacts their ability to anticipate cascading errors, recover from uncertain states, or adapt task plans over time. In practice, robots may overcommit to long navigation paths that are not valid in dynamic or crowded environments.

*3) Uncertainty in HRI*: A central challenge for service robots is managing uncertainty in user-facing decisions, where ambiguous verbal commands, nonverbal gestures, or multimodal cues must be interpreted in real-time. To date, the majority of robots engaged in HRI lack the ability to communicate uncertainty or proactively request clarifications, which often results in social errors and not adhering to social etiquette rules [113]-[116]. These failures typically stem from misaligned intent interpretation, where the robot's confidence is overestimated despite incomplete or noisy human input. Traditional approaches including, plan-based dialog management frameworks [52], behavior-tree-based HRI controllers [53], and Partially Observable Markov Decision

Process (POMDP)-based dialog managers [54] often assume fixed interaction protocols and lack the ability to model, quantify, or express uncertainty in real-time, [117]. Recent studies stress that effective HRI requires not only technical uncertainty quantification but also the ability to express uncertainty in socially meaningful ways, for example, pausing, seeking clarification, or adapting conservatively when unsure [118], [119].

*2.3. Challenge #4: Computational Capabilities*

Mobile service robots have limited onboard computational power, making real-time inference with large-scale symbolic and deep learning models challenging. Key limitations include:

*1) Perception and Planning Computational Overhead*: Large-scale deep learning models often require significant memory [120], [121], and computational resources [122]. For example, models with convolutional backbones [120] or 3D segmentation networks [121] can exceed the real-time processing capacity of edge computing units commonly used in mobile service robots. Classical planning frameworks [123], and the Dynamic Window Approach (DWA) planner [124] were designed for lightweight CPU-based inference with simplified 2D cost maps. While efficient, they are not able to scale to the rich semantic representations needed in human-centered environments, such as multimodal feature maps or RGB-D affordance graphs [125], [126]. Addressing this overhead requires combining efficient model compression and on-chip physical AI with selective edge-cloud collaboration, while maintaining the responsiveness necessary for safe, trustworthy HRI.

*2) Lack of Adaptive Resource Allocation for Real-Time Inference*: Most classical mobile service robot frameworks such as traditional navigation stacks [127] and behavior-based control architectures [128] process sensor data at fixed input resolutions and allocate static CPU or GPU budgets, regardless of changing scene complexity or service robot task urgency [129], [130]. This can lead to onboard resources not being used effectively during routine tasks, however, risks overloading resources when service robots face applications that are interaction-heavy or in cluttered environments [131], [132]. Adaptive techniques have been explored in visual pipelines [133], [134] and mapping frameworks such as ElasticFusion [135] to dynamically adjust computation based on current sensory load or scene complexity. However, their approach is heuristic and they do not generalize across the full perception-planning-control stack required for continuous operation in dynamic human-centered environments. They typically optimize a single subsystem (e.g., perception), without coordinating resource allocation across downstream planning and control modules that must respond to the same real-time constraints [136]-[138]. This results not only in technical inefficiency but also a reduction in service quality. For example, in cluttered domestic environments, increased perception load from dense visual scenes can monopolize computational resources, leaving insufficient capacity for robot motion planning or safety monitoring, leading to hesitations or unsafe navigation behaviors.

Figure 3 situates these four challenges within the broader evolution of embodied AI for mobile service robots. The left portion of the timeline ("*Pre-Foundation Model Era*") highlights how classical pipelines, ranging from STRIPS [39] and PDDL [42], [43] planners to DWA [124], LSD-SLAM [91], and Move Base [127], addressed multimodal perception, language-to-action translation, uncertainty estimation, and computational constraints with task-specific and brittle solutions. The middle segment ("*Rise of Foundation Models*") represents the emergence of large pre-trained models such as GPT [139], BERT [140], CLIP [141], and early robotics integrations (e.g., DeiT [142], SayCan [12], OpenVINS [87]), where scaling laws and transfer learning began to influence service robotics. The right segment ("*Foundation Model Era*") shows recent LLMs, VLMs, MLLMs, and VLAs (e.g., GPT-4 [143], DeepSeek-R1 [144], Magma [145], OpenVLA [146], Aether [147], $\pi_0$ [148], [149]) that directly target the four challenges, as indicated by the color-coded arrows for Challenges #1-#4. This progression motivates Section 3, where we discuss and analyze how foundation models systematically address the challenges and their limitations in more detail.

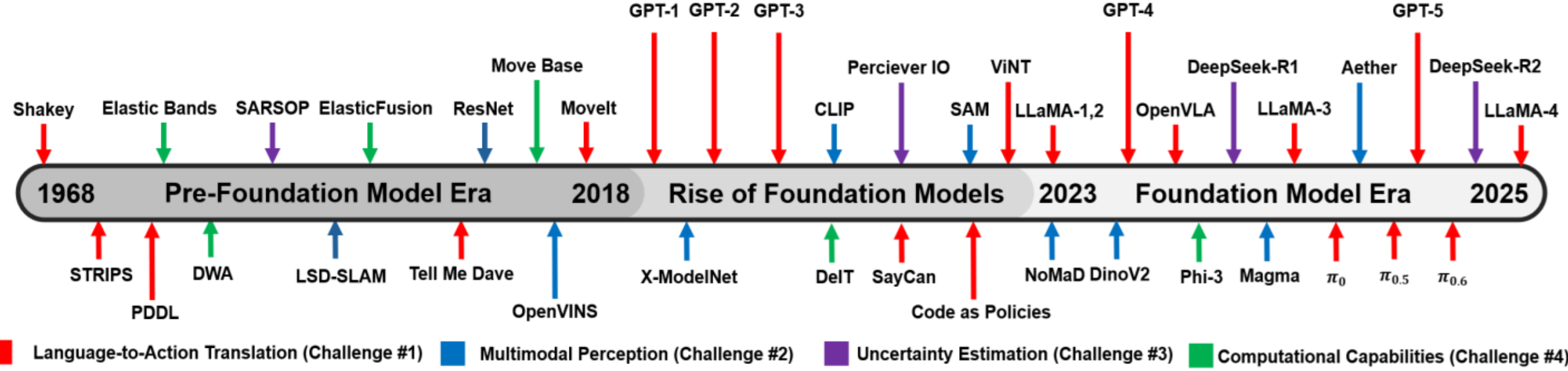

**Figure 3.** Timeline of embodied AI advancements for mobile service robots before and after the rise of foundation models.

## 3. Opportunities in Mobile Service Robots for Foundation Models

This section outlines how foundation models can address the four core challenges identified in Section 2.

To contextualize how foundation models have contributed to embodied intelligence, Figure 4 presents existing mobile service robot capabilities into three functional domains: 1) language communication, 2) vision-language navigation, and 3) manipulation and organization. Language communication capabilities (green box) enable robots to interact with users through language grounding and interactive question answering, supporting clarification, intent inference, and context-aware dialogue. Vision-language navigation capabilities (red box) support multimodal spatial reasoning for both autonomous visual goal reaching and language-guided navigation in complex environments. Manipulation and rearrangement capabilities (orange box) capture object-centric physical interaction, including object and scene manipulation (e.g., grasping, tool use, placing) as well as spatial rearrangement of objects within an environment (e.g., tidying, organizing, or reconfiguring layouts according to task goals). These capabilities go beyond isolated grasps to include sequencing, spatial reasoning, and task-level organization of objects, which are essential for everyday service tasks such as fetching, cleaning, and household reorganization. Together, these three domains provide a practical behavioral decomposition of how foundation models enable mobile service robots to perform meaningful, real-world assistance.

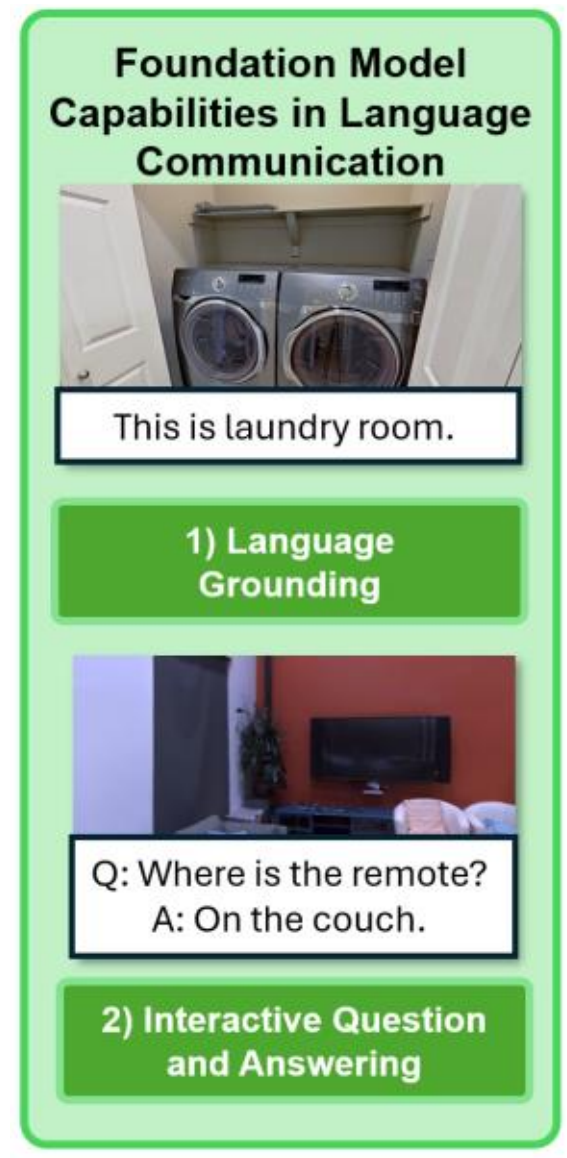

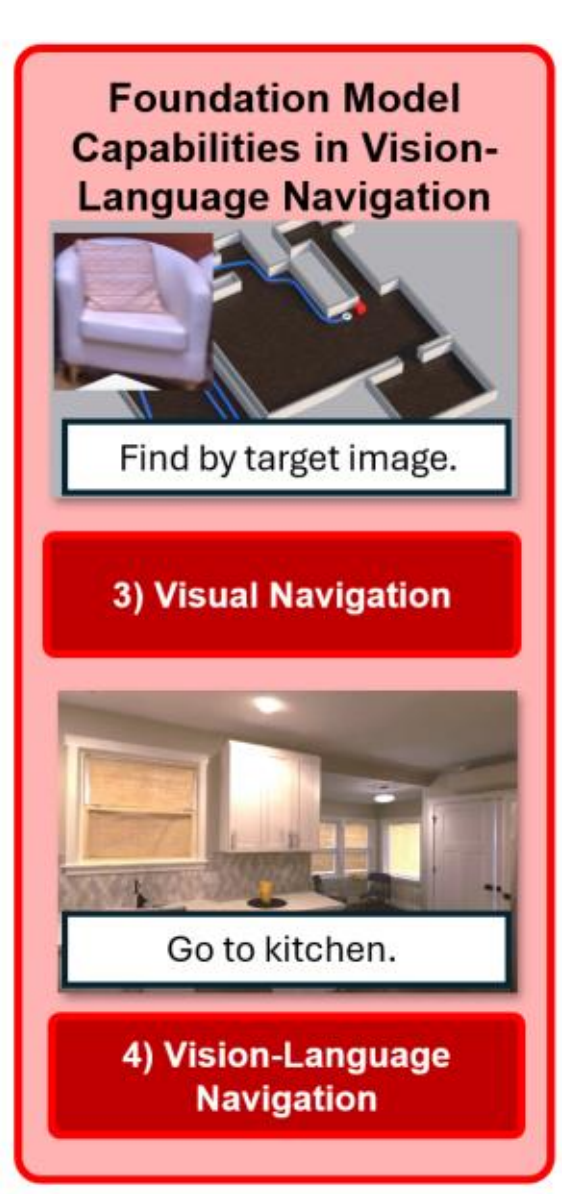

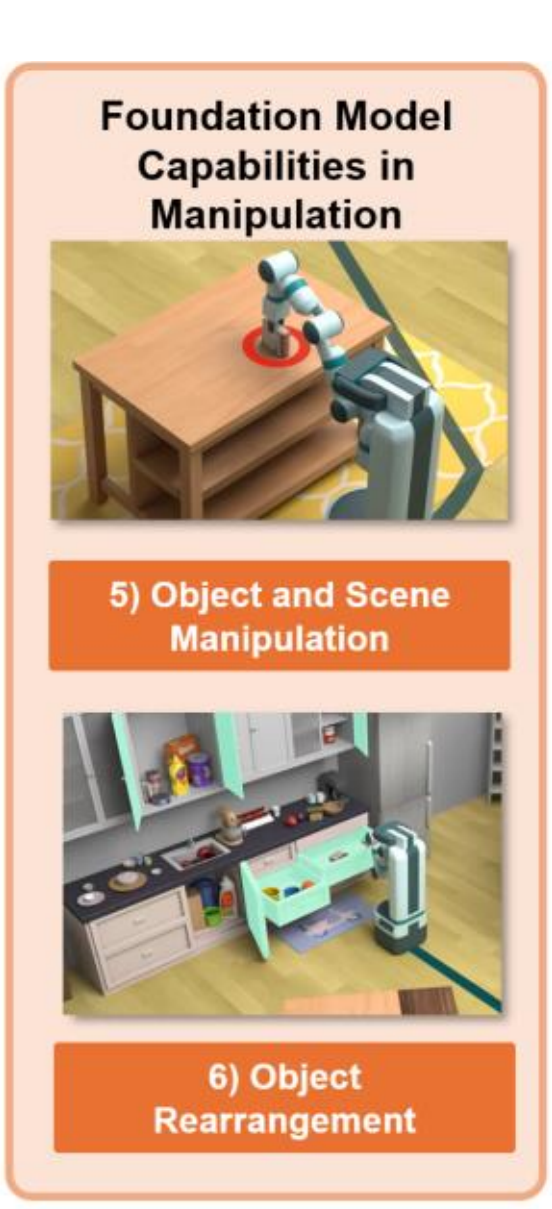

**Figure 4.** Foundation models support communication, navigation, and physical interaction in service robots. Images for 1-4, and for 5 & 6 obtained from the Habitat 1.0 [61] and Habitat 2.0 [150] Simulators, respectively.

### *3.1. Addressing Challenge #1: Translation of Natural Language Instructions into Executable Robot Actions*

Understanding and executing natural language instructions remains a core challenge for mobile service robots. In response, foundation models leverage language-conditioned policies [12], vision-language alignment [151], and multimodal reasoning [152], [153] to bridge the gap between symbolic language and embodied robotic control. In particular, recent foundation models have advanced instruction-following in mobile service robots by addressing the symbolic-to-embodied gap [154]-[160], incorporating physical commonsense [161], [162], and supporting long-horizon planning and adaptation [163]-[165]. We discuss these key advances below.

*1) Symbolic-to-Embodied Mapping and Instruction Ambiguity*: Non-expert users can provide vague or underspecified requests that mobile service robots must translate into precise, embodied actions. Unlike classical planners that depend on brittle rule sets [166], foundation models learn shared semantic spaces where language, perception, and control are jointly represented. This enables robots to resolve ambiguities in context-sensitive ways, for instance, interpreting a parent's request to "*put this somewhere safe*" in a home by linking it to spatial cues like a shelf or cabinet. Models such as LIMO [154], a reasoning-focused language model, extends this capability of resolving ambiguities by retaining semantic context across steps, allowing robots to adapt dynamically when users clarify or modify instructions mid-task. VLMs and MLLMs further support disambiguation by grounding colloquial or multi-modal inputs in real-world layouts, ensuring that actions remain compliant with HRI norms such as safe distances, legibility of motion, and responsiveness to feedback during task implementation.

*2) Lack of Embodied Commonsense and Physical Constraints Awareness*: In general, mobile service robots can use foundation models to address the absence of physical commonsense, by incorporating: 1) physics-based priors [167], 2) affordance reasoning [161], and 3) constraint-aware planning [168] into language-grounded policy architectures. These capabilities address the embodied AI gap where robots often lack real-world intuition about object properties, support stability, or task feasibility which is critical in mobile service contexts. For example, Genesis [167], a generative physics simulation foundation model, encodes differentiable physics into object-centric representations, allowing robots to mentally simulate stacking, collision, or deformation outcomes before acting. This prevents unsafe task errors, such as placing medical supplies on unstable surfaces. By combining affordance reasoning with constraint-aware planning, foundation models enable mobile service robots to anticipate whether actions like "*reposition the IV pole beside the bed without tipping it*" or "*push the luggage cart through a narrow gate*" are physically realizable given their embodiment. This awareness is especially critical in caregiving and domestic contexts, where failure to respect physical and social constraints can potentially endanger others.

*3) Failure in Long-Horizon Task Planning*: Long-horizon service tasks require mobile service robots to maintain memory, adapt to interruptions, and replan under uncertainty. Foundation models address these challenges by: 1) decomposing high-level language into structured subtasks [163], [164], 2) retaining goal-conditioned histories to preserve context [164], [165], and 3) revising plans in response to dynamic sensory input [169], [170]. For instance, Code-as-Policies [163], a code-writing LLM, can translate robot assistive commands into modular control routines that allow task branching, while LLM-Planner [164], an LLM-based planner, adapts ongoing navigation when new verbal clarifications arise. Similarly, s1 [165], a reasoning-optimized LLM, dynamically adapt reasoning depth to prevent cascading errors in extended sequences. These advances allow mobile service robots to sustain coherent, user-aligned task progression in complex environments, whether completing multi-room domestic routines (e.g., fetch-and-deliver sequences that span kitchen → living room → bedroom), replanning mid-shift in hospitals during time-sensitive medical delivery under corridor congestion and changing staff priorities, or adapting customer guidance and wayfinding in public venues (e.g., malls or airports) when routes, destinations, or crowd flow change in real-time.

Prior approaches to language-guided robot planning such as STRIPS [39], PDDL [42], [43], STN [73] and FF [74] relied predominantly on hand-engineered symbolic planners and task-graph formulations, and classical heuristic planning. While effective in static and well-specified domains, these methods were brittle to linguistic ambiguity, underspecified user instructions, and dynamic task variation [39]. Learning-based sequencing methods such as NTGs [77] and option-based policies [78] introduced greater flexibility, but still suffered from limited memory, compounding errors, and weak grounding when deployed in long-horizon mobile service tasks. Recent foundation-model approaches have begun to bridge the gap between symbolic reasoning and embodied execution, as shown in Table 2. In particular, CLIP-based Code-as-Policies (CLIP-CAP) [163] explicitly links vision-language representations to executable control programs, allowing natural language commands to be translated into modular, interpretable robot code that directly invokes perception, navigation, and manipulation primitives. Through this design, CLIP-CAP [163] achieves the highest reported real robot manipulation success (71% across real-robot manipulation tasks) among the evaluated models, while also supporting task branching and online replanning during execution. This capability marks a shift beyond earlier brittle planning pipelines towards more adaptive, language-driven mobile service robot behaviors.

**TABLE 2**

COMPARISON OF MAJOR FOUNDATION MODELS WITH RESPECT TO THE CORE CHALLENGES FOR MOBILE SERVICE ROBOTS

| Model | Language-to-Action (Success Rate) | Perception (Accuracy / Latency) | Uncertainty (Calibration Error) | Computation (GFLOPs (giga floating point operations/s) / FPS) | Notes (Strengths / Limitations) |
| --- | --- | --- | --- | --- | --- |
| Magma [145] | 52.3% (GoogleRobot) | Vision QA: 88.6%<br>~220-300 ms | — | ≈45 GFLOPs / ~30 FPS | + Strong multimodal perception<br>- Limited robotics action success; partial spatial reasoning |
| DeepSeek-R1 [144] | — | MLLU:<br>84%<br>~300–400 ms | 81.8% | ≈90 GFLOPs / ~18 FPS | + Best uncertainty-aware reasoning<br>- No action grounding or real-robot evaluation |
| SAM 2 [171] | — | Segmentation mIoU:<br>86%<br>~50 ms CPU | — | 260–310 GFLOPs / 25-44 FPS | + Highest visual perception accuracy<br>- Cannot produce or ground action commands |
| CLIP-CAP [163] | 71% across 7 manipulation tasks<br>(Everyday Robot) | Goal recognition: 83%<br>~200 ms | — | 175 GFLOPs / 3-5 FPS | + Best physical manipulation success reported<br>- Low FPS hinders onboard deployment |
| Perceiver IO [172]<br>for<br>Perceiver-Actor [173] | 70% across 18 tasks<br>(Franka Panda Robot) | Perceiver-IO scene encoding:<br>84.5%<br>~250–350 ms | — | 9–10 GFLOPs / 2-4 FPS | + Lowest computational cost with multi-task robustness<br>- Uncertainty not explicitly modeled |
| GPT [139] | 85-92% structured question and answering<br>(Franka Panda Robot) | MMLU:<br>90%<br>~700 ms multimodal | 92.0% | ≈800 GFLOPs/token → 1-3 FPS | + Strongest high-level reasoning under ambiguity<br>- Computationally prohibitive for onboard guidance |
| LLaMA [174] | ~91% symbolic planning<br>(Google Robot) | MMLU:<br>83.6%<br>250–350 ms | 92.1% | ≈8 × 10³ GFLOPs/token → ~0.7 Hz | + Effective symbolic planner; strong long-context utility<br>- Slow inference; lacks perception-action grounding |
| **Top Performer Per Challenge** | CLIP-CAP (best success rate in manipulation tasks) | SAM-2 (highest perception accuracy & real-time latency) | DeepSeek-R1 (lowest reported calibrated reasoning error among major GPT-LLaMA models under uncertainty on HLE text-only tasks) | Perceiver-Actor (lowest GFLOPs with viable FPS onboard) | Demonstrates clear trade-offs; no single model excels across all four core challenges |

Note: "—" means not available.

*3.2. Addressing Challenge #2: Multimodal Perception*

Robust multimodal perception is essential for interpreting complex environments. Particularly, multimodal VLMs and MLLMs can address this challenge by providing pre-trained multimodal representations of visual, textual and spatial features that enable the integration of heterogeneous sensory data through unified representation learning, temporal synchronization, uncertainty-aware fusion, and domain generalization [147], [175]. We discuss below the key advances of leveraging foundation models for this challenge.

*1) Cross-Modal Representation Gap*: Foundation models mitigate the fragmentation of sensory streams in mobile service robots by aligning multimodal data into unified latent spaces using either: 1) cross-modal tokenization [145], [141], 2) spatial encoding [147], and/or 3) attention-based fusion [176] to create representations that not only synchronize perception but also remain socially meaningful. Unlike robots which operate in structured factories or warehouses, mobile service robots must integrate heterogeneous inputs, vision, speech, LiDAR, tactile, and physiological signals in real-time, while interacting with humans in cluttered or crowded environments. Magma [145] is a transformer-based multimodal foundation model designed to bridge perceptual understanding and embodied actions through Set-of-Mark (SoM) and Trace-of-Mark (ToM) prompting, which convert visual scenes into actionable spatial tokens and predict future motion trajectories. By integrating images, language commands, and robot traces into a joint token space, Magma enables unified cross-modal grounding and planning. This allows a service robot deployed in a hospital to correctly associate a nurse's spoken instruction with a patient's movement, while in retail environments the same model can interpret gestures such as pointing and ground expressions like "*that one over there*" in crowded shelves. By unifying perception and interaction modes, foundation models can reduce representational misalignment that can result in failures in object association, gesture-speech coordination, and/or scene parsing during real-time HRI.

*2) Latency Issues*: Strategies for reducing latency in the service robot context include: 1) learning time-aware representations [177], 2) aligning asynchronous inputs [178], and 3) dynamically adjusting modality-specific update rates [179]. For learning time-aware representations, video foundation models such as VideoJAM [177] and InternVideo2 [178] jointly model visual appearance and optical-flow motion across time, enabling coherent spatiotemporal grounding of asynchronous frames. In a healthcare environment, for example, such models allow a robot to detect in real-time when a patient begins to stand up from a chair, triggering immediate stabilization assistance or verbal support, before a fall risk escalates. In office environments, object-centric physical modeling frameworks further help synchronize sensor inputs during long hallway robot navigation when groups of employees are sharing the space while having unpredictable movements, enabling timely monitoring and smooth flow management. These advances allow mobile service robots to have real-time responsiveness in environments where even small delays can undermine safety, trust, and quality of assistance.

*3) Uncertainty Propagation Across Modalities*: Mobile service robots must assess the reliability of multimodal inputs in real-time, since sensor degradation in one channel can otherwise cascade into unsafe actions. Foundation models enable uncertainty-aware fusion, where each modality's contribution is dynamically weighted by estimated confidence [180]. For example, DeepSeek-R1 [144], a reinforcement-trained LLM, and Segment Anything Model (SAM) [159], SAM-2 [171] vision foundation models, incorporate confidence-driven suppression of low-quality visual regions and reinforcement-based reasoning to avoid acting on uncertain observations. These mechanisms can be directly leveraged in service environments: in restaurant service contexts, confidence-aware fusion can prevent unsafe misinterpretation of overlapping table orders and simultaneous calls. In service automation, such as busy airports or malls, adaptive reweighting allows robots to filter noisy crowd signals while sustaining situational awareness. These strategies enable mobile service robots to maintain robust perception and socially safe interaction even when modalities degrade.

*4) Domain Adaptation and Transferability*: Service robots need to generalize across high variability, from clutter and lighting variations, to reflective areas, or to reconfigurable layouts and dense crowds. Unlike factory or lab-trained models, foundation model-based perception systems deployed in service robots cannot be continuously retrained during operation due to privacy, compute, and real-time performance constraints. Therefore, strong domain transfer capabilities are essential. Foundation models address this need through: 1) self-supervised pretraining [181], [182], which allows perception systems to adapt without large labeled datasets, 2) cross-domain token alignment [175], [183], which stabilizes multimodal features when sensor inputs vary across environments, and 3) few-shot generalization [184], [168], which dynamically adjusts tokenization for domain-flexible perception. These capabilities help mobile service robots obtain reliable recognition and reasoning under changing environments. This reduces perceptual drift and ensures continuity of safe, trustworthy interaction over long-term deployments.

Prior to foundation models, mobile service robots struggled with cross-modal alignment, noisy sensing, and limited generalization across objects and environments [47]-[49]. The perception results in Table 2 demonstrate that foundation models can address these concerns. Namely, SAM-2 [171] achieves the highest visual perception accuracy with real-time performance, while Magma [145] improves semantic grounding through unified image-language embeddings maintain competitive sensing at low compute cost.

### *3.3. Addressing Challenge #3: Uncertainty Estimation*

Robust uncertainty estimation for mobile service robots enables such robots to work in unpredictable environments, where sensor noise, ambiguous input, and unexpected events can compromise decision-making. Foundation models address this challenge by supporting explicit confidence modeling, prediction long-horizon risk prediction, and uncertainty-aware HRI [172]-[195]. We discuss these below.

*1) Lack of Explicit Uncertainty Quantification*: In human-facing environments, robots must be able to avoid unsafe or socially awkward actions. Foundation models support this through: 1) reinforcement-guided value modeling [144], 2) adaptive attention-based reweighting [172], [185], [139], and 3) temporal uncertainty tracking [186]. The LLM DeepSeek-R1 [144] calibrates confidence using reinforcement signals, while the MLLM GPT-4o [139] models token-level reliability. These capabilities allow mobile service robots to cautiously perform actions in everyday contexts. Benchmarks like Humanity's Last Exam (HLE) [196] and MANNERS-DB [197] further highlight the importance of expressing uncertainty in socially acceptable ways, such as apologizing before asking for clarification or signaling doubt through tone and timing, which is essential for maintaining trust in domestic and public service interactions.

*2) Failure in Long-Horizon Uncertainty Estimation*: Planning over extended time horizons presents a major challenge for mobile service robots, as perception or reasoning errors can accumulate into significant risks during prolonged operation. Foundation models mitigate this through latent world modeling and temporal abstraction [187]-[189], which simulate outcomes in compressed latent space to anticipate risks such as localization drift, occlusion, or sensor degradation before unsafe actions are executed. Vision-based self-supervised models such as MAE [190] and Swin Transformer [191] further enhance robustness by reconstructing corrupted spatiotemporal patches, and improving confidence tracking under partial observability. In service applications, this enables long-horizon replanning for robotic tasks such as guiding visitors through multi-floor buildings where the availability of elevators or access restrictions can vary dynamically. By modeling uncertainty accumulation explicitly, these approaches support continuity of assistance and maintain reliability during extended, socially embedded tasks.

*3) Uncertainty in Human-Robot Interactions (HRI)*: Effective HRI requires mobile service robots to detect uncertain or ambiguous human behaviors such as vague speech, inconsistent gestures, or shifting gaze, and to communicate their own uncertainty clearly during task execution. Foundation models support this by integrating behavior forecasting, social intent grounding, and natural language grounding. For example, Lumiere [192], a diffusion-based generative video model, generates motion-consistent forecasts of human activity, enabling robots to anticipate possible human actions and avoid socially disruptive responses. In a domestic setting, this may involve the robot briefly pausing to signal uncertainty and then proactively asking the user a clarifying question, for example, "*Did you mean the unwashed bowls on the table or the ones by the sink?*" thereby preventing incorrect actions while maintaining smooth interaction flow.

Traditional mobile service robots often did not account for uncertainty, leading to unsafe behaviors and brittle recovery when unexpected human interactions occurred [55], [56]. In contrast, recent foundation models demonstrate measurable progress toward calibrated reasoning under uncertainty. For instance, DeepSeek-R1 [144] reports the lowest calibration error (~81.8%) on the HLE benchmark [196], indicating closer alignment between predicted confidence and actual correctness, whereas GPT [139] and LLaMA [174] class models generally exhibit higher calibration error values (~92%), demonstrating overconfidence. As summarized in Table 2, this improvement suggests that certain foundation models are beginning to quantify and communicate uncertainty more reliably, an ability that is essential for safe, trust-preserving service robot deployment in real-world environments.

### *3.4. Addressing Challenge #4: Computational Capabilities*

Mobile service robots often operate under constrained budgets, making real-time perception and planning with large deep learning and symbolic models difficult. Foundation models address this challenge by enabling scalable model compression [198], [199], efficient attention distillation [200], [170], adaptive token processing [201], [202], and on-device execution [203], [204]. These advances allow mobile robots to maintain real-time responsiveness, energy efficiency, and task throughput even under strict power, memory, and latency constraints. We discuss these key advances below.

*1) Perception and Planning Computational Overhead*: Real-time perception and planning can often exceed the computational limits of embedded mobile service robot systems. This creates latency bottlenecks that impair decision-making during navigation, interaction, or manipulation. Foundation models address computation limits through: 1) architectural streamlining [205], [146], 2) model compression [198], [199], and 3) efficient attention mechanisms tailored for edge devices [170], [200]. The LLM EdgeFormer [205] applies dynamic token control to sustain responsiveness during dialogue-driven task execution. The VLM DeiT [198], [142] a distilled vision transformer, compresses large transformer architectures into compact models suitable for identifying small objects in cluttered environments. The MLLM VisionLLaMA [199] adapts LLaMA-style transformers for lightweight multimodal reasoning. These advances reduce inference latency while preserving reliability, allowing mobile service robots to operate efficiently under power and memory constraints while maintaining safe, socially responsive interactions.

*2) Lack of Adaptive Resource Allocation for Real-Time Inference*: Efficient task execution under fluctuating computational loads is essential for mobile service robots, which must remain responsive despite strict latency and energy limits. Foundation models address this challenge through adaptive computation strategies that allocate resources in proportion to service robot task complexity. For example, the LLMs Switch Transformer [201] and GShard [202] use Mixture-of-Experts (MoE) architectures to selectively activate the modules required for the current context, conserving energy during routine operations while scaling up capacity during socially or perceptually demanding interactions. This dynamic allocation directly enhances service quality: a domestic assistant robot can transition from monitoring to full-capacity multimodal parsing of requests. By dynamically rebalancing inference loads, foundation models enable mobile service robots to maintain timely, safe, and socially fluent interactions.

Onboard processors and battery constraints restricted mobile service robots to simplified models with reduced perception and reasoning abilities. Meaningful improvements have been achieved with foundation models. Namely, Perceiver-Actor offers the lowest compute cost (~9-10 GFLOPs) while still achieving multi-task visuomotor control, and SAM-2 and Magma deliver strong perception with higher computation budgets.

Addressing the core challenges of multimodal perception, language grounding, task generalization, uncertainty estimation, and computational efficiency, foundation models offer scalable solutions for mobile service robotics.

## 4. Real-World Applications for Mobile Service Robots with Embedded Foundation Models

The global service robotics market is projected to experience rapid growth, approximately doubling from $47.10 billion USD in 2024 to $98.65 billion USD by 2029, with an average compound annual growth rate of 15.9% [206]. This surge reflects broader trends toward integrating intelligent service robots into homes, healthcare facilities, and public spaces [207], [208]. As foundation models are increasingly being embedded into mobile service robots, we examine how these models address real-world needs and enable safer, more adaptive, and socially intelligent robot behaviors across these domains.

We categorize applications into three primary domains, domestic assistance, healthcare assistance, and service automation, with each decomposed into representative sub-tasks, Figure 5.

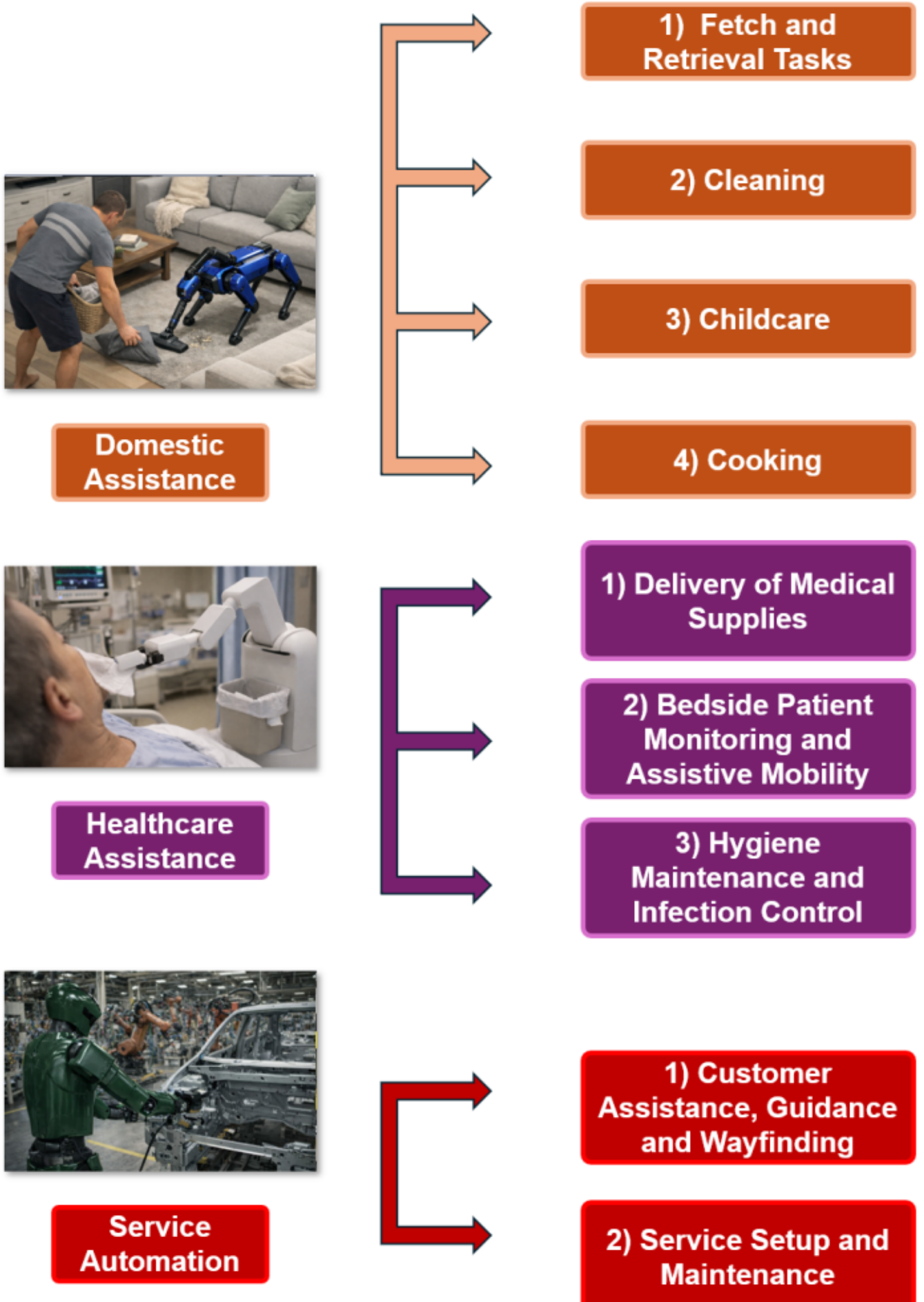

**Figure 5.** Mobile service robots operate across domestic, healthcare, and service automation domains. Images generated using the Nano Banana Pro AI Image Generator.

Domestic assistance consists of fetch-and-retrieval, cleaning, childcare, and cooking activities; healthcare assistance includes medical delivery, bedside monitoring, assistive mobility, and hygiene maintenance; service automation encompasses customer assistance, guidance and wayfinding, as well as service setup and maintenance tasks such as venue preparation, layout arrangement, equipment positioning, and post-event cleanup in public spaces. Representative frameworks and robot platforms in each application domain, are presented in Table 3, linking existing robotic systems (e.g., Stretch RE-1, Fetch, Everyday Robot, Jackal, and mobile manipulation platforms) to the tasks they perform. Together, Figure 5 and Table 3 provide a mapping from application domains and subtasks to specific foundation model-enabled mobile service robot implementations that we examine in the following subsections.

**TABLE 3**

FOUNDATION MODEL APPLICATIONS ACROSS SEVERAL ROBOT DOMAINS

| Domain | Task | Representative Frameworks | Robots |
|---|---|---|---|
| **Domestic Assistance** | **Fetch and Carry Tasks** | Ok-Robot [209] | Stretch RE-1 Robot |
| | | Octo [210] | Franka Emika Panda Robot |
| | | CogNav [30] | Fetch mobile manipulator |
| | | PARTNR [211] | Franka Emika Panda Robot<br>Spot + 7-DoF arm |
| | **Cleaning** | TidyBot [37] | Holonomic mobile base<br>and a Kinova Gen3 arm |
| | | RT-2 [14] | Everyday Robots mobile manipulator |
| | **Childcare** | Smart Help [28] | Custom mobile manipulator robot<br>with Kinova Gen3 arm |
| | | SAFE [212] | Franka Emika Panda Robot |
| | **Cooking** | OVQSRFRW [213] | Everyday Robots mobile manipulator |
| | | TPLC [214] | 6-DoF Universal Robots UR5e arm |
| | | CTGN [215] | Dual-arm Universal Robots UR5e |
| **Healthcare Assistance** | **Delivery of Medical Supplies** | ProgPrompt [216] | Franka Emika Panda Robot |
| | | LLM-Grounder [217] | Franka Emika Panda Robot |
| | | VLM-Social-Nav [193] | Clearpath Jackal Robot |
| | **Bedside Patient Monitoring** | MoMa-LLM [193] | Fetch mobile manipulator |
| | **Assistive Mobility** | OLiVia-Nav [194] | Clearpath Jackal Robot |
| | | GSON [218] | Custom differential-drive robot |
| | **Hygiene Maintenance and Infection Control** | AutoRT [219] | Everyday Robots mobile manipulator |
| | | SayPlan [220] | Franka Emika Panda Robot |
| **Service Automation** | **Customer Assistance, Guidance and Wayfinding** | OVSG [221] | Ackermann Steering Robot |
| | | LM-Nav [222] | Clearpath Jackal Robot |
| | | ViNG [223] | Vizbot, Unitree Go1, Clearpath Jackal, and LoCoBot |
| | **Service Setup and Maintenance** | Hi Robot [224] | Dual-arm mobile manipulator<br>Mobile ALOHA robot |
| | | $\pi_{0.5}$ [149] | Dual-arm mobile manipulator<br>Mobile ALOHA robot |
| | | PhysObjects [225] | Franka Emika Panda Robot |

*4.1. Domestic Assistance*

Domestic assistance refers to everyday tasks to support comfort, safety, and quality of life [226], [227]. These tasks include the fetch and retrieval of household objects [210], [211], [228], [229], cleaning [37], [230], childcare assistance [231], [148], and cooking activities [232]-[234].

*1) Fetch and Retrieval Tasks*: These tasks, often triggered by natural language, gestures, or contextual cues, involve multimodal perception, ambiguous instruction grounding, uncertainty management, and efficient planning. Fetch and retrieval tasks decompose into sub-tasks including: 1) object localization, 2) grasp planning and action refinement, 3) navigation and multi-step delivery, and 4) handover and placement reasoning. Using foundation models such as CLIP [141] and CLIP-Fields [235] for spatial-language grounding, Octo [210] for visuomotor control, GPT-4V [158] and CogNav [30] for high-level reasoning and route sequencing, and LLaMA3.1-8B [174] or CodeLLaMA-70B [236] for symbolic task decomposition, mobile service robots can achieve end-to-end perception and manipulation across these sub-tasks. VLM CLIP-Fields [235] support object localization by constructing a continuous 3D semantic field from RGB-D images and odometry. Each point in the environment is encoded using CLIP's text-vision embedding space, allowing a robot to search for objects using natural language descriptions (e.g., "*the yellow sponge near the sink, go get it and bring it to me so I can clean these dishes*") without predefined labels. This method has been deployed on the Stretch RE-1 [209], a commercial mobile manipulator equipped with an RGB-D sensor, telescoping mast, and a compliant gripper. The Stretch uses the semantic field to resolve which surfaces and object clusters have the highest likelihood to match the user's query, enabling reliable localization in typical household scenes where objects may be partially occluded, moved, or surrounded by visually similar items. Grasp execution and refinement are produced by the VLA model Octo [210], which blends a T5 language encoder with a diffusion-based visuomotor policy. On a Franka Emika Panda Robot arm, Octo generates stable grasp trajectories when object poses are uncertain or cluttered, refining motions based on both visual input and the language-specified goal. High-level instruction reasoning and route sequencing are modeled through GPT-4 [143], GPT-4V [158], and segmentation backbones such as SAM [159]. Integrated onto the Fetch mobile manipulator, these models interpret user requests, segment relevant regions of the scene, and adjust navigation plans as environmental obstacles or room layouts change. The CogNav system [30] further extends spatial reasoning by using GPT-4V [158] to decide when to explore, when to navigate, and how to map new rooms during longer fetch-and-deliver routines. Context-sensitive handover and placement actions are enabled through the PARTNR framework [211], which applies LLaMA3.1-8B [174] and CodeLLaMA-70B [236] to infer appropriate approach angles, placement surfaces, and human-aware delivery strategies. PARTNR demonstrated these behaviors on both a Franka Panda and a Boston Dynamics Spot robot equipped with a 7-degrees-of-freedom (DoF) arm, enabling the retrieval and socially compliant delivery of household items during domestic assistance tasks.

*2) Cleaning*: Cleaning tasks involve a robot autonomously organizing, decluttering, and sorting household items into designated locations [237]. Unlike object-specific fetch tasks, cleaning often requires generalizing abstract commands such as "*put everything away*" into sequential, context-aware behaviors across many objects and categories. Cleaning tasks can be decomposed into the following sub-tasks: 1) user preference inference and object categorization, 2) object localization and sorting, and 3) task execution with multimodal grounding and action control. Foundation models such as GPT-3.5 [168] or LLaMA-based LLMs [238], [174], [236] for preference reasoning, VLM CLIP [141] for visual-semantic association, and embodied VLA models such as RT-2 can be used for physical execution. These models collectively support the subtasks in real-world cleaning robots. User preference inference and category reasoning are demonstrated in TidyBot [37], which integrates LLM GPT-3.5 [168] with a holonomic mobile base and a Kinova Gen3 arm. GPT-3.5 [168] is prompted with a small set of examples (e.g., "*put light clothes in the drawer*," "*place toys in the bin*") and infers general placement rules that align with household organization habits. The model predicts appropriate storage locations for previously unseen objects, enabling the robot to follow human-specific conventions rather than relying on rigid predefined templates. Object localization and sorting are enabled through the combination of GPT-3.5 and CLIP [141] on the same TidyBot robot, which together match visual observations to semantic categories. CLIP provides an open-vocabulary visual embedding space, allowing the TidyBot to recognize a wide range of everyday objects, even those not present in its training set, and associate them with the inferred categories generated by the LLM. This supports reliable sorting in typical home scenes where objects vary in appearance, lighting, or placement.

Task execution and multimodal control can be supported by VLA models such as RT-2 [14], which unify vision, language, and motor control. RT-2 policies map high-level cleaning commands (e.g., "*clear the table,*" "*place everything neatly on the shelf*") into low-level robot actions while maintaining awareness of object poses, workspace boundaries, and user-defined organization rules. This method has been deployed on the Everyday Robot [14], a mobile manipulator with a single 6-DoF arm. The VLA model enables the Everyday Robot to carry out multi-step cleaning routines with consistency and adaptability.

*3) Childcare*: Childcare tasks require a mobile service robot to assist with daily routines, object retrievals, and monitoring of young children [239]. Example tasks include responding to prompts such as "*bring me my blanket*," guiding hygiene routines like teeth brushing after meals, and detecting safety risks such as toddlers approaching stairs or handling small objects. Childcare can be decomposed into following sub-tasks: 1) behavioral intent grounding, 2) dynamic instruction decomposition, and 3) real-time multimodal control and safety adaptation. Using foundation models such as GPT-3.5/3.5-turbo-instruct [168] for intent interpretation and multi-step reasoning, and VLA safety-adapter frameworks such as SAFE [212] for embodied risk detection, mobile service robots can integrate perceptual cues, language inputs, and feedback signals to support safe and context-aware childcare routines. LLM GPT-3.5-turbo-instruct [168] supports language-grounded behavioral reasoning by learning probabilistic mappings between behavioral cues and task goals, allowing a robot to interpret vague requests such as "*help me*" in context-specific ways; for example, in retrieving a dropped toy, handing over a comfort item, or alerting a caregiver. This has been demonstrated on the Smart Help system [28] which uses a custom mobile manipulator robot with an onboard camera and a Kinova Gen3 arm. Smart Help uses GPT-3.5-turbo-instruct [168] to infer user intent from short or underspecified prompts. The LLM supports dynamic instruction decomposition, where, for example, longer caregiver commands (e.g., "*check if the baby is asleep and then turn off the light*") can be translated into ordered subgoals that coordinate perception, navigation, and verification steps, allowing a robot to maintain situational context and align timing with child behaviors. Real-time multimodal control and safety adaptation are provided by SAFE [212], which integrates a multimodal safety classifier atop pretrained LLM- and VLM-based policies. The Franka Emika Panda robot has used SAFE to continuously analyze vision, proprioception, and trajectory predictions to halt unsafe grasps and prevent collisions with fragile objects. SAFE gives a timely alert such that the robot can stop, backtrack, or ask for help, enabling early intervention before failures cause harm to humans, adults and children, around.

*4) Cooking*: Cooking tasks involve mobile service robots executing multi-step procedures to assist elderly individuals, busy families, or users with disabilities. Cooking includes such subtasks as: 1) ingredient and utensil localization, 2) sequential cooking task decomposition and planning, and 3) bimanual manipulation and coordinated action execution. Foundation models such as LLM PaLM 540B [6] for visual-language grounding, LLMs BERT [140] and DistilBERT [240] for instruction parsing, and VLM CLIP [141] for visual-semantic correspondence, can map natural language cooking requests to perception-driven subgoals and physically executable motion plans in the home. In particular, the LLM PaLM 540B supports ingredient and utensil localization by visually identifying objects such as spoons, milk, eggs, or bread from natural language cues. This allows a robot to ground ingredient or utensil references directly to camera observations and infer their spatial context within the kitchen. This was demonstrated on the Everyday Robots mobile manipulator [213] which integrated the PaLM 540B LLM. Sequential task decomposition and planning have been demonstrated using the fixed-base 6-DoF Universal Robots UR5e arm system [214] . The system used the BERT and DistilBERT as well as CLIP to transform high-level goals such as "*in kitchen, make a sandwich*" into grounded action sequences like "*pick bread from cabinet above*" or "*retrieve cooked meat from microwave.*" These models decompose abstract cooking goals into structured subtasks linked to corresponding visual targets. Coordinated bimanual manipulation are enabled through the combination of GPT-3 [2] and a graph-based control network to synchronize grasping, tool use, ingredient placement, and force-sensitive interactions. This method has been deployed on a dual-arm Universal Robots UR5e configuration [215] utilizing GPT-3 as the high-level planner taking language instruction and RGB camera data as inputs. This integration allows the robot to perform LLM-guided actions such as stabilizing a bowl with one arm while stirring or transferring ingredients with the other, all while maintaining real-time perceptual grounding and safety constraints.

*4.2. Healthcare Assistance*

Healthcare assistance tasks involve supporting clinical workflows, assisting people, and infection control in dynamic hospital environments. These tasks include the delivery of medical supplies [216], [217], bedside patient monitoring, assistive mobility [241], [218], and hygiene maintenance and infection control [219], [220], [242].

*1) Delivery of Medical Supplies*: Delivery of medical supplies involves mobile service robots autonomously transporting medications, IV fluids, lab samples, and surgical instruments throughout hospitals and clinics. These deliveries are essential to maintaining smooth clinical workflows, reducing staff workload, and minimizing human error in time-sensitive medical tasks. Robotic delivery tasks can be decomposed into the following sub-tasks: 1) symbolic-to-action mapping and long-horizon planning, 2) spatial grounding of objects and target locations, 3) feasibility evaluation and adaptive plan selection, and 4) social navigation in dynamic clinical settings. LLMs such as Codex [243], GPT-3 [2], and GPT-4 [143] have been used for symbolic program synthesis and task decomposition, and VLMs such as OpenScene [244] for spatial grounding, and GPT-4V [158] combined with YOLO [245] for human-aware navigation. These models allow robots to translate requests into structured action sequences, infer 3D target locations, and adapt their motion to human activity in real-time. Codex [243] and GPT-3 [2] support symbolic-to-action reasoning by converting instructions such as "*take these surgical tools to the OR room*" into structured control sequences (e.g., functionally decomposed like grab(kit) → goto(prep_room) → handover) allowing a robot to autonomously generate interpretable, executable programs using an LLM-API and predefined spatial primitives to produce interpretable, long-horizon delivery routines. This is demonstrated on a custom mobile manipulator base equipped with a Franka Emika Panda arm [216]. Spatial grounding was enabled by a combination of models such as GPT-4 [143] and OpenScene [244]. Demonstrated on the same custom mobile manipulator base in [217], GPT-4 and OpenScene together resolves spatial expressions such as "*on the left tray by the door*" into 3D coordinates suitable for pick-and-place execution, enabling accurate supply localization and placement even when trays, carts, or equipment partially occlude the scene. GPT-4V [158] and YOLO [245] supports social navigation in dynamic environments as was shown with VLM-Social-Nav [193] on a Clearpath Jackal mobile robot equipped with RGB cameras and wheel odometry. VLM-Social-Nav merged GPT-4V and YOLO to interpret real-time human gestures and motion cues to infer social intent, dynamically adjusting its navigation policy to yield, reroute, or proceed safely through crowded corridors.

*2) Bedside Patient Monitoring:* Bedside patient monitoring focuses on understanding a patient's state, intent, and safety within the immediate vicinity of the bed, under strict clinical constraints on proximity, etiquette, and timely response. These tasks decompose into: 1) patient intent and state interpretation, and 2) bedside assistive decision-making. LLMs such as GPT-4 [143] and GPT-3.5-turbo-1106 [168] support these capabilities by reasoning over verbal cues, contextual room information, and clinical routines. This functionality is demonstrated in MoMa-LLM [241] on the Fetch robot, where GPT-4 [143] performs high-level reasoning over requests and situational context, while GPT-3.5-turbo-1106 [168] provides real-time room and object context classification. Together, these models enable robot bedside monitoring behaviors such as recognizing requests for assistance, detecting unsafe situations (e.g., a patient attempting to reach beyond safe range), and deciding when to reposition objects, alert staff, or initiate assistive actions, all while remaining localized to the bedside environment.

*3) Assistive Mobility:* Assistive mobility focuses on safely navigating dynamic hospital environments while anticipating human motion and proactively adapting behavior. These tasks decompose into: 1) safe co-navigation and obstacle adaptation and 2) multi-agent forecasting and proactive intervention. Vision-language models such as GPT-4V [158] and SC-CLIP [246] enable socially compliant navigation by grounding human motion cues, spatial layouts, and interaction affordances in real-time. This approach is demonstrated in OLiVia-Nav [194] on a mobile Jackal robot, which combines GPT-4V (offline) [158] for social-context captioning with SC-CLIP (real-time) [246] for responsive perception, allowing the robot to maneuver around people while maintaining appropriate spacing. For longer-horizon anticipation, GPT-4o [158] supports multi-agent behavior forecasting by analyzing pedestrian formations and predicting future motion patterns. This capability is demonstrated on a custom differential-drive robot equipped with RGB cameras and 2D LiDAR [218], where GPT-4o [158] enable proactive actions such as yielding to people, avoiding emerging congestion.

*4) Hygiene Maintenance and Infection Control*: In hospitals, hygiene maintenance tasks involve disinfecting high-touch surfaces, restocking personal protective equipment (PPE), managing biomedical waste, and monitoring air quality in areas such as ICUs, isolation wards, and operating rooms. Hygiene maintenance and infection control robot tasks decompose into sub-tasks: 1) spatial perception and contamination zone identification, 2) hygiene task grounding and low-level action planning, and 3) real-time PPE tool handling and proactive environmental monitoring. LLM GPT-4 [143], MLLMs such as AutoRT [219] and VLAs like SayPlan [220] enable robots to detect contamination risks, interpret abstract cleaning commands, and execute safe, context-aware disinfection procedures in clinical environments. AutoRT in [219] supports spatial perception and contamination zone identification by interpreting environment descriptions and linking them to actionable cleaning targets. AutoRT was integrated on the Everyday Robots mobile manipulator. AutoRT's MLLM is able to parse high-level natural language queries (e.g., "*disinfect all handles near patient beds")* under a robot-constitution safety framework and convert them into targeted manipulation instructions. This enables autonomous exploration of clinical rooms and prioritization of high-touch contamination zones. LLMs such as GPT-4 [143] support hygiene task grounding and low-level action planning by performing semantic reasoning over structured scene representations and generating long-horizon disinfection plans. This capability is demonstrated in SayPlan [220] on a Franka Emika Panda arm mounted on an Omron LD-60 mobile base, where GPT-4 [143] performed hierarchical semantic search over 3D scene graphs, which refines feasible disinfection steps across multi-room ICU layouts. This allows the robot to convert abstract instructions such as "*sanitize the entire ICU wing*" into sequential, spatially grounded actions. Real-time tool handling and proactive environmental monitoring is provided by VLA systems such as PaLM-E [13] by jointly coordinating visual perception, language reasoning, and manipulation. PaLM-E [13] is integrated with mobile manipulators to restock, manage tools, and with additional sensors can monitor indicators such as air quality, UV sanitation cycles, or filter status while avoiding contaminated regions during operation. PaLM-E's multimodal grounding allows a robot to adapt actions when new hazards emerge, such as say when staff flow alters the environment.

### *4.3. Service Automation*

Service automation tasks involve helping customers, managing events, and navigating dynamic public venues such as malls, airports, and museums. These tasks include customer assistance, guidance, and wayfinding [221]-[248] as well as service setup and maintenance [224], [225].

*1) Customer Assistance, Guidance and Wayfinding:* Mobile service robots in public venues such as malls, airports, and museums assist customers by interpreting requests, navigating large and dynamic spaces, and supporting multi-stage guidance under social constraints. These tasks decompose into sub-tasks: 1) language-to-goal spatial grounding, 2) multi-stage waypoint sequencing, and 3) symbolic-spatial planning for real-time urgency-based redirection. LLMs such as GPT-3 [2], GPT-3.5 [168], LLaMA [238], and VLMs like CLIP [141], and ViNG [223] enable robots to interpret open-ended user queries, associate them with real-world spatial layouts, and generate adaptive navigation plans that remain interpretable and socially aware in complex public environments. Language-to-goal grounding is achieved using GPT-3.5 [168] or LLaMA [238], which convert natural language queries into spatial targets and semantic constraints. This approach was used in OVSG [221] on the Ackermann-steered mobile platform ROSMASTER R2 equipped with RGB-D cameras. OVSG fuses GPT-3.5 [168] / LLaMA [238] with Detic-CLIP embeddings to build open-vocabulary 3D scene graphs encoding object labels, relations, and spatial geometry. As a result, the robot can interpret queries such as "*find the nearest pharmacy across from the information desk*" and resolve them into precise navigation goals in complex public spaces. LLMs such as GPT-3 [2] combined with VLMs CLIP and ViNG support multi-stage waypoint sequencing by breaking down language instructions into landmark-based routes and grounding those landmarks visually. This was shown in LM-Nav [222] on a Clearpath Jackal robot. GPT-3 [2] parsed instructions into ordered landmark sequences, CLIP matched these landmarks to visual observations, and ViNG predicted traversability between them via a topological connectivity graph. This enables robust execution of multi-stop instructions such as "*first go to security, then continue to Gate A12*," even when crowds, occlusions, or temporary barriers appear. Real-time symbolic-spatial redirection is provided by models such as GPT-3.5 [168], which can re-evaluate priorities and update navigation plans as conditions change. This functionality was demonstrated in MLLM-Search [249] and HAM-Nav [250] on a Jackal, where GPT-3.5 reprioritizes goals based on new constraints, such as flight gate changes, service desk relocations, or temporary crowding. This integration of symbolic reasoning and live spatial cues allows the robot to dynamically reroute while preserving safety and interpretability in busy public environments.

*2) Service Setup and Maintenance:* Service setup and maintenance tasks involve preparing venues, arranging equipment, adjusting layouts, and cleaning up during conferences, trade shows, or public spaces. Robots must interpret vague and evolving instructions such as "*set up chairs near the left projector*," "*tidy this booth area,*" or "*build this structure based on the manual.*" These sub-tasks decompose into: 1) adaptive motion skill chaining, 2) generalized layout planning and setup, and 3) spatial safety validation and manipulation. Using foundation models such as PaliGemma-3B [251] for multimodal grounding, $\pi_0$ [148] and $\pi_{0.5}$ [149] for language-conditioned visuomotor control, and PG-InstructBLIP [252] for physical attribute reasoning, mobile service robots perform adaptive skill sequencing, layout planning, and safe manipulation across these service setup and maintenance sub-tasks. LLM PaliGemma-3B [251] and VLA $\pi_0$ [148] support adaptive motion skill chaining, which together can translate vague service-setup instructions into structured sequences of low-level actions. This capability was demonstrated in Hi-Robot [224] using the dual-arm mobile manipulator Mobile ALOHA robot [234]. PaliGemma-3B interprets vague commands such as "*tidy this booth area*" or "*organize the display materials*," while $\pi_0$ sequences grasping, placing, folding, or lining motions. Human-in-the-loop feedback allows the model to correct or refine behaviors online, which enables robots to adapt skill chains as service conditions change. Generalized layout planning and setup is achieved through the VLA $\pi_{0.5}$ [149], which extends LLM [251] by co-training across heterogeneous tasks and spatial rearrangement demonstrations. This generalization is also shown on the Mobile ALOHA platform, where $\pi_{0.5}$ can learn to interpret commands such as "*set up chairs near the left projector*" or "*assemble the booth according to this layout*" even in unfamiliar rooms. By grounding natural language against perceived furniture, landmarks, and geometry, the robot can infer feasible arrangements and execute multi-step venue setup procedures. Spatial safety validation and manipulation are enabled by PG-InstructBLIP [252], a fine-tuned VLM capable of inferring physical attributes such as fragility, density, or structural integrity from images. PhysObjects [225] used a Franka Emika

Panda robot, where PG-InstructBLIP conditioned the robot's motion policy on object properties. For example, slowing movement near glass displays, adapting grip force for heavy metal stands, or selecting compliant strategies when handling sensitive electronics. This allows a robot to manipulate equipment safely and reliably.

An overview three-level framework is presented in Figure 6 which connects application domains to embodied AI challenges and shared foundation model capabilities. At Level 1, the framework spans the three aforementioned domains of domestic assistance, healthcare, and service automation. Level 2 organizes specific systems according to their perception, reasoning, and action components, highlighting how different frameworks (e.g., CLIP-Fields [235], Octo [21], MoMa-LLM [241], OVSG [221], Hi-Robot [224]) combine vision, language, and control. Level 3 captures shared model capabilities across domains, such as CLIP-style VLMs for perception, GPT-family LLMs for reasoning, and VLA models such as RT-2 [14] and SayCan [12] for action. This shows that many foundation models serve as reusable building blocks rather than domain-specific one-offs. This taxonomy clarifies how embedded foundation models generalize across diverse service settings while still supporting domain-specialized behaviors.

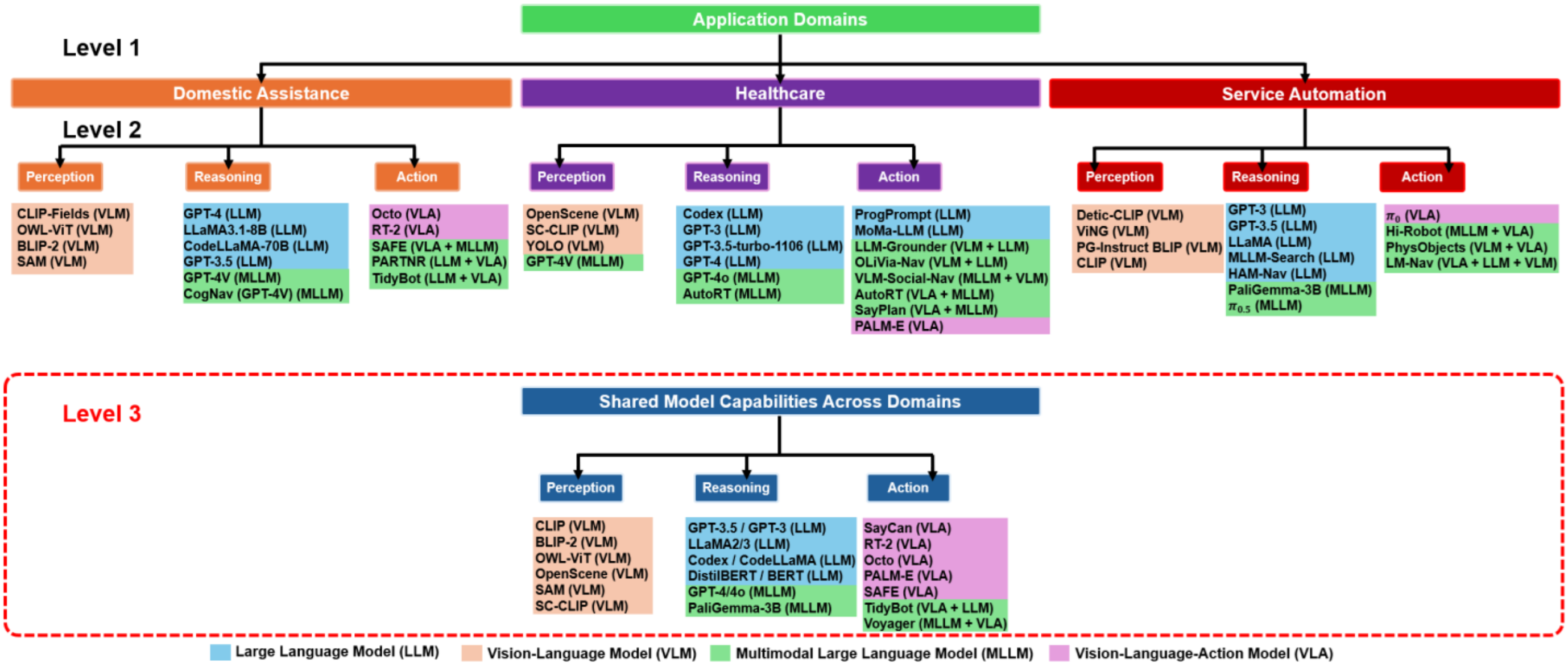

**Figure 6.** Three-level framework maps foundation models to perception, reasoning, and action across domains.

## 5. Ethical, Societal and Human-Interaction Implications for Mobile Service Robots with Embedded Foundation Models

Foundation models help close key technical gaps for mobile service robots (Section 3), but their use also creates deployment-time risks and responsibilities. The implications below should be read as requirements for safe adoption, not as arguments against foundation models. Each subsection pairs a risk with concrete mitigation hooks that align with the solutions discussed in Section 3-4.

Figure 7 provides an overview of these three interconnected domains of responsibility, ethical, societal, and human-interaction implications, showing how they collectively converge on the shared objective of ensuring safe and trustworthy mobile service robots. The triangular structure emphasizes that risks in one domain often propagate into the others, motivating integrated mitigation strategies throughout Sections 5.1 to 5.3.

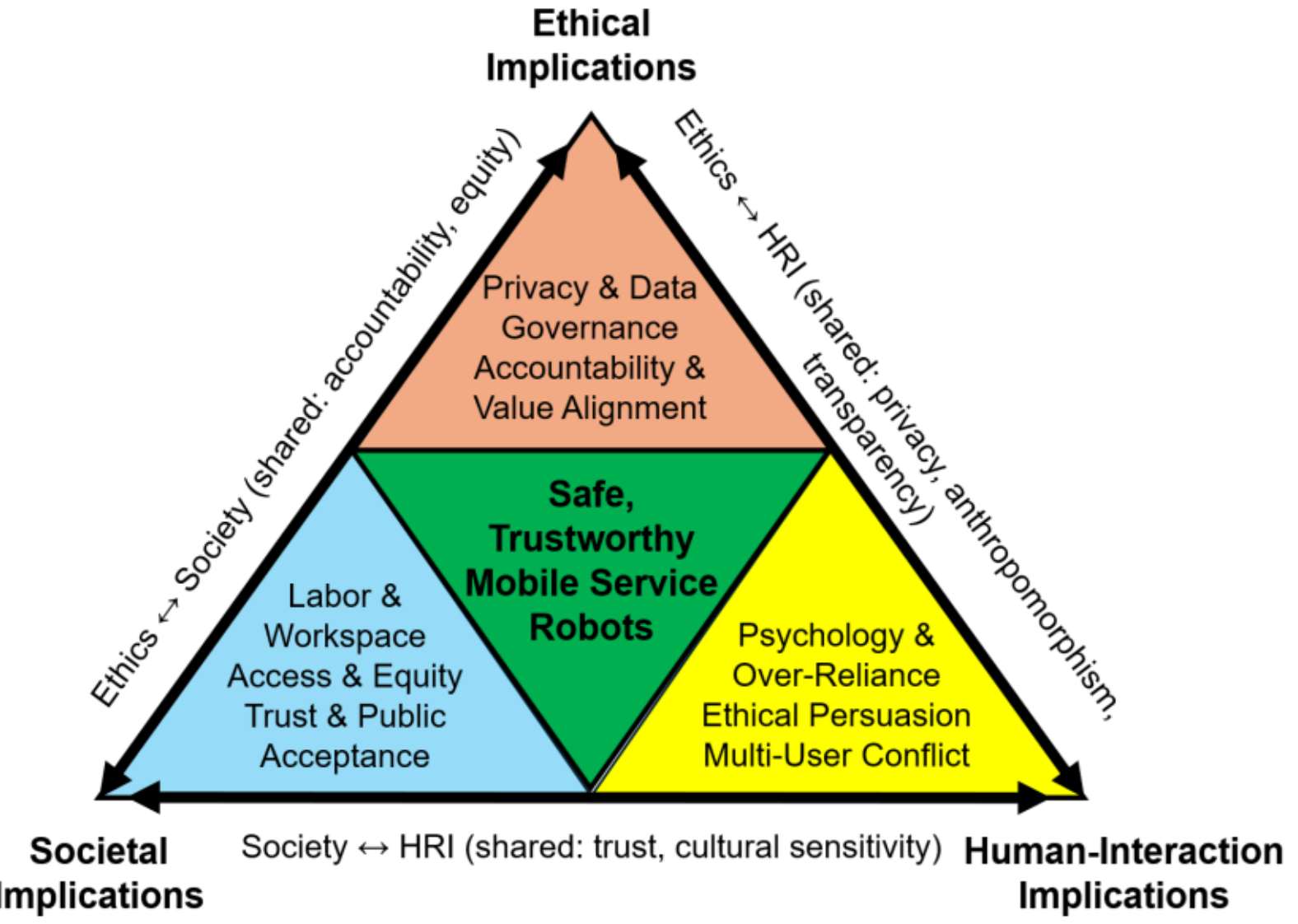

**Figure 7.** Responsible robotics spans ethical, societal, and human-interaction domains.

### *5.1. Ethical Implications*

Ethical implications concern the responsible use of foundation models in mobile service robots, focusing on protecting sensitive user data, ensuring accountability for autonomous decisions, and aligning robot behavior with human values.

*1) Privacy and Data Governance:* Mobile service robots deployed in homes or hospitals collect sensitive multimodal data (visual, audio, physiological) [253]. When coupled with foundation models trained on web-scale corpora, risks of leakage [254], [255], re-identification, or misuse increase [256]. Ethical deployment requires privacy-preserving architectures (on-device inference; selective edge–cloud collaboration), data minimization and purpose limitation, and auditable data-handling protocols aligned with the target domain.

*2) Accountability and Value Alignment:* When foundation models generate autonomous action plans, it becomes unclear who bears responsibility for harmful outcomes: the developer, operator, or model provider [257], [258]. Ethical frameworks should mandate action provenance (logging how language was grounded into plans), human-over-the-loop approval for high-impact tasks, and verifiable policy constraints that encode domain values (e.g., clinical safety norms).

### *5.2. Societal Implications*

Societal implications address how foundation model-powered service robots shape access, labor, and public trust, highlighting equity of deployment, transformation of human work roles, and the cultural acceptance of robots in daily life.

*1) Accessibility and Equity of Deployment:* If foundation model-powered service robots remain expensive, access will skew toward wealthier institutions [259], leaving vulnerable populations (elderly, rural communities) behind. Equitable deployment favors cost-efficient, compressed models (Section 3.4) and procurement or reimbursement mechanisms that extend access to underserved homes and care facilities [260].

*2) Labor and Workspace Transformation:* As foundation model-enabled robots become capable of handling service automation (hospital deliveries, retail stocking, cleaning), societal concerns arise about job displacement [261], [262]. At the same time, they may augment human workers by offloading repetitive or physically demanding tasks. Policy and design should prioritize augmentation (cobotic workflows and role redesign) rather than substitution, with training pathways for clinicians and staff to oversee foundation model-enabled service robots.

*3) Trust and Public Acceptance:* Societal trust in robots depends on transparency, reliability, and cultural sensitivity. Public acceptance will require mobile service robots which adapt to cultural norms [263], [264] and institutional standards, transparent behavior rationales (“*why this action and is this action a socially non-discriminatory one?*”), calibrated self-expression, and certification regimes appropriate to domestic and healthcare contexts [265].

### 5.3. Human-Interaction Implications

Human-interaction implications capture how foundation model-equipped service robots reshape user expectations, emotions, and trust, requiring careful design of emotional, psychological impact management, transparency and conflict resolution.

*1) Emotional and Cognitive Impact on Users:* Prolonged interaction with foundation model-equipped robots can influence user psychology [266]. While service robots may enhance companionship and reduce isolation in eldercare, they can also induce over-reliance, anxiety, or emotional confusion if their social behaviors mimic humans too closely without reliability [267]. Designers should tune anthropomorphism and disclosure (e.g., system status and capabilities) to reduce over-reliance and confusion, especially in eldercare.

*2) Ethical Persuasion and Influence:* LLM-based dialogue of mobile service robots enables persuasive interactions, reminding an elder to take medication or guiding a child to safe play. But persuasion risks manipulation [268], [269] if not carefully bounded. Research is needed on ethical persuasion policies that define limits for motivational dialogue in health, safety, or childcare contexts.

*3) Multi-User Interaction and Conflict Resolution:* Unlike lab settings, service robots often face competing instructions from multiple people (e.g., family members, nurses). Foundation models excel at multi-turn reasoning but lack arbitration mechanisms [270]. Research should explore multi-user dialogue management and conflict-resolution strategies for embodied agents.

The same properties that make foundation models powerful for perception, language-to-action, uncertainty handling, and efficient execution also necessitate explicit governance and HRI design, so that the benefits of Section 3 are realized without incurring the harms outlined here.

## 6. Future Research Directions

We have identified the significant research progress and of the open challenges of foundation models when applied to mobile service robots. We outline three forward-looking research directions that are directly motivated by the unresolved gaps identified across Sections 2-5: 1) reliability and lifelong adaptation, 2) privacy-aware and further resource constrained deployment, and 3) governance, standards, and human-in-the-loop frameworks. Figure 8 presents a forward-looking deployment roadmap that aligns our technical recommendations with practical adoption stages: *(i)* short-term integration of pretrained models into existing service settings, *(ii)* mid-term domain-specific fine-tuning to tackle unique clinical and domestic constraints, and *(iii)* long-term deployment of transparent, trustworthy, and human-aligned service robots. Each of the following research directions map onto one or more stages of this roadmap, helping transition foundation model-enabled robots from experimental prototypes into reliable assistants for everyday environments.

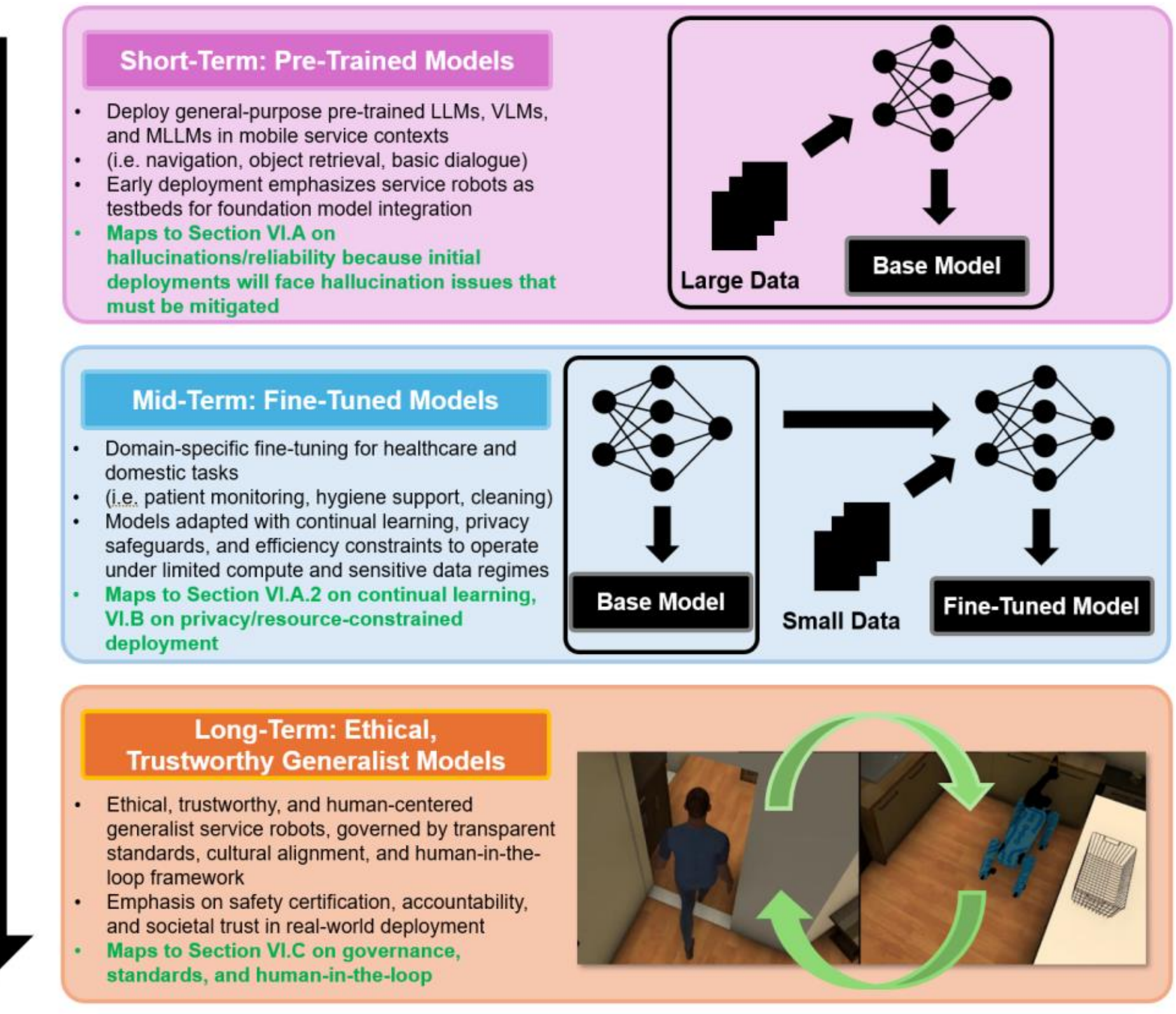

**Figure 8.** Roadmap shows short-, mid-, and long-term deployment of foundation-model service robots. Image of a robot and image of a human obtained from the Habitat Simulator 3.0 [271].

## 6.1. Reliability and Lifelong Adaptation

Reliability and lifelong adaptation address the need for mobile service robots with embedded foundation models to operate safely over time, avoiding hallucinations and preserving knowledge as environments and tasks evolve.

*1) Mitigating Hallucinations in Safety-Critical Domains:* LLMs and MLLMs are prone to generating plausible but incorrect outputs. In mobile service robots, hallucinations can cause serious risks [272], such as falsely identifying medical equipment or proposing unsafe navigation paths. Future research must develop hallucination-detection pipelines that ground foundation model outputs against real-time sensory consistency checks, ensuring that every proposed action is validated before execution [273].

*2) Addressing Catastrophic Forgetting During Continual Learning:* Mobile service robots deployed in hospitals or public spaces must adapt to evolving environments (e.g., service space rearrangements, new medical routines). Yet, online fine-tuning of foundation models risks catastrophic forgetting of previously learned tasks. Future work should explore continual adaptation methods [274], [275] with domain-specific rehearsal buffers, elastic weight consolidation at the edge, or modular adapters that preserve core reasoning while integrating local updates.

## 6.2. Privacy-Aware and Further Resource-Constrained Deployment

Privacy-aware and resource-constrained deployment highlights the challenge of adapting foundation models to sensitive domains like homes and hospitals, where strict data governance and limited onboard compute demand new approaches for secure, efficient, and context-specific execution.

*1) Data Scarcity and Privacy in Sensitive Domains:* Unlike web-scale training, robots in homes and hospitals face data scarcity and strict privacy constraints. Foundation models cannot rely on large-scale collection of in-situ data [276]. Future directions include federated foundation model adaptation, synthetic scene generation grounded in domestic and clinical layouts, and on-device fine-tuning under limited compute and memory budgets.

*2) Efficient Scaling for Embedded Execution:* Although Section 3.4 showed advances in compressed and efficient architectures, real deployments still face energy, latency, and throughput ceilings [277]. Further research is required on task-aware scaling laws, where model complexity dynamically better adapts to robot workload (e.g., low-power monitoring vs. high-bandwidth multi-person dialogue), ensuring safe operation without exhausting onboard resources. There is always room in optimizing computation load.

### *6.3. Governance, Standards, and Human-in-the-Loop Frameworks*

Governance, standards, and human-in-the-loop frameworks emphasize the need for systematic evaluation environments, regulatory guidance, and supervisory protocols to ensure that foundation model-enabled service robots operate safely, transparently, and under accountable oversight in human-centered domains.

*1) Standardized Benchmarks for Mobile Service Robots:* Existing foundation model benchmarks such as Habitat provide strong environments for testing navigation and embodied perception, but they remain limited in capturing the complexity of human-centered service tasks [31]. Current frameworks often evaluate vision, language, or manipulation in isolation, without integrating multimodal grounding, safety norms, or real-time human interaction. Future research should focus on developing multi-room, human-in-the-loop benchmarks that embed realistic human agents, unpredictable activities, and task-level evaluation metrics relevant to homes, hospitals, and service venues. Such benchmarks would enable systematic testing of foundation model-powered service robots under conditions that better approximate real-world domestic assistance, patient care, and service automation.

*2) Human-over-the-Loop Interaction Protocols:* Even with strong autonomy, mobile service robots must allow human supervisors to override or confirm foundation model-generated plans in safety-critical domains [278], [279] . Research should explore interaction protocols for seamless fallback control, combining foundation model reasoning with predictable human-in-the-loop interfaces.

## 7. Conclusions

This systematic review examined the integration of foundation models into mobile service robotics, focusing on how they address four core challenges: multimodal perception, translation of natural language into executable actions, uncertainty estimation across perception and planning, and computational efficiency for real-time operation. We analyzed how recent advances in LLMs, VLMs, MLLMs, and VLAs provide technical solutions to these challenges, such as cross-modal fusion, instruction grounding, confidence-weighted reasoning, and lightweight scalable architectures. We also discussed real-world applications in domestic assistance, healthcare, and service automation, highlighting how foundation models enable context-aware, socially responsive, and generalizable robot behaviors for service robots. While significant progress has been made in bridging semantic reasoning and embodied robot actions in real-world environments, critical limitations remain for achieving reliable and scalable deployment, specifically in model reliability and lifelong adaptation, privacy-aware and resource-constrained deployment, and the development of regulatory guidance and supervisory protocols for accountable operation in human-centered domains. Research in these areas will allow mobile service robots to safely, adaptively, and collaboratively interact in dynamic human environments to become proactive assistants in everyday life.

**Funding:** This work was supported in part by the Natural Sciences and Engineering Research Council of Canada (NSERC), NSERC Alliance and the NSERC CREATE ADVENTOR fellowship.

**Conflict of Interest:** The authors declare no conflicts of interest.

**References**

1. Radford, A.; Narasimhan, K.; Salimans, T.; Sutskever, I. Improving Language Understanding by Generative Pre-Training; Technical Report; OpenAI: San Francisco, CA, USA, 2018. Available Online: https://cdn.openai.com/research-covers/language-unsupervised/language_understanding_paper.pdf.
2. Brown, T.B.; Mann, B.; Ryder, N.; Subbiah, M.; Kaplan, J.D.; Dhariwal, P.; Neelakantan, A.; Shyam, P.; Sastry, G.; Askell, A.; et al. Language Models Are Few-Shot Learners. In Proceedings of the 34th International Conference on Neural Information Processing Systems (NeurIPS 2020), Vancouver, BC, Canada, 6–12 December 2020; Curran Associates, Inc.: Red Hook, NY, USA, 2020; Article No. 159, pp. 1877–1901. https://dl.acm.org/doi/10.5555/3495724.3495883.
3. Thirunavukarasu, A.J.; Ting, D.S.J.; Elangovan, K.; Gutierrez, L.; Tan, T.F.; Ting, D.S.W. Large Language Models in Medicine. Nat. Med. 2023, 29, 1930–1940. https://doi.org/10.1038/s41591-023-02448-8.
4. Touvron, H.; Martin, L.; Stone, K.; Albert, P.; Almahairi, A.; Babaei, Y.; Bashlykov, N.; Batra, S.; Bhargava, P.; Bhosale, S.; et al. LLaMA 2: Open Foundation and Fine-Tuned Chat Models. arXiv 2023, arXiv:2307.09288. https://doi.org/10.48550/arXiv.2307.09288.
5. Taori, R.; Gulrajani, I.; Zhang, T.; Dubois, Y.; Li, X.; Guestrin, C.; Liang, P.; Hashimoto, T.B. Alpaca: A Strong, Replicable Instruction-Following Model. Stanford Center for Research on Foundation Models, 2023. Available Online: https://crfm.stanford.edu/2023/03/13/alpaca.html.
6. Chowdhery, A.; Narang, S.; Devlin, J.; Bosma, M.; Mishra, G.; Roberts, A.; Barham, P.; Chung, H.W.; Sutton, C.; Gehrmann, S.; et al. PaLM: Scaling Language Modeling with Pathways. J. Mach. Learn. Res. 2023, 24, 240, 11324–11436. https://dl.acm.org/doi/10.5555/3648699.3648939.
7. Zhang, J.; Huang, J.; Jin, S.; Lu, S. Vision-Language Models for Vision Tasks: A Survey. IEEE Trans. Pattern Anal. Mach. Intell. 2024, 46, 5625–5644. https://doi.org/10.1109/TPAMI.2024.3369699.
8. Anderson, P.; He, X.; Buehler, C.; Teney, D.; Johnson, M.; Gould, S.; Zhang, L. Bottom-Up and Top-Down Attention for Image Captioning and Visual Question Answering. In Proceedings of the 2018 IEEE/CVF Conference on Computer Vision and Pattern Recognition (CVPR 2018), Salt Lake City, UT, USA, 18–22 June 2018; IEEE: 2018; pp. 6077–6086. https://doi.org/10.1109/CVPR.2018.00636.
9. Li, L.H.; Yatskar, M.; Yin, D.; Hsieh, C.-J.; Chang, K.-W. VisualBERT: A Simple and Performant Baseline for Vision and Language. arXiv 2019, arXiv:1908.03557. https://doi.org/10.48550/arXiv.1908.03557.
10. Ramesh, A.; Dhariwal, P.; Nichol, A.; Chu, C.; Chen, M. Hierarchical Text-Conditional Image Generation with CLIP Latents. arXiv 2022, arXiv:2204.06125. https://doi.org/10.48550/arXiv.2204.06125.
11. Bao, H.; Dong, L.; Piao, S.; Wei, F. BEiT: BERT Pre-Training of Image Transformers. arXiv 2021, arXiv:2106.08254. https://doi.org/10.48550/arXiv.2106.08254.
12. Ichter, B.; Brohan, A.; Chebotar, Y.; Finn, C.; Hausman, K.; Herzog, A.; Ho, D.; Ibarz, J.; Irpan, A.; Jang, E.; et al. Do as I Can, Not as I Say: Grounding Language in Robotic Affordances. In Proceedings of The 6th Conference on Robot Learning (CoRL 2022), Auckland, New Zealand, 14–18 December 2022; PMLR: 2023; pp. 287–318. https://proceedings.mlr.press/v205/ichter23a.html.
13. Driess, D.; Xia, F.; Sajjadi, M.S.M.; Lynch, C.; Chowdhery, A.; Ichter, B.; Wahid, A.; Tompson, J.; Vuong, Q.; Yu, T.; et al. PaLM-E: An Embodied Multimodal Language Model. In Proceedings of the 40th International Conference on Machine Learning (ICML 2023), Honolulu, HI, USA, 23–29 July 2023; JMLR.org: Red Hook, NY, USA, 2023; Article No. 340, 20 pp. https://dl.acm.org/doi/10.5555/3618408.3618748.
14. Zitkovich, B.; Yu, T.; Xu, S.; Xu, P.; Xiao, T.; Xia, F.; Wu, J.; Wohlhart, P.; Welker, S.; Wahid, A.; et al. RT-2: Vision-Language-Action Models Transfer Web Knowledge to Robotic Control. In Proceedings of the 7th Conference on Robot Learning (CoRL 2023), Atlanta, GA, USA, 6–9 November 2023; PMLR: 2023; Volume 229, pp. 2165–2183. https://proceedings.mlr.press/v229/zitkovich23a.html.

15. Rivkin, D.; Hogan, F.; Feriani, A.; Konar, A.; Sigal, A.; Liu, X.; Dudek, G. AIot Smart Home via Autonomous LLM Agents. IEEE Internet of Things Journal 2024. https://doi.org/10.1109/JIOT.2024.3471904.
16. Giudici, M.; Padalino, L.; Paolino, G.; Paratici, I.; Pascu, A.I.; Garzotto, F. Designing Home Automation Routines Using An LLM-Based Chatbot. Designs 2024, 8, 43. https://doi.org/10.3390/designs8030043.
17. Li, Y.; Wen, H.; Wang, W.; Li, X.; Yuan, Y.; Liu, G.; Liu, J.; Xu, W.; Wang, X.; Sun, Y.; et al. Personal LLM Agents: Insights and Survey About the Capability, Efficiency and Security. arXiv 2024, arXiv:2401.05459. https://doi.org/10.48550/arXiv.2401.05459.
18. Pandya, A. ChatGPT-Enabled daVinci Surgical Robot Prototype: Advancements and Limitations. Robotics 2023, 12, 97. https://doi.org/10.3390/robotics12040097.
19. Salichs, M.A.; Castro-González, Á.; Salichs, E.; Fernández-Rodicio, E.; Maroto-Gómez, M.; Gamboa-Montero, J.J.; Marques-Villarroya, S.; Castillo, J.C.; Alonso-Martín, F.; Malfaz, M. Mini: A New Social Robot for the Elderly. Int. J. Soc. Robot. 2020, 12, 1231–1249. https://doi.org/10.1007/s12369-020-00687-0.
20. Paiva, A.; Leite, I.; Boukricha, H.; Wachsmuth, I. Empathy in Virtual Agents and Robots: A Survey. ACM Trans. Interact. Intell. Syst. 2017, 7, 11. https://doi.org/10.1145/2912150.
21. Zhao, X.; Li, M.; Weber, C.; Hafez, M.B.; Wermter, S. Chat with the Environment: Interactive Multimodal Perception using Large Language Models. In Proceedings of the IEEE/RSJ International Conference on Intelligent Robots and Systems (IROS 2023), Detroit, MI, USA, 1–5 October 2023; IEEE: 2023; pp. 3590–3596. https://doi.org/10.1109/IROS55552.2023.10342363.
22. Xia, Y.; Zhang, J.; Jazdi, N.; Weyrich, M. Incorporating Large Language Models into Production Systems for Enhanced Task Automation and Flexibility. arXiv 2024, arXiv:2407.08550. https://doi.org/10.48550/arXiv.2407.08550.
23. Wang, Z.; Qin, H. Intelligent Industrial Production Process Automatic Regulation System Based on LLM Agents. In Proceedings of the 5th International Conference on Artificial Intelligence and Electromechanical Automation (AIEA 2024), 14–16 June 2024; IEEE: 2024; pp. 133–137. https://doi.org/10.1109/AIEA62095.2024.10692701.
24. Huang, W.; Abbeel, P.; Pathak, D.; Mordatch, I. Language Models as Zero-Shot Planners: Extracting Actionable Knowledge for Embodied Agents. In Proceedings of the 39th International Conference on Machine Learning (ICML 2022); PMLR: 2022; pp. 9118–9147. https://proceedings.mlr.press/v162/huang22a.html.
25. Huang, W.; Xia, F.; Xiao, T.; Chan, H.; Liang, J.; Florence, P.; Zeng, A.; Tompson, J.; Mordatch, I.; Chebotar, Y.; Sermanet, P.; Jackson, T.; Brown, N.; Luu, L.; Levine, S.; Hausman, K.; Ichter, B. Inner Monologue: Embodied Reasoning through Planning with Language Models. In Proceedings of the 6th Conference on Robot Learning (CoRL 2023), 14–18 December 2023; PMLR: Red Hook, NY, USA, 2023; Volume 205, pp. 1769–1782. https://proceedings.mlr.press/v205/huang23c.html.
26. Shridhar, M.; Manuelli, L.; Fox, D. CLIPort: What and Where Pathways for Robotic Manipulation. In Proceedings of the 5th Conference on Robot Learning (CoRL 2021), 8–11 November 2021; PMLR: Red Hook, NY, USA, 2022; Volume 164, pp. 894–906. https://proceedings.mlr.press/v164/shridhar22a.html.
27. Narcomey, A.; Tsoi, N.; Desai, R.; Vázquez, M. Learning human preferences over robot behavior as soft planning constraints. arXiv 2024, arXiv:2403.19795. https://doi.org/10.48550/arXiv.2403.19795.
28. Cao, Z.; Wang, Z.; Xie, S.; Liu, A.; Fan, L. Smart Help: Strategic Opponent Modeling for Proactive and Adaptive Robot Assistance in Households. In Proceedings of the IEEE/CVF Conference on Computer Vision and Pattern

Recognition (CVPR 2024), Seattle, WA, USA, 17–21 June 2024; IEEE: 2024; pp. 18091–18101. https://doi.org/10.1109/CVPR52733.2024.01713.

29. Xiao, A.; Janaka, N.; Hu, T.; Gupta, A.; Li, K.; Yu, C. Robi Butler: Multimodal Remote Interaction with a Household Robot Assistant. In Proceedings of the IEEE International Conference on Robotics and Automation (ICRA 2025), Atlanta, GA, USA, 19–23 May 2025; IEEE: Piscataway, NJ, USA, 2025. https://doi.org/10.1109/ICRA55743.2025.11128329.
30. Cao, Y.; Zhang, J.; Yu, Z.; Liu, S.; Qin, Z.; Zou, Q.; Du, B.; Xu, K. CogNav: Cognitive Process Modeling for Object Goal Navigation with LLMs. In Proceedings of the IEEE/CVF International Conference on Computer Vision (ICCV 2025), Paris, France, October 2025; IEEE/CVF: Piscataway, NJ, USA, 2025; pp. 9550–9560. https://openaccess.thecvf.com/content/ICCV2025/papers/Cao_CogNav_Cognitive_Process_Modeling_for_Object_Goal_Navigation_with_LLMs_ICCV_2025_paper.pdf.
31. Ji, Y.; Zhang, Z.; Tang, D.; et al. Foundation Models Assist in Human–Robot Collaboration Assembly. Sci. Rep. 2024, 14, 24828. https://doi.org/10.1038/s41598-024-75715-4.
32. Eftekhar, A.; Weihs, L.; Hendrix, R.; Caglar, E.; Salvador, J.; Herrasti, A.; Han, W.; VanderBil, E.; Kembhavi, A.; Farhadi, A.; et al. The One RING: A Robotic Indoor Navigation Generalist. arXiv 2024, arXiv:2412.14401. https://doi.org/10.48550/arXiv.2412.14401.
33. Hu, J.; Hendrix, R.; Farhadi, A.; Kembhavi, A.; Martín-Martín, R.; Stone, P. FLaRe: Achieving Masterful and Adaptive Robot Policies with Large-Scale Reinforcement Learning Fine-Tuning. In Proceedings of the IEEE International Conference on Robotics and Automation (ICRA 2025), Atlanta, GA, USA, 19–23 May 2025; IEEE: Piscataway, NJ, USA, 2025. https://doi.org/10.1109/ICRA55743.2025.11127934.
34. Fan, H.; Liu, X.; Fuh, J.Y.H.; Lu, W.F.; Li, B. Embodied Intelligence in Manufacturing: Leveraging Large Language Models for Autonomous Industrial Robotics. J. Intell. Manuf. 2025, 36, 1141–1157. https://doi.org/10.1007/s10845-023-02294-y.
35. Salierno, G.; Leonardi, L.; Cabri, G. Generative AI and large language models in Industry 5.0: Shaping Smarter Sustainable Cities. Encyclopedia 2025, 5, 30. https://doi.org/10.3390/encyclopedia5010030.
36. Li, S.; Wang, J.; Dai, R.; Ma, W.; Ng, W.Y.; Hu, Y. RoboNurse-VLA: Robotic Scrub Nurse System Based on Vision-Language-Action Model. In Proceedings of the IEEE/RSJ International Conference on Intelligent Robots and Systems (IROS 2025), Detroit, MI, USA, 19–25 October 2025; IEEE: Piscataway, NJ, USA, 2025. https://doi.org/10.1109/IROS60139.2025.11246030.
37. Wu, J.; Antonova, R.; Kan, A.; Lepert, M.; Zeng, A.; Song, S.; Bohg, J.; Rusinkiewicz, S.; Funkhouser, T. Tidybot: Personalized Robot Assistance with Large Language Models. Auton. Robots 2023, 47, 1087–1102. https://doi.org/10.1007/s10514-023-10139-z.
38. Zhao, Z.; Tang, H.; Yan, Y. Audio-Visual Navigation with Anti-Backtracking. In Pattern Recognition; Antonacopoulos, A.; Chaudhuri, S.; Chellappa, R.; Liu, C.-L.; Bhattacharya, S.; Pal, U., Eds.; Lecture Notes in Computer Science; Springer Nature Switzerland: Cham, Switzerland, 2025; Volume 15318, pp. 358–372. https://doi.org/10.1007/978-3-031-78456-9_23.
39. Fikes, R.E.; Nilsson, N.J. STRIPS: A New Approach to the Application of Theorem Proving to Problem Solving. Artif. Intell. 1971, 2, 189–208. https://doi.org/10.1016/0004-3702(71)90010-5.
40. Nau, D.S.; Au, T.-C.; Ilghami, O.; Kuter, U.; Murdock, J.W.; Wu, D.; Yaman, F. SHOP2: An HTN Planning System. J. Artif. Intell. Res. 2003, 20, 379–404. https://doi.org/10.1613/jair.1141.
41. Nau, D.; Au, T.-C.; Ilghami, O.; Kuter, U.; Wu, D.; Yaman, F.; Munoz-Avila, H.; Murdock, J.W. Applications of SHOP and SHOP2. IEEE Intell. Syst. 2005, 20, 34–41. https://doi.org/10.1109/MIS.2005.20.

42. Fox, M.; Long, D. PDDL2.1: An Extension to PDDL for Expressing Temporal Planning Domains. J. Artif. Intell. Res. 2003, 20, 61–124. https://dl.acm.org/doi/10.5555/1622452.1622454.
43. Ghallab, M.; Nau, D.; Traverso, P. Automated Planning: Theory and Practice; Morgan Kaufmann (Elsevier): San Francisco, CA, USA, 2004; ISBN 978-1-55860-856-6. https://doi.org/10.1016/B978-1-55860-856-6.X5000-5.
44. Misra, D.K.; Sung, J.; Lee, K.; Saxena, A. Tell me Dave: Context-Sensitive Grounding of Natural Language to Manipulation Instructions. Int. J. Robot. Res. 2016, 35, 281–300. https://doi.org/10.1177/0278364915602060.
45. MacGlashan, J.; Babes-Vroman, M.; desJardins, M.; Littman, M.; Muresan, S.; Squire, S.; Tellex, S.; Arumugam, D.; Yang, L. Grounding English commands to reward functions. In Proceedings of Robotics: Science and Systems XI (RSS 2015), Rome, Italy, 13–17 July 2015; Robotics: Science and Systems Foundation: 2015. https://doi.org/10.15607/RSS.2015.XI.018.
46. Chen, D.; Mooney, R. Learning to Interpret Natural Language Navigation Instructions from Observations. In Proceedings of the AAAI Conference on Artificial Intelligence (AAAI 2011), San Francisco, CA, USA, 7–11 August 2011; pp. 859–865. https://doi.org/10.1609/aaai.v25i1.7974.
47. Hazirbas, C.; Ma, L.; Domokos, C.; Cremers, D. FuseNet: Incorporating Depth into Semantic Segmentation Via Fusion-Based CNN Architecture. In Computer Vision – ACCV 2016; Lai, S.-H.; Lepetit, V.; Nishino, K.; Sato, Y., Eds.; Lecture Notes in Computer Science; Springer International Publishing: Cham, Switzerland, 2017; Volume 10111, pp. 213–228. https://doi.org/10.1007/978-3-319-54181-5_14.
48. Hong, D.; Yokoya, N.; Xia, G.-S.; Chanussot, J.; Zhu, X.X. X-ModalNet: A Semi-Supervised Deep Cross-Modal Network for Classification of Remote Sensing Data. ISPRS J. Photogramm. Remote Sens. 2020, 167, 12–23. https://doi.org/10.1016/j.isprsjprs.2020.06.014.
49. Zuñiga-Noël, D.; Ruiz-Sarmiento, J.-R.; Gomez-Ojeda, R.; Gonzalez-Jimenez, J. Automatic Multi-Sensor Extrinsic Calibration for Mobile Robots. IEEE Robot. Autom. Lett. 2019, 4, 2862–2869. https://doi.org/10.1109/LRA.2019.2922618.
50. Cobus, V.; Heuten, W. To Beep or Not to Beep? Evaluating Modalities for Multimodal ICU Alarms. Multimodal Technol. Interact. 2019, 3, 15. https://doi.org/10.3390/mti3010015.
51. Xia, B.; Zhou, J.; Kong, F.; You, Y.; Yang, J.; Lin, L. Enhancing 3D Object Detection Through Multi-Modal Fusion for Cooperative Perception. Alex. Eng. J. 2024, 104, 46–55. https://doi.org/10.1016/j.aej.2024.06.025.
52. Bohus, D.; Rudnicky, A.I. The Ravenclaw Dialog Management Framework: Architecture and systems. Comput. Speech Lang. 2009, 23, 332–361. https://doi.org/10.1016/j.csl.2008.10.001.
53. Colledanchise, M.; Ögren, P. How Behavior Trees Modularize Hybrid Control Systems and Generalize Sequential Behavior Compositions, The Subsumption Architecture, And Decision Trees. IEEE Trans. Robot. 2017, 33, 372–389. https://doi.org/10.1109/TRO.2016.2633567.
54. Young, S.; Gašić, M.; Thomson, B.; Williams, J.D. POMDP-Based Statistical Spoken Dialog Systems: A Review. Proc. IEEE 2013, 101, 1160–1179. https://doi.org/10.1109/JPROC.2012.2225812.
55. Gal, Y.; Ghahramani, Z. Dropout as a Bayesian Approximation: Representing Model Uncertainty in Deep Learning. In Proceedings of the 33rd International Conference on Machine Learning (ICML 2016), New York, NY, USA, 19–24 June 2016; PMLR: 2016; Vol. 48, pp. 1050–1059. https://dl.acm.org/doi/10.5555/3045390.3045502.
56. Beetz, M.; Tenorth, M.; Jain, D.; Bandouch, J. Towards Automated Models of Activities of Daily Life. Technol. Disabil. 2010, 22, 27–40. https://doi.org/10.3233/TAD-2010-0285.
57. Levine, S.; Finn, C.; Darrell, T.; Abbeel, P. End-to-End Training of Deep Visuomotor Policies. J. Mach. Learn. Res. 2016, 17, 1334–1373. https://dl.acm.org/doi/10.5555/2946645.2946684.

58. Duan, Y.; Chen, X.; Houthooft, R.; Schulman, J.; Abbeel, P. Benchmarking Deep Reinforcement Learning for Continuous Control. In Proceedings of the 33rd International Conference on Machine Learning (ICML 2016), New York, NY, USA, 20–22 June 2016; JMLR.org: 2016; pp. 1329–1338. https://dl.acm.org/doi/10.5555/3045390.3045531.
59. Kehoe, B.; Patil, S.; Abbeel, P.; Goldberg, K. A Survey Of Research on Cloud Robotics and Automation. IEEE Trans. Autom. Sci. Eng. 2015, 12, 398–409. https://doi.org/10.1109/TASE.2014.2376492.
60. Wang, X.; Tang, Z.; Guo, J.; Meng, T.; Wang, C.; Wang, T.; Jia, W. Empowering Edge Intelligence: A Comprehensive Survey on On-Device AI Models. ACM Comput. Surv. 2025. https://doi.org/10.1145/3724420.
61. Savva, M.; Kadian, A.; Maksymets, O.; Zhao, Y.; Wijmans, E.; Jain, B.; Straub, J.; Liu, J.; Koltun, V.; Malik, J.; et al. Habitat: A Platform for Embodied AI Research. In Proceedings of the IEEE/CVF International Conference on Computer Vision (ICCV 2019), Seoul, Korea, 27 October–2 November 2019; IEEE: 2019; pp. 9338–9346. https://doi.org/10.1109/ICCV.2019.00943.
62. Hu, Y.; Xie, Q.; Jain, V.; Francis, J.; Patrikar, J.; Keetha, N.; Kim, S.; Xie, Y.; Zhang, T.; Fang, H.-S.; et al. Toward General-Purpose Robots Via Foundation Models: A Survey and Meta-Analysis. arXiv 2024, arXiv:2312.08782. https://doi.org/10.48550/arXiv.2312.08782.
63. Zhou, H.; Yao, X.; Mees, O.; Meng, Y.; Xiao, T.; Bisk, Y.; Oh, J.; Johns, E.; Shridhar, M.; Shah, D.; et al. Bridging Language and Action: A Survey of Language-Conditioned Robot Manipulation. arXiv 2024, arXiv:2312.10807. https://doi.org/10.48550/arXiv.2312.10807.
64. Firoozi, R.; Tucker, J.; Tian, S.; Majumdar, A.; Sun, J.; Liu, W.; Zhu, Y.; Song, S.; Kapoor, A.; Hausman, K.; et al. Foundation Models in Robotics: Applications, Challenges, and the Future. Int. J. Robot. Res. 2024, 02783649241281508. https://doi.org/10.1177/02783649241281508.
65. Priem, J.; Piwowar, H.; Orr, R. OpenAlex: A Fully Open Index of Scholarly Works, Authors, Venues, Institutions, And Concepts. arXiv 2022, arXiv:2205.01833. https://doi.org/10.48550/arXiv.2205.01833.
66. Görner, M.; Haschke, R.; Ritter, H.; Zhang, J. MoveIt! Task Constructor for Task-Level Motion Planning. In Proceedings of the IEEE International Conference on Robotics and Automation (ICRA 2019), Montreal, QC, Canada, 20–24 May 2019; IEEE: 2019; pp. 190–196. https://doi.org/10.1109/ICRA.2019.8793898.
67. Tellex, S.; Kollar, T.; Dickerson, S.; Walter, M.R.; Banerjee, A.G.; Teller, S.; Roy, N. Approaching the Symbol Grounding Problem with Probabilistic Graphical Models. AI Mag. 2011, 32, 64–76. https://doi.org/10.1609/aimag.v32i4.2384.
68. Williams, T.; Cantrell, R.; Briggs, G.; Schermerhorn, P.; Scheutz, M. Grounding Natural Language References to Unvisited and Hypothetical Locations. In Proceedings of the AAAI Conference on Artificial Intelligence (AAAI), Bellevue, WA, USA, 14–18 July 2013; pp. 947–953. https://doi.org/10.1609/aaai.v27i1.8563.
69. Kollar, T.; Tellex, S.; Roy, D.; Roy, N. Toward Understanding Natural Language Directions. In Proceedings of the 5th ACM/IEEE International Conference on Human-Robot Interaction (HRI), Osaka, Japan, 2–5 March 2010; IEEE: Piscataway, NJ, USA, 2010; pp. 259–266. https://doi.org/10.1109/HRI.2010.5453186.
70. Walter, M.R.; Patki, S.; Daniele, A.F.; Fahnestock, E.; Duvallet, F.; Hemachandra, S.; Oh, J.; Stentz, A.; Roy, N.; Howard, T.M. Language Understanding For Field and Service Robots in A Priori Unknown Environments. Field Robot. 2022, 2, 1191–1231. https://doi.org/10.55417/fr.2022040.
71. Paul, R.; Arkin, J.; Roy, N.; Howard, T.M. Grounding Abstract Spatial Concepts for Language Interaction with Robots. In Proceedings of the Twenty-Sixth International Joint Conference on Artificial Intelligence (IJCAI 2017), Melbourne, Australia, 19–25 August 2017; IJCAI: 2017; pp. 4929–4933. https://doi.org/10.24963/ijcai.2017/696.

72. Kaelbling, L.P.; Lozano-Pérez, T. Hierarchical Task and Motion Planning in The Now. In Proceedings of the IEEE International Conference on Robotics and Automation (ICRA 2011), Shanghai, China, 9–13 May 2011; IEEE: 2011; pp. 1470–1477. https://doi.org/10.1109/ICRA.2011.5980391.
73. Mohr, F.; Lettmann, T.; Hüllermeier, E. Planning with Independent Task Networks. In KI 2017: Advances in Artificial Intelligence; Kern-Isberner, G.; Fürnkranz, J.; Thimm, M., Eds.; Lecture Notes in Computer Science; Springer International Publishing: Cham, Switzerland, 2017; Volume 10505, pp. 193–206. https://doi.org/10.1007/978-3-319-67190-1_15.
74. Hoffmann, J. FF: The Fast-Forward Planning System. AI Mag. 2001, 22, 57–62. https://doi.org/10.1609/aimag.v22i3.1572.
75. Helmert, M. The Fast Downward Planning System. J. Artif. Intell. Res. 2006, 26, 191–246. https://doi.org/10.1613/jair.1705.
76. Dornhege, C.; Eyerich, P.; Keller, T.; Trüg, S.; Brenner, M.; Nebel, B. Semantic Attachments for Domain-Independent Planning Systems. In Towards Service Robots for Everyday Environments; Prassler, E.; Zöllner, M.; Bischoff, R.; Burgard, W.; Haschke, R.; Hägele, M.; Lawitzky, G.; Nebel, B.; Plöger, P.; Reiser, U., Eds.; Springer: Berlin, Heidelberg, Germany, 2012; pp. 99–115. https://doi.org/10.1007/978-3-642-25116-0_9.
77. Mao, W.; Desai, R.; Iuzzolino, M.L.; Kamra, N. Action Dynamics Task Graphs for Learning Plannable Representations of Procedural Tasks. arXiv 2023, arXiv:2302.05330. https://doi.org/10.48550/arXiv.2302.05330.
78. Bacon, P.-L.; Harb, J.; Precup, D. The Option-Critic Architecture. In Proceedings of the Thirty-First AAAI Conference on Artificial Intelligence (AAAI 2017), San Francisco, CA, USA, 4–9 February 2017; AAAI Press: Palo Alto, CA, USA, 2017; pp. 1726–1734. https://dl.acm.org/doi/10.5555/3298483.3298491.
79. Kurniawati, H.; Du, Y.; Hsu, D.; Lee, W.S. Motion planning under uncertainty for robotic tasks with long time horizons. Int. J. Robot. Res. 2011, 30, 308–323. https://doi.org/10.1177/0278364910386986.
80. Lauri, M.; Hsu, D.; Pajarinen, J. Partially Observable Markov Decision Processes in Robotics: A Survey. IEEE Trans. Robot. 2023, 39, 21–40. https://doi.org/10.1109/TRO.2022.3200138.
81. Silver, D.; Veness, J. Monte-Carlo Planning in Large POMDPs. In Proceedings of the 24th International Conference on Neural Information Processing Systems (NIPS 2010), Vancouver, BC, Canada, 6–9 December 2010; Curran Associates, Inc.: Red Hook, NY, USA, 2010; pp. 2164–2172. https://dl.acm.org/doi/10.5555/2997046.2997137.
82. Ramachandram, D.; Taylor, G.W. Deep Multimodal Learning: A Survey on Recent Advances and Trends. IEEE Signal Process. Mag. 2017, 34, 96–108. https://doi.org/10.1109/MSP.2017.2738401.
83. Li, S.; Tang, H. Multimodal Alignment and Fusion: A Survey. arXiv 2024, arXiv:2411.17040. https://doi.org/10.48550/arXiv.2411.17040.
84. Wang, T.; Zheng, P.; Li, S.; Wang, L. Multimodal Human–Robot Interaction for Human-Centric Smart Manufacturing: A Survey. Adv. Intell. Syst. 2024, 6, 2300359. https://doi.org/10.1002/aisy.202300359.
85. Mora, A.; Prados, A.; Mendez, A.; Barber, R.; Garrido, S. Sensor Fusion for Social Navigation on a Mobile Robot Based on Fast Marching Square and Gaussian Mixture Model. Sensors 2022, 22, 8728. https://doi.org/10.3390/s22228728.
86. Zuo, X.; Geneva, P.; Lee, W.; Liu, Y.; Huang, G. LIC-Fusion: Lidar-Inertial-Camera Odometry. In Proceedings of the IEEE/RSJ International Conference on Intelligent Robots and Systems (IROS 2019), Macau, China, 3–8 November 2019; IEEE: 2019; pp. 5848–5854. https://doi.org/10.1109/IROS40897.2019.8967746.
87. Geneva, P.; Eckenhoff, K.; Lee, W.; Yang, Y.; Huang, G. OpenVINS: A research platform for visual-inertial estimation. In Proceedings of the IEEE International Conference on Robotics and Automation (ICRA 2020), Paris, France, 31 May–31 August 2020; IEEE: 2020; pp. 4666–4672. https://doi.org/10.1109/ICRA40945.2020.9196524.

88. Rehder, J.; Siegwart, R.; Furgale, P. A General Approach to Spatiotemporal Calibration in Multisensor Systems. IEEE Trans. Robot. 2016, 32, 383–398. https://doi.org/10.1109/TRO.2016.2529645.
89. Cadena, C.; Carlone, L.; Carrillo, H.; Latif, Y.; Scaramuzza, D.; Neira, J.; Reid, I.; Leonard, J.J. Past, Present, and Future of Simultaneous Localization and Mapping: Toward the Robust-Perception Age. IEEE Trans. Robot. 2016, 32, 1309–1332. https://doi.org/10.1109/TRO.2016.2624754.
90. Mur-Artal, R.; Tardós, J.D. ORB-SLAM2: An Open-Source SLAM System for Monocular, Stereo, and RGB-D Cameras. IEEE Trans. Robot. 2017, 33, 1255–1262, https://doi.org/10.1109/TRO.2017.2705103.
91. Engel, J.; Schöps, T.; Cremers, D. LSD-SLAM: Large-scale direct monocular SLAM. In Computer Vision – ECCV 2014; Fleet, D.; Pajdla, T.; Schiele, B.; Tuytelaars, T., Eds.; Springer International Publishing: Cham, Switzerland, 2014; pp. 834–849. https://doi.org/10.1007/978-3-319-10605-2_54.
92. Forster, C.; Carlone, L.; Dellaert, F.; Scaramuzza, D. On-Manifold Preintegration for Real-Time Visual–Inertial Odometry. IEEE Trans. Robot. 2017, 33, 1–21. https://doi.org/10.1109/TRO.2016.2597321.
93. Lupton, T.; Sukkarieh, S. Visual-Inertial-Aided Navigation for High-Dynamic Motion in Built Environments Without Initial Conditions. IEEE Trans. Robot. 2012, 28, 61–76. https://doi.org/10.1109/TRO.2011.2170332.
94. Gawlikowski, J.; Tassi, C.R.N.; Ali, M.; Lee, J.; Humt, M.; Feng, J.; Kruspe, A.; Triebel, R.; Jung, P.; Roscher, R.; et al. A Survey of Uncertainty in Deep Neural Networks. Artif. Intell. Rev. 2023, 56, 1513–1589. https://doi.org/10.1007/s10462-023-10562-9.
95. Jung, J.H.; Choe, Y.; Park, C.G. Photometric Visual-Inertial Navigation with Uncertainty-Aware Ensembles. IEEE Trans. Robot. 2022, 38, 2039–2052. https://doi.org/10.1109/TRO.2021.3139964.
96. Sünderhauf, N.; Brock, O.; Scheirer, W.; Hadsell, R.; Fox, D.; Leitner, J.; Upcroft, B.; Abbeel, P.; Burgard, W.; Milford, M.; et al. The Limits and Potentials of Deep Learning for Robotics. Int. J. Robot. Res. 2018, 37, 405–420. https://doi.org/10.1177/0278364918770733.
97. Maddern, W.; Pascoe, G.; Linegar, C.; Newman, P. 1 Year, 1000 Km: The Oxford RobotCar Dataset. Int. J. Robot. Res. 2017, 36, 3–15. https://doi.org/10.1177/0278364916679498.
98. Tremblay, J.; Prakash, A.; Acuna, D.; Brophy, M.; Jampani, V.; Anil, C.; To, T.; Cameracci, E.; Boochoon, S.; Birchfield, S. Training Deep Networks with Synthetic Data: Bridging the Reality Gap by Domain Randomization. In Proceedings of the IEEE/CVF Conference on Computer Vision and Pattern Recognition Workshops (CVPRW 2018), Salt Lake City, UT, USA, 18–22 June 2018; IEEE: 2018; pp. 1082–10828. https://doi.org/10.1109/CVPRW.2018.00143.
99. Sofman, B.; Lin, E.; Bagnell, J.A.; Cole, J.; Vandapel, N.; Stentz, A. Improving robot navigation through self-supervised online learning. J. Field Robot. 2006, 23, 1059–1075. https://doi.org/10.1002/rob.20169.
100. Porav, H.; Maddern, W.; Newman, P. Adversarial Training for Adverse Conditions: Robust Metric Localization Using Appearance Transfer. In Proceedings of the IEEE International Conference on Robotics and Automation (ICRA 2018), Brisbane, QLD, Australia, 21–25 May 2018; IEEE: 2018; pp. 1011–1018. https://doi.org/10.1109/ICRA.2018.8462894.
101. Lakshminarayanan, B.; Pritzel, A.; Blundell, C. Simple and Scalable Predictive Uncertainty Estimation Using Deep Ensembles. In Proceedings of the 31st International Conference on Neural Information Processing Systems (NeurIPS 2017), Long Beach, CA, USA, 4–9 December 2017; Curran Associates, Inc.: Red Hook, NY, USA, 2017; pp. 6405–6416. https://dl.acm.org/doi/10.5555/3295222.3295387.
102. Mohanan, M.G.; Salgoankar, A. A Survey of Robotic Motion Planning in Dynamic Environments. Robot. Auton. Syst. 2018, 100, 171–185. https://doi.org/10.1016/j.robot.2017.10.011.

103. Guo, C.; Pleiss, G.; Sun, Y.; Weinberger, K.Q. On Calibration of Modern Neural Networks. In Proceedings of the 34th International Conference on Machine Learning (ICML 2017), Sydney, NSW, Australia, 6–11 August 2017; PMLR: 2017; Vol. 70, pp. 1321–1330. https://dl.acm.org/doi/10.5555/3305381.3305518.
104. Simon, D. Kalman Filtering with State Constraints: A Survey of Linear and Nonlinear Algorithms. IET Control Theory Appl. 2010, 4, 1303–1318. https://doi.org/10.1049/iet-cta.2009.0032.
105. Khodarahmi, M.; Maihami, V. A Review on Kalman Filter Models. Arch. Comput. Methods Eng. 2023, 30, 727–747. https://doi.org/10.1007/s11831-022-09815-7.
106. Kurniawati, H.; Hsu, D.; Lee, W.S. SARSOP: Efficient Point-Based POMDP Planning by Approximating Optimally Reachable Belief Spaces. In Proceedings of Robotics: Science and Systems IV (RSS 2008), Zurich, Switzerland, June 2008. https://doi.org/10.15607/RSS.2008.IV.009.
107. Chen, S.Y. Kalman Filter for Robot Vision: A Survey. IEEE Trans. Ind. Electron. 2012, 59, 4409–4420. https://doi.org/10.1109/TIE.2011.2162714.
108. Wan, E.A.; Van der Merwe, R. The Unscented Kalman Filter for Nonlinear Estimation. In Proceedings of the IEEE Adaptive Systems for Signal Processing, Communications, and Control Symposium, Lake Louise, AB, Canada, 1–4 October 2000; IEEE: 2000; pp. 153–158. https://doi.org/10.1109/ASSPCC.2000.882463.
109. Van der Merwe, R.; Wan, E.A. The Square-Root Unscented Kalman Filter for State and Parameter Estimation. In Proceedings of the 2001 IEEE International Conference on Acoustics, Speech, and Signal Processing (ICASSP 2001), Salt Lake City, UT, USA, 7–11 May 2001; IEEE: 2001; Vol. 6, pp. 3461–3464. https://doi.org/10.1109/ICASSP.2001.940586.
110. Brouk, J.D.; DeMars, K.J. Kalman Filtering with Uncertain and Asynchronous Measurement Epochs. Navig. J. Inst. Navig. 2024, 71. https://doi.org/10.33012/navi.652.
111. Naab, C.; Zheng, Z. Application of the Unscented Kalman Filter in Position Estimation a Case Study on a Robot for Precise Positioning. Robot. Auton. Syst. 2022, 147, 103904. https://doi.org/10.1016/j.robot.2021.103904.
112. Platt, R.; Tedrake, R.; Kaelbling, L.P.; Lozano-Pérez, T. Belief space planning assuming maximum likelihood observations. Proceedings of Robotics: Science and Systems (RSS), Zaragoza, Spain, June 2010. https://doi.org/10.15607/RSS.2010.VI.037.
113. Leusmann, J.; Wang, C.; Gienger, M.; Schmidt, A.; Mayer, S. Understanding the Uncertainty Loop of Human–Robot Interaction. arXiv 2023, arXiv:2303.07889. https://doi.org/10.48550/arXiv.2303.07889.
114. Cumbal, R.; Lopes, J.; Engwall, O. Uncertainty in Robot Assisted Second Language Conversation Practice. In Companion of the 2020 ACM/IEEE International Conference on Human-Robot Interaction (HRI '20), Cambridge, UK, 23–26 March 2020; ACM: New York, NY, USA, 2020; pp. 171–173. https://doi.org/10.1145/3371382.3378306.
115. Hough, J.; Schlangen, D. It's Not What You Do, It's How You Do It: Grounding Uncertainty for a Simple Robot. In Proceedings of the 12th ACM/IEEE International Conference on Human-Robot Interaction (HRI 2017), Vienna, Austria, 6–9 March 2017; IEEE/ACM: 2017; pp. 274–282. https://doi.org/10.1145/2909824.3020214.
116. Trick, S.; Koert, D.; Peters, J.; Rothkopf, C.A. Multimodal Uncertainty Reduction for Intention Recognition in Human–Robot Interaction. In Proceedings of the 2019 IEEE/RSJ International Conference on Intelligent Robots and Systems (IROS), Macau, China, 3–8 November 2019; IEEE: 2019; pp. 7009–7016. https://doi.org/10.1109/IROS40897.2019.8968171.
117. Leidner, J.L.; Lieberman, M.D. Detecting Geographical References in the Form of Place Names and Associated Spatial Natural Language. SIGSPATIAL Spec. 2011, 3, 5–11. https://doi.org/10.1145/2047296.2047298.
118. Tellex, S.; Knepper, R.; Li, A.; Rus, D.; Roy, N. Asking for Help Using Inverse Semantics. In Proceedings of Robotics: Science and Systems (RSS 2014), Berkeley, CA, USA, July 2014. https://doi.org/10.15607/RSS.2014.X.024.

119. Dragan, A.D.; Lee, K.C.T.; Srinivasa, S.S. Legibility and Predictability of Robot Motion. In Proceedings of the 8th ACM/IEEE International Conference on Human–Robot Interaction (HRI 2013), Tokyo, Japan, 3–6 March 2013; pp. 301–308. https://doi.org/10.1109/HRI.2013.6483603.
120. He, K.; Zhang, X.; Ren, S.; Sun, J. Deep Residual Learning for Image Recognition. In Proceedings of the 2016 IEEE Conference on Computer Vision and Pattern Recognition (CVPR), Las Vegas, NV, USA, 27–30 June 2016; IEEE: 2016; pp. 770–778. https://doi.org/10.1109/CVPR.2016.90.
121. Qi, X.; Liao, R.; Jia, J.; Fidler, S.; Urtasun, R. 3D Graph Neural Networks For RGB-D Semantic Segmentation. In Proceedings of the 2017 IEEE International Conference on Computer Vision (ICCV), Venice, Italy, 22–29 October 2017; IEEE: 2017; pp. 5209–5218. https://doi.org/10.1109/ICCV.2017.556.
122. Zhou, T.; Porikli, F.; Crandall, D.J.; Van Gool, L.; Wang, W. A Survey on Deep Learning Techniques For Video Segmentation. IEEE Trans. Pattern Anal. Mach. Intell. 2022, 45, 7099–7122. https://doi.org/10.1109/TPAMI.2022.3225573.
123. Quinlan, S.; Khatib, O. Elastic bands: Connecting path planning and control. In Proceedings of the 1993 IEEE International Conference on Robotics and Automation (ICRA), Atlanta, GA, USA, 2–6 May 1993; IEEE: 1993; Vol. 2, pp. 802–807. https://doi.org/10.1109/ROBOT.1993.291936.
124. Fox, D.; Burgard, W.; Thrun, S. The Dynamic Window Approach to Collision Avoidance. IEEE Robot. Autom. Mag. 1997, 4, 23–33. https://doi.org/10.1109/100.580977.
125. Zhu, Y.; Gordon, D.; Kolve, E.; Fox, D.; Fei-Fei, L.; Gupta, A.; Mottaghi, R.; Farhadi, A. Visual Semantic Planning Using Deep Successor Representations. In Proceedings of the 2017 IEEE International Conference on Computer Vision (ICCV), Venice, Italy, 22–29 October 2017; IEEE: 2017; pp. 483–492. https://doi.org/10.1109/ICCV.2017.60.
126. Zhang, H.-B.; Zhang, Y.-X.; Zhong, B.; Lei, Q.; Yang, L.; Du, J.-X.; Chen, D.-S. A Comprehensive Survey of Vision-Based Human Action Recognition Methods. Sensors 2019, 19, 1005. https://doi.org/10.3390/s19051005.
127. Pütz, S.; Simón, J.S.; Hertzberg, J. Move Base Flex: A Highly Flexible Navigation Framework for Mobile Robots. In Proceedings of the 2018 IEEE/RSJ International Conference on Intelligent Robots and Systems (IROS), Madrid, Spain, 1–5 October 2018; IEEE: 2018; pp. 3416–3421. https://doi.org/10.1109/IROS.2018.8593829.
128. Arkin, R.C. Behavior-Based Robotics; MIT Press: Cambridge, MA, USA, 1998; ISBN 978-0-262-01165-5. Available Online: https://books.google.ca/books/about/Behavior_Based_Robotics.html?id=mRWT6alZt9oC&redir_esc=y.
129. Noroozi, F.; Daneshmand, M.; Fiorini, P. Conventional, Heuristic and Learning-Based Robot Motion Planning: Reviewing Frameworks of Current Practical Significance. Machines 2023, 11, 722. https://doi.org/10.3390/machines11070722.
130. Pandey, Dr.A. Mobile Robot Navigation and Obstacle Avoidance Techniques: A Review. Int. Robot. Autom. J. 2017, 2, 1–12. https://doi.org/10.15406/iratj.2017.02.00023.
131. Mohammed, M.Q.; Kwek, L.C.; Chua, S.C.; Al-Dhaqm, A.; Nahavandi, S.; Eisa, T.A.E.; Miskon, M.F.; Al-Mhiqani, M.N.; Ali, A.; Abaker, M. Review of Learning-Based Robotic Manipulation in Cluttered Environments. Sensors 2022, 22, 7938. https://doi.org/10.3390/s22207938.
132. Lewis, F.L.; Ge, S.S. Autonomous Mobile Robots: Sensing, Control, Decision Making and Applications; CRC Press: Boca Raton, FL, USA, 2018; ISBN 978-1-351-83711-8. https://doi.org/10.1201/9781315221229.
133. Guo, X.; Lyu, M.; Xia, B.; Zhang, K.; Zhang, L. An Improved Visual SLAM Method with Adaptive Feature Extraction. Appl. Sci. 2023, 13, 10038. https://doi.org/10.3390/app131810038.
134. Kurz, G.; Holoch, M.; Biber, P. Geometry-Based Graph Pruning for Lifelong SLAM. In Proceedings of the 2021 IEEE/RSJ International Conference on Intelligent Robots and Systems (IROS), Prague, Czech Republic, 27 September–1 October 2021; IEEE: 2021; pp. 3313–3320. https://doi.org/10.1109/IROS51168.2021.9636530.

135. Whelan, T.; Salas-Moreno, R.F.; Glocker, B.; Davison, A.J.; Leutenegger, S. ElasticFusion: Real-Time Dense SLAM and Light Source Estimation. Int. J. Rob. Res. 2016, 35, 1697–1716. https://doi.org/10.1177/0278364916669237.
136. Murali, A.; Liu, W.; Marino, K.; Chernova, S.; Gupta, A. Same Object, Different Grasps: Data and Semantic Knowledge for Task-Oriented Grasping. In Proceedings of the 2020 Conference on Robot Learning (CoRL), PMLR, Virtual Event, October 2021; pp. 1540–1557. https://proceedings.mlr.press/v155/murali21a.html.
137. Mayya, S.; Ramachandran, R.K.; Zhou, L.; Senthil, V.; Thakur, D.; Sukhatme, G.S.; Kumar, V. Adaptive and Risk-Aware Target Tracking for Robot Teams with Heterogeneous Sensors. IEEE Robot. Autom. Lett. 2022, 7, 5615–5622. https://doi.org/10.1109/LRA.2022.3155805.
138. Wang, J.; Lin, S.; Liu, A. Bioinspired Perception and Navigation of Service Robots in Indoor Environments: A Review. Biomimetics 2023, 8, 350, https://doi.org/10.3390/biomimetics8040350.
139. OpenAI; Hurst, A.; Lerer, A.; Goucher, A.P.; Perelman, A.; Ramesh, A.; Clark, A.; Ostrow, A.J.; Welihinda, A.; Hayes, A.; Radford, A.; et al. GPT-4o System Card. arXiv 2024, arXiv:2410.21276. https://doi.org/10.48550/arXiv.2410.21276.
140. Devlin, J.; Chang, M.-W.; Lee, K.; Toutanova, K. BERT: Pre-Training of Deep Bidirectional Transformers for Language Understanding. In Proceedings of the 2019 Conference of the North American Chapter of the Association for Computational Linguistics: Human Language Technologies (NAACL-HLT), Minneapolis, MN, USA, June 2019; Association for Computational Linguistics: Minneapolis, MN, USA, 2019; pp. 4171–4186. https://doi.org/10.18653/v1/N19-1423.
141. Radford, A.; Kim, J.W.; Hallacy, C.; Ramesh, A.; Goh, G.; Agarwal, S.; Sastry, G.; Askell, A.; Mishkin, P.; Clark, J.; et al. Learning Transferable Visual Models From Natural Language Supervision. In Proceedings of the Proceedings of the 38th International Conference on Machine Learning; PMLR, July 1 2021; pp. 8748–8763. https://proceedings.mlr.press/v139/radford21a.html.
142. Touvron, H.; Cord, M.; Jégou, H. DeiT III: Revenge of the ViT. In Computer Vision – ECCV 2022: 17th European Conference, Tel Aviv, Israel, 23–27 October 2022; Proceedings, Part XXIV; Springer: Berlin, Heidelberg, Germany, 2022; pp. 516–533. https://doi.org/10.1007/978-3-031-20053-3_30.
143. OpenAI; Achiam, J.; Adler, S.; Agarwal, S.; Ahmad, L.; Akkaya, I.; Aleman, F.L.; Almeida, D.; Altenschmidt, J.; Altman, S.; et al. GPT-4 Technical Report. arXiv 2023, arXiv:2303.08774. https://doi.org/10.48550/arXiv.2303.08774.
144. Guo, D.; Yang, D.; Zhang, H.; Song, J.; Zhang, R.; Xu, R.; Zhu, Q.; Ma, S.; Wang, P.; et al. DeepSeek-R1 Incentivizes Reasoning in LLMs through Reinforcement Learning. Nature 2025, 645, 633–638. https://doi.org/10.1038/s41586-025-09422-z.
145. Yang, J.; Tan, R.; Wu, Q.; Zheng, R.; Peng, B.; Liang, Y. Magma: A Foundation Model for Multimodal AI Agents. In Proceedings of the IEEE/CVF Conference on Computer Vision and Pattern Recognition (CVPR 2025), Nashville, TN, USA, 10–17 June 2025; IEEE: Piscataway, NJ, USA, 2025. https://doi.org/10.1109/CVPR52734.2025.01325.
146. Kim, M.J.; Pertsch, K.; Karamcheti, S.; Xiao, T.; Balakrishna, A.; Nair, S.; Rafailov, R.; Foster, E.P.; Sanketi, P.R.; Vuong, Q.; et al. OpenVLA: An Open-Source Vision-Language-Action Model. In Proceedings of the 8th Conference on Robot Learning (CoRL 2025), Atlanta, GA, USA, 6–9 November 2025; Proceedings of Machine Learning Research; PMLR: Cambridge, MA, USA, 2025; Volume 270, pp. 2679–2713. https://proceedings.mlr.press/v270/kim25c.html.
147. Zhu, H.; Wang, Y.; Zhou, J.; Chang, W.; Zhou, Y.; Li, Z.; Chen, J.; Shen, C.; Pang, J.; He, T. Aether: Geometric-Aware Unified World Modeling. In Proceedings of the IEEE/CVF International Conference on Computer Vision (ICCV 2025), Paris, France, October 2025; pp. 8535–8546. https://openaccess.thecvf.com/content/ICCV2025/papers/Zhu_Aether_Geometric-Aware_Unified_World_Modeling_ICCV_2025_paper.pdf.

148. Black, K.; Brown, N.; Driess, D.; Esmail, A.; Equi, M.R.; Finn, C.; Fusai, N.; Groom, L.; Hausman, K.; Ichter, B.; Jakubczak, S.; Jones, T.; Ke, L.; Levine, S.; Li-Bell, A.; Mothukuri, M.; Nair, S.; Pertsch, K.; Shi, L.X.; Smith, L.; Tanner, J.; Vuong, Q.; Walling, A.; Wang, H.; Zhilinsky, U. π0: A Vision-Language-Action Flow Model for General Robot Control. In Proceedings of Robotics: Science and Systems (RSS 2025), Los Angeles, CA, USA, June 2025. https://doi.org/10.15607/RSS.2025.XXI.010.
149. Black, K.; Brown, N.; Darpinian, J.; Dhabalia, K.; Driess, D.; Esmail, A.; Equi, M.R.; Finn, C.; Fusai, N.; Galliker, M.Y.; Ghosh, D.; Groom, L.; Hausman, K.; Ichter, B.; Jakubczak, S.; Jones, T.; Ke, L.; LeBlanc, D.; Levine, S.; Li-Bell, A.; Mothukuri, M.; Nair, S.; Pertsch, K.; Ren, A.Z.; Shi, L.X.; Smith, L.; Springenberg, J.T.; Stachowicz, K.; Tanner, J.; Vuong, Q.; Walke, H.; Walling, A.; Wang, H.; Yu, L.; Zhilinsky, U. π0.5: A Vision–Language–Action Model with Open-World Generalization. In Proceedings of the 9th Conference on Robot Learning (CoRL 2025), 27–30 September 2025; Proceedings of Machine Learning Research, Vol. 305; Lim, J.; Song, S.; Park, H.-W., Eds.; PMLR: 2025; pp. 17–40. https://proceedings.mlr.press/v305/black25a.html.
150. Szot, A.; Clegg, A.; Undersander, E.; Wijmans, E.; Zhao, Y.; Turner, J.; Maestre, N.; Mukadam, M.; Chaplot, D.S.; Maksymets, O.; Gokaslan, A.; Vondrus, V.; Dharur, S.; Meier, F.; Galuba, W.; Chang, A.; Kira, Z.; Koltun, V.; Malik, J.; Savva, M.; Batra, D. Habitat 2.0: Training Home Assistants to Rearrange Their Habitat. In Proceedings of the 35th International Conference on Neural Information Processing Systems (NeurIPS 2021); Curran Associates, Inc.: Red Hook, NY, USA, 2021; pp. 251–266. https://dl.acm.org/doi/10.5555/3540261.3540281.
151. Shu, D.; Zhao, H.; Hu, J.; Liu, W.; Payani, A.; Cheng, L.; Du, M. Large Vision–Language Model Alignment and Misalignment: A Survey through the Lens of Explainability. In Findings of the Association for Computational Linguistics: EMNLP 2025; Christodoulopoulos, C.; Chakraborty, T.; Rose, C.; Peng, V., Eds.; Association for Computational Linguistics: Suzhou, China, 2025; pp. 1713–1735. https://doi.org/10.18653/v1/2025.findings-emnlp.90.
152. Chen, Z.; Wang, W.; Tian, H.; et al. How Far Are We to GPT-4V? Closing the Gap to Commercial Multimodal Models with Open-Source Suites. Sci. China Inf. Sci. 2024, 67, 220101. https://doi.org/10.1007/s11432-024-4231-5.
153. NVIDIA; Agarwal, N.; Ali, A.; Bala, M.; Balaji, Y.; Barker, E.; Cai, T.; Chattopadhyay, P.; Chen, Y.; Cui, Y.; et al. Cosmos World Foundation Model Platform for Physical AI. arXiv 2025, arXiv:2501.03575. https://doi.org/10.48550/arXiv.2501.03575.
154. Ye, Y.; Huang, Z.; Xiao, Y.; Chern, E.; Xia, S.; Liu, P. LIMO: Less Is More for Reasoning. arXiv 2025, arXiv:2502.03387. https://doi.org/10.48550/arXiv.2502.03387.
155. Dong, X.; Zhang, P.; Zang, Y.; Cao, Y.; Wang, B.; Ouyang, L.; Zhang, S.; Duan, H.; Zhang, W.; Li, Y.; Yan, H.; Gao, Y.; Chen, Z.; Zhang, X.; Li, W.; Li, J.; Wang, W.; Chen, K.; He, C.; Zhang, X.; Dai, J.; Qiao, Y.; Lin, D.; Wang, J. InternLM-XComposer2-4KHD: A Pioneering Large Vision-Language Model Handling Resolutions from 336 Pixels to 4K HD. In *Proceedings of the 38th International Conference on Neural Information Processing Systems (NeurIPS 2024)*, Vancouver, BC, Canada, 10–15 December 2024; Curran Associates, Inc.: Red Hook, NY, USA, 2024; Article No. 1348. https://dl.acm.org/doi/10.5555/3737916.3739264.
156. Gu, Q.; Kuwajerwala, A.; Morin, S.; Jatavallabhula, K.M.; Sen, B.; Agarwal, A.; Rivera, C.; Paul, W.; Ellis, K.; Chellappa, R. Conceptgraphs: Open-Vocabulary 3D Scene Graphs for Perception and Planning. In Proceedings of the 2024 IEEE International Conference on Robotics and Automation (ICRA); IEEE: Yokohama, Japan, 2024; pp. 5021–5028. https://doi.org/10.1109/ICRA57147.2024.10610243.
157. Raffel, C.; Shazeer, N.; Roberts, A.; Lee, K.; Narang, S.; Matena, M.; Zhou, Y.; Li, W.; Liu, P.J. Exploring the Limits of Transfer Learning with a Unified Text-to-Text Transformer. J. Mach. Learn. Res. 2020, 21, 140, 1–67. https://dl.acm.org/doi/10.5555/3455716.3455856.

158. Yang, Z.; Li, L.; Lin, K.; Wang, J.; Lin, C.-C.; Liu, Z.; Wang, L. The dawn of LMMs: Preliminary Explorations with GPT-4V(ision). arXiv 2023, arXiv:2309.17421. https://doi.org/10.48550/arXiv.2309.17421.
159. Kirillov, A.; Mintun, E.; Ravi, N.; Mao, H.; Rolland, C.; Gustafson, L.; Xiao, T.; Whitehead, S.; Berg, A.C.; Lo, W.-Y.; et al. Segment Anything. In Proceedings of the 2023 IEEE/CVF International Conference on Computer Vision (ICCV); IEEE: Paris, France, October 2023; pp. 3992–4003. https://doi.org/10.1109/ICCV51070.2023.00371.
160. Défossez, A.; Mazaré, L.; Orsini, M.; Royer, A.; Pérez, P.; Jégou, H.; Grave, E.; Zeghidour, N. Moshi: A Speech-Text Foundation Model for Real-Time Dialogue. arXiv 2024, arXiv:2410.00037. https://doi.org/10.48550/arXiv.2410.00037.
161. Zuo, G.; Tong, J.; Liu, H.; Chen, W.; Li, J. Graph-Based Visual Manipulation Relationship Reasoning in Object-Stacking Scenes. In Proceedings of the International Joint Conference on Neural Networks (IJCNN); IEEE: 2021; pp. 1–8. https://doi.org/10.1109/IJCNN52387.2021.9534389.
162. Chen, X.; Djolonga, J.; Padlewski, P.; Mustafa, B.; Changpinyo, S.; Wu, J. On Scaling Up a Multilingual Vision and Language Model. In Proceedings of the IEEE/CVF Conference on Computer Vision and Pattern Recognition (CVPR 2024), Seattle, WA, USA, 16–22 June 2024. https://doi.org/10.1109/CVPR52733.2024.01368.
163. Liang, J.; Huang, W.; Xia, F.; Xu, P.; Hausman, K.; Ichter, B.; Florence, P.; Zeng, A. Code as Policies: Language Model Programs for Embodied Control. In Proceedings of the IEEE International Conference on Robotics and Automation (ICRA); IEEE: 2023; pp. 9493–9500. https://doi.org/10.1109/ICRA48891.2023.10160591.
164. Song, C.H.; Sadler, B.M.; Wu, J.; Chao, W.-L.; Washington, C.; Su, Y. LLM-Planner: Few-shot grounded planning for embodied agents with large language models. In Proceedings of the IEEE/CVF International Conference on Computer Vision (ICCV); IEEE: Paris, France, 2023; pp. 2986–2997. https://doi.org/10.1109/ICCV51070.2023.00280.
165. Muennighoff, N.; Yang, Z.; Shi, W.; Li, X.L.; Fei-Fei, L.; Hajishirzi, H.; Zettlemoyer, L.; Liang, P.; Candes, E.; Hashimoto, T. s1: Simple Test-Time Scaling. In Proceedings of the 2025 Conference on Empirical Methods in Natural Language Processing (EMNLP 2025), Suzhou, China, November 2025; pp. 20275–20321. https://doi.org/10.18653/v1/2025.emnlp-main.1025.
166. Kaelbling, L.P.; Lozano-Pérez, T. Hierarchical Task and Motion Planning in the Now. In Proceedings of the IEEE International Conference on Robotics and Automation (ICRA); IEEE: 2011; pp. 1470–1477. https://doi.org/10.1109/ICRA.2011.5980391.
167. Zhou, X.; Gandhi, S.; Fan, L.; Lin, Z.; Du, Y.; Abbeel, P.; Wu, J.; Xia, F. GENESIS: A Generative and Universal Physics Engine for Robotics and Beyond. Available Online: https://genesis-embodied-ai.github.io/.
168. Ye, J.; Chen, X.; Xu, N.; Zu, C.; Shao, Z.; Liu, S.; Cui, Y.; Zhou, Z.; Gong, C.; Shen, Y.; et al. A Comprehensive Capability Analysis of GPT-3 and GPT-3.5 Series Models. arXiv 2023, arXiv:2303.10420. https://doi.org/10.48550/arXiv.2303.10420.
169. Song, X.; Chen, W.; Liu, Y.; Chen, W.; Li, G.; Lin, L. Towards Long-Horizon Vision-Language Navigation: Platform, Benchmark and Method. In Proceedings of the IEEE/CVF Conference on Computer Vision and Pattern Recognition (CVPR 2025), Seattle, WA, USA, 10–17 June 2025. https://doi.org/10.1109/CVPR52734.2025.01128.
170. Abdin, M.; Jacobs, S.A.; Awan, A.A.; Aneja, J.; Awadallah, A.; Awadalla, H.; Bach, N.; Bahree, A.; Bakhtiari, A.; Behl, H.; et al. Phi-3 Technical Report: A Highly Capable Language Model Locally on Your Phone. arXiv 2024, arXiv:2404.14219. https://doi.org/10.48550/arXiv.2404.14219.
171. Ravi, N.; Gabeur, V.; Hu, Y.-T.; Hu, R.; Ryali, C.; Ma, T.; Khedr, H.; Rädle, R.; Rolland, C.; Gustafson, L.; Mintun, E.; Pan, J.; Alwala, K.V.; Carion, N.; Wu, C.-Y.; Girshick, R.; Dollár, P.; Feichtenhofer, C. SAM 2: Segment Anything in Images and Videos. In Proceedings of the International Conference on Learning Representations (ICLR 2025); Yue, Y.; Garg, A.; Peng, N.; Sha, F.; Yu, R., Eds.; 2025; pp. 28085–28128. https://proceedings.iclr.cc/paper_files/paper/2025/file/45c1f6a8cbf2da59ebf2c802b4f742cd-Paper-Conference.pdf.

172. Jaegle, A.; Borgeaud, S.; Alayrac, J.-B.; Doersch, C.; Ionescu, C.; Ding, D.; Koppula, S.; Zoran, D.; Brock, A.; Shelhamer, E.; Hénaff, O.; Botvinick, M.M.; Zisserman, A.; Vinyals, O.; Carreira, J. Perceiver IO: A General Architecture for Structured Inputs and Outputs. arXiv 2021, arXiv:2107.14795. https://doi.org/10.48550/arXiv.2107.14795.

173. Shridhar, M.; Manuelli, L.; Fox, D. Perceiver-Actor: A Multi-Task Transformer for Robotic Manipulation. In Proceedings of the 6th Conference on Robot Learning (CoRL 2023); PMLR: 2023; Vol. 205, pp. 785–799. https://proceedings.mlr.press/v205/shridhar23a.html.

174. Grattafiori, A.; Dubey, A.; Jauhri, A.; Pandey, A.; Kadian, A.; Al-Dahle, A.; Letman, A.; Mathur, A.; Schelten, A.; Vaughan, A.; Yang, A.; Fan, A.; Goyal, A.; Hartshorn, A.; Yang, A.; Mitra, A.; Sravankumar, A.; Korenev, A.; Hinsvark, A.; Rao, A.; Zhang, A.; et al. The LLaMA 3 herd of models. arXiv 2024, arXiv:2407.21783. https://doi.org/10.48550/arXiv.2407.21783.

175. Li, J.; Wu, J.; Zhao, W.; Bai, S.; Bai, X. PartGLEE: A Foundation Model for Recognizing and Parsing Any Objects. In Computer Vision – ECCV 2024; Leonardis, A., Ricci, E., Roth, S., Russakovsky, O., Sattler, T., Varol, G., Eds.; Lecture Notes in Computer Science, Vol. 15133; Springer: Cham, Switzerland, 2024; pp. 1–17. https://doi.org/10.1007/978-3-031-73226-3_27.

176. Diao, H.; Cui, Y.; Li, X.; Wang, Y.; Lu, H.; Wang, X. Unveiling Encoder-Free Vision-Language Models. In Proceedings of the 38th International Conference on Neural Information Processing Systems (NeurIPS 2024); Curran Associates, Inc.: Red Hook, NY, USA, 2024; Article 1665, 23p. https://dl.acm.org/doi/10.5555/3737916.3739581.

177. Chefer, H.; Singer, U.; Zohar, A.; Kirstain, Y.; Polyak, A.; Taigman, Y.; Wolf, L.; Sheynin, S. VideoJAM: Joint Appearance–Motion Representations for Enhanced Motion Generation in Video Models. In Proceedings of the 42nd International Conference on Machine Learning (ICML 2025); PMLR: 2025; Vol. 267, pp. 7595–7616. https://proceedings.mlr.press/v267/chefer25a.html.

178. Wang, Y.; Li, K.; Li, X.; Yu, J.; He, Y.; Wang, C.; Chen, G.; Pei, B.; Yan, Z.; Zheng, R.; et al. InternVideo2: Scaling Foundation Models for Multimodal Video Understanding. In Computer Vision – ECCV 2024; Leonardis, A.; Ricci, E.; Roth, S.; Russakovsky, O.; Sattler, T.; Varol, G., Eds.; Lecture Notes in Computer Science, Vol. 15143; Springer: Cham, Switzerland, 2024; https://doi.org/10.1007/978-3-031-73013-9_23.

179. Chu, X.; Qiao, L.; Zhang, X.; Xu, S.; Wei, F.; Yang, Y.; Sun, X.; Hu, Y.; Lin, X.; Zhang, B.; et al. MobileVLM V2: Faster And Stronger Baseline for Vision-Language Models. arXiv 2024, arXiv:2402.03766. https://doi.org/10.48550/arXiv.2402.03766.

180. Cho, M.; Cao, Y.; Sun, J.; Zhang, Q.; Pavone, M.; Park, J.J.; Yang, H.; Mao, Z.M. Cocoon: Robust Multi-Modal Perception with Uncertainty-Aware Sensor Fusion. arXiv 2024, arXiv:2410.12592. https://doi.org/10.48550/arXiv.2410.12592.

181. Ren, T.; Chen, Y.; Jiang, Q.; Zeng, Z.; Xiong, Y.; Liu, W.; Ma, Z.; Shen, J.; Gao, Y.; Jiang, X.; et al. DINO-X: A Unified Vision Model for Open-World Object Detection and Understanding. arXiv 2024, arXiv:2411.14347. https://doi.org/10.48550/arXiv.2411.14347.

182. Oquab, M.; Darcet, T.; Moutakanni, T.; Vo, H.; Szafraniec, M.; Khalidov, V.; Fernandez, P.; Haziza, D.; Massa, F.; El-Nouby, A.; Assran, M.; Ballas, N.; Galuba, W.; Howes, R.; Huang, P.-Y.; Li, S.-W.; Misra, I.; Rabbat, M.; Sharma, V.; Synnaeve, G.; Xu, H.; Jegou, H.; Mairal, J.; Labatut, P.; Joulin, A.; Bojanowski, P. DINOv2: Learning Robust Visual Features without Supervision. arXiv 2023, arXiv:2304.07193. https://doi.org/10.48550/arXiv.2304.07193.

183. Liu, H.; Li, C.; Li, Y.; Lee, Y.J. Improved Baselines with Visual Instruction Tuning. In Proceedings of the IEEE/CVF Conference on Computer Vision and Pattern Recognition (CVPR); IEEE: Seattle, WA, USA, 2024. https://doi.org/10.1109/CVPR52733.2024.02484.
184. Tong, S.; Brown, E.L., II; Wu, P.; Woo, S.; Iyer, A.J.; Akula, S.C.; Yang, S.; Yang, J.; Middepogu, M.; Wang, Z.; Pan, X.; Fergus, R.; LeCun, Y.; Xie, S. Cambrian-1: A Fully Open, Vision-Centric Exploration of Multimodal LLMs. In Proceedings of the 38th Conference on Neural Information Processing Systems (NeurIPS 2024), Vancouver, BC, Canada, 10–15 December 2024. https://dl.acm.org/doi/10.5555/3737916.3740687.
185. Li, H.; Zhu, J.; Jiang, X.; Zhu, X.; Li, H.; Yuan, C.; Wang, X.; Qiao, Y.; Wang, X.; Wang, W.; et al. Uni-Perceiver V2: A Generalist Model for Large-Scale Vision and Vision-Language Tasks. In Proceedings of the IEEE/CVF Conference on Computer Vision and Pattern Recognition (CVPR); IEEE: Vancouver, BC, Canada, 2023; pp. 2691–2700. https://doi.org/10.1109/CVPR52729.2023.00264.
186. Awadalla, A.; Gao, I.; Gardner, J.; Hessel, J.; Hanafy, Y.; Zhu, W.; Marathe, K.; Bitton, Y.; Gadre, S.; Sagawa, S.; et al. OpenFlamingo: An Open-Source Framework for Training Large Autoregressive Vision-Language Models. arXiv 2023, arXiv:2308.01390. https://doi.org/10.48550/arXiv.2308.01390.
187. Hafner, D.; Lillicrap, T.; Ba, J.; Norouzi, M. Dream to Control: Learning Behaviors by Latent Imagination. arXiv 2019, arXiv:1912.01603. https://doi.org/10.48550/arXiv.1912.01603.
188. Hafner, D.; Lillicrap, T.; Norouzi, M.; Ba, J. Mastering Atari with Discrete World Models. arXiv 2020, arXiv:2010.02193. https://doi.org/10.48550/arXiv.2010.02193.
189. Li, Q.; Lin, Y.; Luo, Q.; Yu, L. DreamerV3 for Traffic Signal Control: Hyperparameter Tuning and Performance. arXiv 2025, arXiv:2503.02279. https://doi.org/10.48550/arXiv.2503.02279.
190. Feichtenhofer, C.; Fan, H.; Li, Y.; He, K. Masked Autoencoders as Spatiotemporal Learners. In Proceedings of the 36th International Conference on Neural Information Processing Systems (NeurIPS 2022); Curran Associates, Inc.: Red Hook, NY, USA, 2022; pp. 2605–2617. https://dl.acm.org/doi/10.5555/3600270.3602875.
191. Liu, Z.; Lin, Y.; Cao, Y.; Hu, H.; Wei, Y.; Zhang, Z.; Lin, S.; Guo, B. Swin Transformer: Hierarchical Vision Transformer Using Shifted Windows. In Proceedings of the IEEE/CVF International Conference on Computer Vision (ICCV); IEEE: Montreal, QC, Canada, 2021; pp. 9992–10002. https://doi.org/10.1109/ICCV48922.2021.00986.
192. Bar-Tal, O.; Chefer, H.; Tov, O.; Herrmann, C.; Paiss, R.; Zada, S.; Ephrat, A.; Hur, J.; Liu, G.; Raj, A.; et al. Lumiere: A space-time diffusion model for video generation. In SIGGRAPH Asia 2024 Conference Papers; Association for Computing Machinery: New York, NY, USA, 2024; pp. 1–11. https://doi.org/10.1145/3680528.3687614.
193. Song, D.; Liang, J.; Payandeh, A.; Raj, A.H.; Xiao, X.; Manocha, D. VLM-Social-Nav: Socially Aware Robot Navigation Through Scoring Using Vision-Language Models. IEEE Robot. Autom. Lett. 2025, 10, 508–515. https://doi.org/10.1109/LRA.2024.3511409.
194. Narasimhan, S.; Tan, A.H.; Choi, D.; Nejat, G. OLiVia-Nav: An Online Lifelong Vision-Language Approach for Mobile Robot Social Navigation. In Proceedings of the 2025 IEEE International Conference on Robotics and Automation (ICRA 2025); IEEE: Atlanta, GA, USA, 19–23 May 2025. https://doi.org/10.1109/ICRA55743.2025.11128004.
195. Huang, Y.; Zhang, Q.; Y, P.S.; Sun, L. TrustGPT: A Benchmark for Trustworthy And Responsible Large Language Models. arXiv 2023, arXiv:2306.11507. https://doi.org/10.48550/arXiv.2306.11507.
196. Phan, L.; Gatti, A.; Han, Z.; Li, N.; Hu, J.; Zhang, H.; Zhang, C.B.C.; Shaaban, M.; Ling, J.; Shi, S.; et al. Humanity's Last Exam. arXiv 2025, arXiv:2501.14249. https://doi.org/10.48550/arXiv.2501.14249.

197. Tjomsland, J.; Kalkan, S.; Gunes, H. Mind Your Manners! A Dataset and a Continual Learning Approach for Assessing Social Appropriateness of Robot Actions. Front. Robot. AI 2022, 9. https://doi.org/10.3389/frobt.2022.669420.
198. Touvron, H.; Cord, M.; Douze, M.; Massa, F.; Sablayrolles, A.; Jégou, H. Training Data-Efficient Image Transformers and Distillation through Attention. In Proceedings of the 38th International Conference on Machine Learning (ICML 2021); PMLR: 2021; Vol. 139, pp. 10347–10357. https://proceedings.mlr.press/v139/touvron21a.html.
199. Chu, X.; Su, J.; Zhang, B.; Shen, C. VisionLLaMA: A Unified LLaMA Backbone for Vision Tasks. In Computer Vision – ECCV 2024; Leonardis, A.; Ricci, E.; Roth, S.; Russakovsky, O.; Sattler, T.; Varol, G., Eds.; Lecture Notes in Computer Science, Vol. 15148; Springer: Cham, Switzerland, 2024; pp. 1–18. https://doi.org/10.1007/978-3-031-72848-8_1.
200. MiniMax; Li, A.; Gong, B.; Yang, B.; Shan, B.; Liu, C.; Zhu, C.; Zhang, C.; Guo, C.; Chen, D.; et al. Minimax-01: Scaling Foundation Models with Lightning Attention. arXiv 2025, arXiv:2501.08313. https://doi.org/10.48550/arXiv.2501.08313.
201. Fedus, W.; Zoph, B.; Shazeer, N. Switch Transformers: Scaling to Trillion Parameter Models with Simple and Efficient Sparsity. J. Mach. Learn. Res. 2022, 23, 120. https://dl.acm.org/doi/abs/10.5555/3586589.3586709.
202. Lepikhin, D.; Lee, H.; Xu, Y.; Chen, D.; Firat, O.; Huang, Y.; Krikun, M.; Shazeer, N.; Chen, Z. GShard: Scaling Giant Models with Conditional Computation and Automatic Sharding. arXiv 2020, arXiv:2006.16668. https://doi.org/10.48550/arXiv.2006.16668.
203. Chen, W.; Huang, W.; Du, X.; Song, X.; Wang, Z.; Zhou, D. Auto-Scaling Vision Transformers without Training. arXiv 2022, arXiv:2202.11921. https://doi.org/10.48550/arXiv.2202.11921.
204. Qu, G.; Chen, Q.; Wei, W.; Lin, Z.; Chen, X.; Huang, K. Mobile Edge Intelligence for Large Language Models: A Contemporary Survey. IEEE Commun. Surv. Tutor. 2025, 1–1. https://doi.org/10.1109/COMST.2025.3527641.
205. Ge, T.; Chen, S.-Q.; Wei, F. EdgeFormer: A parameter-efficient transformer for on-device seq2seq generation. arXiv 2022, arXiv:2202.07959. https://doi.org/10.48550/arXiv.2202.07959.
206. MarketsandMarkets. Service Robotics Market Size, Share and Trends. 2024. Available Online: https://www.marketsandmarkets.com/Market-Reports/service-robotics-market-681.html.
208. Chiang, A.-H.; Trimi, S. Impacts of service robots on service quality. Serv. Bus. 2020, 14, 439–459. https://doi.org/10.1007/s11628-020-00423-8.
209. Liu, P.; Orru, Y.; Vakil, J.; Paxton, C.; Shafiullah, N.M.M.; Pinto, L. Demonstrating OK-Robot: What Really Matters in Integrating Open-Knowledge Models for Robotics. In Proceedings of Robotics: Science and Systems (RSS 2024); Delft, The Netherlands, July 2024. https://doi.org/10.15607/RSS.2024.XX.091.
210. Ghosh, D.; Walke, H.R.; Pertsch, K.; Black, K.; Mees, O.; Dasari, S.; Hejna, J.; Kreiman, T.; Xu, C.; Luo, J.; Tan, Y.L.; Chen, L.Y.; Vuong, Q.; Xiao, T.; Sanketi, P.R.; Sadigh, D.; Finn, C.; Levine, S. Octo: An Open-Source Generalist Robot Policy. In Proceedings of Robotics: Science and Systems (RSS 2024); Delft, The Netherlands, July 2024. https://doi.org/10.15607/RSS.2024.XX.090.
211. Chang, M.; Chhablani, G.; Clegg, A.; Dallaire Cote, M.; Desai, R.; Hlavac, M.; Karashchuk, V.; Krantz, J.; Mottaghi, R.; Parashar, P.; Patki, S.; Prasad, I.; Puig, X.; Rai, A.; Ramrakhya, R.; Tran, D.; Truong, J.; Turner, J.M.; Undersander, E.; Yang, T.-Y. PARTNR: A Benchmark for Planning and Reasoning in Embodied Multi-Agent Tasks. arXiv 2024, arXiv:2411.00081. https://doi.org/10.48550/arXiv.2411.00081.
212. Gu, Q.; Ju, Y.; Sun, S.; Gilitschenski, I.; Nishimura, H.; Itkina, M.; Shkurti, F. SAFE: Multitask Failure Detection for Vision-Language-Action Models. arXiv 2025, arXiv:2506.09937. https://doi.org/10.48550/arXiv.2506.09937.

213. Chen, B.; Xia, F.; Ichter, B.; Rao, K.; Gopalakrishnan, K.; Ryoo, M.S. Open-Vocabulary Queryable Scene Representations for Real World Planning. In Proceedings of the IEEE International Conference on Robotics and Automation (ICRA 2023); IEEE: 2023. https://doi.org/10.1109/ICRA48891.2023.10161534.
214. Shin, J.; Han, J.; Kim, S.; Oh, Y.; Kim, E. Task Planning for Long-Horizon Cooking Tasks Based On Large Language Models. In Proceedings of the IEEE/RSJ International Conference on Intelligent Robots and Systems (IROS); IEEE: Kyoto, Japan, 14–18 October 2024; pp. 13613–13619. https://doi.org/10.1109/IROS58592.2024.10801687.
215. Takebayashi, R.; Isume, V.H.; Kiyokawa, T.; Wan, W.; Harada, K. Cooking Task Planning Using LLM and Verified by Graph Network. In Proceedings of the 21st IEEE International Conference on Automation Science and Engineering (CASE 2025), 17–21 August 2025; IEEE: 2025. https://doi.org/10.1109/CASE58245.2025.11164158.
216. Singh, I.; Blukis, V.; Mousavian, A.; Goyal, A.; Xu, D.; Tremblay, J.; Fox, D.; Thomason, J.; Garg, A. ProgPrompt: Generating Situated Robot Task Plans Using Large Language Models. In Proceedings of the IEEE International Conference on Robotics and Automation (ICRA); IEEE: London, UK, May 2023; pp. 11523–11530. https://doi.org/10.1109/ICRA48891.2023.10161317.
217. Yang, J.; Chen, X.; Qian, S.; Madaan, N.; Iyengar, M.; Fouhey, D.F.; Chai, J. LLM-Grounder: Open-Vocabulary 3D Visual Grounding with Large Language Model as an Agent. In Proceedings of the IEEE International Conference on Robotics and Automation (ICRA); IEEE: Yokohama, Japan, 13–17 May 2024; pp. 7694–7701. https://doi.org/10.1109/ICRA57147.2024.10610443.
218. Luo, S.; Sun, P.; Zhu, J.; Deng, Y.; Yu, C.; Xiao, A. GSON: A Group-Based Social Navigation Framework with Large Multimodal Model. IEEE Robot. Autom. Lett. 2025, 10, 9646–9653. https://doi.org/10.1109/LRA.2025.3595038.
219. Ahn, M.; Dwibedi, D.; Finn, C.; Arenas, M.G.; Gopalakrishnan, K.; Hausman, K.; Ichter, B.; Irpan, A.; Joshi, N.; Julian, R.; et al. AutoRT: Embodied Foundation Models for Large-Scale Orchestration of Robotic Agents. arXiv 2024, arXiv:2401.12963. https://doi.org/10.48550/arXiv.2401.12963.
220. Rana, K.; Haviland, J.; Garg, S.; Abou-Chakra, J.; Reid, I.; Sünderhauf, N. SayPlan: Grounding Large Language Models using 3D Scene Graphs for Scalable Robot Task Planning. Proc. 7th Conf. Robot Learn. (CoRL 2023), PMLR 229, 2023, pp. 23–72. https://proceedings.mlr.press/v229/rana23a.html.
221. Chang, H.; Boyalakuntla, K.; Lu, S.; Cai, S.; Jing, E.P.; Keskar, S.; Geng, S.; Abbas, A.; Zhou, L.; Bekris, K.; Boularias, A. Context-Aware Entity Grounding with Open-Vocabulary 3D Scene Graphs. Proc. 7th Conf. Robot Learn. (CoRL 2023), PMLR 229, 2023, pp. 1950–1974. https://proceedings.mlr.press/v229/chang23b.html.
222. Shah, D.; Osiński, B.; Ichter, B.; Levine, S. LM-Nav: Robotic Navigation with Large Pre-Trained Models of Language, Vision, and Action. In Proceedings of the 6th Conference on Robot Learning (CoRL); PMLR: 2023; pp. 492–504. https://proceedings.mlr.press/v205/shah23b.html.
223. Shah, D.; Eysenbach, B.; Kahn, G.; Rhinehart, N.; Levine, S. ViNG: Learning Open-World Navigation with Visual Goals. In Proceedings of the IEEE International Conference on Robotics and Automation (ICRA); IEEE: Xi'an, China, May 2021; pp. 13215–13222. https://doi.org/10.1109/ICRA48506.2021.9561936.
224. Shi, L.X.; Ichter, B.; Equi, M.; Ke, L.; Pertsch, K.; Vuong, Q.; Tanner, J.; Walling, A.; Wang, H.; Fusai, N.; Li-Bell, A.; Driess, D.; Groom, L.; Levine, S.; Finn, C. Hi Robot: Open-Ended Instruction Following with Hierarchical Vision-Language-Action Models. arXiv 2025, arXiv:2502.19417. https://doi.org/10.48550/arXiv.2502.19417.
225. Gao, J.; Sarkar, B.; Xia, F.; Xiao, T.; Wu, J.; Ichter, B.; Majumdar, A.; Sadigh, D. Physically Grounded Vision-Language Models for Robotic Manipulation. In Proceedings of the IEEE International Conference on Robotics

and Automation (ICRA); IEEE: Yokohama, Japan, 2024; pp. 12462–12469. https://doi.org/10.1109/ICRA57147.2024.10610090.
226. Webster, C.; Ivanov, S. Robots, Artificial Intelligence and Service Automation in Tourism and Quality of Life. In Handbook of Tourism and Quality-of-Life Research II; Uysal, M.; Sirgy, M.J., Eds.; Springer: Cham, Switzerland, 2023; https://doi.org/10.1007/978-3-031-31513-8_36.
227. Choi, Y.; Choi, M.; Oh, M.; Kim, S. Service Robots in Hotels: Understanding the Service Quality Perceptions of Human–Robot Interaction. J. Hosp. Mark. Manag. 2020, 29, 613–635. https://doi.org/10.1080/19368623.2020.1703871.
228. Yuan, W.; Duan, J.; Blukis, V.; Pumacay, W.; Krishna, R.; Murali, A.; Mousavian, A.; Fox, D. RoboPoint: A Vision-Language Model for Spatial Affordance Prediction in Robotics. In Proceedings of the 8th Conference on Robot Learning (CoRL 2025); Agrawal, P.; Kroemer, O.; Burgard, W., Eds.; PMLR: 2025; Vol. 270, pp. 4005–4020; https://proceedings.mlr.press/v270/yuan25c.html.
229. Zheng, R.; Liang, Y.; Huang, S.; Gao, J.; Daumé III, H.; Kolobov, A.; Huang, F.; Yang, J. TraceVLA: Visual Trace Prompting Enhances Spatial-Temporal Awareness for Generalist Robotic Policies. arXiv 2024, arXiv:2412.10345. https://doi.org/10.48550/arXiv.2412.10345.
230. Jiang, Y.; Gupta, A.; Zhang, Z.; Wang, G.; Dou, Y.; Chen, Y.; Fei-Fei, L.; Anandkumar, A.; Zhu, Y.; Fan, L. VIMA: Robot Manipulation with Multimodal Prompts. In Proceedings of the 40th International Conference on Machine Learning (ICML 2023); 2023; https://dl.acm.org/doi/10.5555/3618408.3619019.
231. Min, S.Y.; Puig, X.; Chaplot, D.S.; Yang, T.-Y.; Rai, A.; Parashar, P.; Salakhutdinov, R.; Bisk, Y.; Mottaghi, R. Situated Instruction Following. In Computer Vision – ECCV 2024, Proceedings of the 18th European Conference on Computer Vision; Springer: Milan, Italy, 2024; pp. 202–228. https://doi.org/10.1007/978-3-031-73030-6_12.
232. Dalal, M.; Chiruvolu, T.; Chaplot, D.; Salakhutdinov, R. Plan-Seq-Learn: Language Model Guided RL for Solving Long-Horizon Robotics Tasks. In Proceedings of the International Conference on Learning Representations (ICLR 2024); 2024; https://proceedings.iclr.cc/paper_files/paper/2024/file/2e9f9cde1b709281a06dd14f679e4c51-Paper-Conference.pdf.
233. Zhao, T.Z.; Kumar, V.; Levine, S.; Finn, C. Learning Fine-Grained Bimanual Manipulation with Low-Cost Hardware. In Proceedings of Robotics: Science and Systems (RSS 2023), Daegu, Republic of Korea, July 2023. https://doi.org/10.15607/RSS.2023.XIX.016.
234. Fu, Z.; Zhao, T.Z.; Finn, C. Mobile ALOHA: Learning Bimanual Mobile Manipulation Using Low-Cost Whole-Body Teleoperation. In Proceedings of the 8th Conference on Robot Learning (CoRL 2025); PMLR: 2025; Vol. 270, pp. 4066–4083. https://proceedings.mlr.press/v270/fu25b.html.
235. Shafiullah, N.M.; Paxton, C.; Pinto, L.; Chintala, S.; Szlam, A. CLIP-Fields: Weakly Supervised Semantic Fields for Robotic Memory. In Proceedings of Robotics: Science and Systems (RSS 2023), Daegu, Republic of Korea, July 2023. https://doi.org/10.15607/RSS.2023.XIX.074.
236. Rozière, B.; Gehring, J.; Gloeckle, F.; Sootla, S.; Gat, I.; Tan, X.E.; Adi, Y.; Liu, J.; Sauvestre, R.; Remez, T.; et al. Code Llama: Open Foundation Models for Code. arXiv 2024, arXiv:2308.12950. https://doi.org/10.48550/arXiv.2308.12950.
237. Kim, J.; Mishra, A.K.; Limosani, R.; Scafuro, M.; Cauli, N.; Santos-Victor, J.; Mazzolai, B.; Cavallo, F. Control Strategies for Cleaning Robots in Domestic Applications: A Comprehensive Review. Int. J. Adv. Robot. Syst. 2019, 16, 1729881419857432. https://doi.org/10.1177/1729881419857432.

238. Touvron, H.; Lavril, T.; Izacard, G.; Martinet, X.; Lachaux, M.-A.; Lacroix, T.; Rozière, B.; Goyal, N.; Hambro, E.; Azhar, F.; et al. LLaMA: Open and Efficient Foundation Language Models. arXiv 2023, arXiv:2302.13971. https://doi.org/10.48550/arXiv.2302.13971.
239. Lym, H.J.; Son, H.I.; Kim, D.-Y.; Kim, J.; Kim, M.-G.; Chung, J.H. Child-Centered Home Service Design for a Family Robot Companion. Front. Robot. AI 2024, 11, 1346257. https://doi.org/10.3389/frobt.2024.1346257.
240. Sanh, V.; Debut, L.; Chaumond, J.; Wolf, T. DistilBERT: A distilled version of BERT—Smaller, faster, cheaper and lighter. arXiv 2020, arXiv:1910.01108. https://doi.org/10.48550/arXiv.1910.01108.
241. Honerkamp, D.; Büchner, M.; Despinoy, F.; Welschehold, T.; Valada, A. Language-Grounded Dynamic Scene Graphs for Interactive Object Search with Mobile Manipulation. IEEE Robot. Autom. Lett. 2024. https://doi.org/10.1109/LRA.2024.3441495.
242. Wang, Y.; Xian, Z.; Chen, F.; Wang, T.-H.; Wang, Y.; Fragkiadaki, K.; Erickson, Z.; Held, D.; Gan, C. RoboGen: Towards unleashing infinite data for automated robot learning via generative simulation. In Proceedings of the 41st International Conference on Machine Learning (ICML 2024), Vienna, Austria; JMLR.org: 2024; Article 2127. https://dl.acm.org/doi/10.5555/3692070.3694197.
243. Chen, M.; Tworek, J.; Jun, H.; Yuan, Q.; de Oliveira Pinto, H.P.; Kaplan, J.; Edwards, H.; Burda, Y.; Joseph, N.; Brockman, G.; et al. Evaluating Large Language Models Trained on Code. arXiv 2021, arXiv:2107.03374. https://doi.org/10.48550/arXiv.2107.03374.
244. Peng, S.; Genova, K.; Jiang, C.; Tagliasacchi, A.; Pollefeys, M.; Funkhouser, T. OpenScene: 3D scene understanding with open vocabularies. In Proceedings of the IEEE/CVF Conference on Computer Vision and Pattern Recognition (CVPR); IEEE: Vancouver, BC, Canada, 2023; pp. 815–824. https://doi.org/10.1109/CVPR52729.2023.00085.
245. Redmon, J.; Divvala, S.; Girshick, R.; Farhadi, A. You Only Look Once: Unified, Real-Time Object Detection. In Proceedings of the IEEE Conference on Computer Vision and Pattern Recognition (CVPR 2016), Las Vegas, NV, USA, 27–30 June 2016; IEEE: 2016. https://doi.org/10.1109/CVPR.2016.91.
246. Bai, S.; Liu, Y.; Han, Y.; Zhang, H.; Tang, Y.; Zhou, J. Self-Calibrated CLIP for Training-Free Open-Vocabulary Segmentation. IEEE Trans. Image Process. 2025, 34, 8271–8284. https://doi.org/10.1109/TIP.2025.3639996.
247. Shah, D.; Sridhar, A.; Dashora, N.; Stachowicz, K.; Black, K.; Hirose, N.; Levine, S. ViNT: A Foundation Model for Visual Navigation. In Proceedings of the 7th Conference on Robot Learning (CoRL 2023); PMLR: 2023; Vol. 229, pp. 711–733. https://proceedings.mlr.press/v229/shah23a.html.
248. Narasimhan, S.; Lisondra, M.; Wang, H.; Nejat, G. SplatSearch: Instance Image Goal Navigation For Mobile Robots Using 3D Gaussian Splatting and Diffusion Models. arXiv 2025, arXiv:2511.12972. https://doi.org/10.48550/arXiv.2511.12972.
249. Fung, A.; Tan, A.H.; Wang, H.; Benhabib, B.; Nejat, G. MLLM-Search: A Zero-Shot Approach to Finding People Using Multimodal Large Language Models. Robotics 2025, 14, 102. https://doi.org/10.3390/robotics14080102.
250. Tan, A.H.; Fung, A.; Wang, H.; Nejat, G. Mobile Robot Navigation Using Hand-Drawn Maps: A Vision Language Model Approach. IEEE Robot. Autom. Lett. 2025, 10, 7667–7674. https://doi.org/10.1109/LRA.2025.3577456.
251. Beyer, L.; Steiner, A.; Pinto, A.S.; Kolesnikov, A.; Wang, X.; Salz, D.; Neumann, M.; Alabdulmohsin, I.; Tschannen, M.; Bugliarello, E.; et al. Paligemma: A Versatile 3B Vision-Language Model for Transfer. arXiv 2024, arXiv:2407.07726. https://doi.org/10.48550/arXiv.2407.07726.
252. Dai, W.; Li, J.; Li, D.; Tiong, A.M.H.; Zhao, J.; Wang, W.; Li, B.; Fung, P.; Hoi, S. InstructBLIP: Towards General-Purpose Vision–Language Models with Instruction Tuning. In Proceedings of the 37th International Conference on Neural Information Processing Systems (NeurIPS 2023); Curran Associates, Inc.: Red Hook, NY, USA, 2023; Art. No. 2142. https://dl.acm.org/doi/10.5555/3666122.3668264.

253. Yao, Y.; Duan, J.; Xu, K.; Cai, Y.; Sun, Z.; Zhang, Y. A Survey on Large Language Model (LLM) Security and Privacy: The Good, The Bad, and The Ugly. High-Confid. Comput. 2024, 4, 100211. https://doi.org/10.1016/j.hcc.2024.100211.
254. Spreitzer, R.; Moonsamy, V.; Korak, T.; Mangard, S. Systematic Classification of Side-Channel Attacks: A Case Study for Mobile Devices. IEEE Commun. Surv. Tutor. 2018, 20, 465–488. https://doi.org/10.1109/COMST.2017.2779824.
255. Hettwer, B.; Gehrer, S.; Güneysu, T. Applications of Machine Learning Techniques in Side-Channel Attacks: A Survey. J. Cryptogr. Eng. 2020, 10, 135–162. https://doi.org/10.1007/s13389-019-00212-8.
256. Pa Pa, Y.M.; Tanizaki, S.; Kou, T.; van Eeten, M.; Yoshioka, K.; Matsumoto, T. An Attacker's Dream? Exploring the Capabilities of ChatGPT for Developing Malware. In Proceedings of the 16th Cyber Security Experimentation and Test Workshop (CSET 2023); Association for Computing Machinery: New York, NY, USA, 2023; pp. 10–18. https://doi.org/10.1145/3607505.3607513.
257. Alami, A.; Jensen, V.V.; Ernst, N.A. Accountability in Code Review: The Role of Intrinsic Drivers and the Impact of LLMs. ACM Trans Softw Eng Methodol 2025. https://doi.org/10.1145/3721127.
258. Raptis, E.K.; Kapoutsis, A.C.; Kosmatopoulos, E.B. Agentic LLM-Based Robotic Systems for Real-World Applications: A Review on Their Agenticness and Ethics. Front. Robot. AI 2025, 12. https://doi.org/10.3389/frobt.2025.1605405.
259. Sathish, V.; Lin, H.; Kamath, A.K.; Nyayachavadi, A. LLeMpower: Understanding Disparities in the Control and Access of Large Language Models. arXiv 2024, arXiv:2404.09356. https://doi.org/10.48550/arXiv.2404.09356.
260. Nagarajan, R.; Kondo, M.; Salas, F.; Sezgin, E.; Yao, Y.; Klotzman, V.; Godambe, S.A.; Khan, N.; Limon, A.; Stephenson, G.; et al. Economics and Equity of Large Language Models: Health Care Perspective. J. Med. Internet Res. 2024, 26, e64226. https://doi.org/10.2196/64226.
261. Frank, M.R.; Ahn, Y.-Y.; Moro, E. AI Exposure Predicts Unemployment Risk: A New Approach to Technology-Driven Job Loss. PNAS Nexus 2025, 4, pgaf107. https://doi.org/10.1093/pnasnexus/pgaf107.
262. Eloundou, T.; Manning, S.; Mishkin, P.; Rock, D. GPTs Are GPTs: Labor Market Impact Potential of LLMs. Science 2024, 384, 1306–1308. https://doi.org/10.1126/science.adj0998.
263. Adilazuarda, M.F.; Mukherjee, S.; Lavania, P.; Singh, S.S.; Aji, A.F.; O'Neill, J.; Modi, A.; Choudhury, M. Towards Measuring and Modeling "Culture" in LLMs: A Survey. In Proceedings of the 2024 Conference on Empirical Methods in Natural Language Processing (EMNLP 2024), Miami, FL, USA, November 2024; pp. 15763–15784. Association for Computational Linguistics. https://doi.org/10.18653/v1/2024.emnlp-main.882.
264. Li, C.; Chen, M.; Wang, J.; Sitaram, S.; Xie, X. CultureLLM: Incorporating Cultural Differences into Large Language Models. In Proceedings of the 38th International Conference on Neural Information Processing Systems (NeurIPS 2024), Vancouver, BC, Canada, December 2024; Curran Associates Inc.: Red Hook, NY, USA; Article No. 2693. https://dl.acm.org/doi/10.5555/3737916.3740609.
265. Pal, A.; Wangmo, T.; Bharadia, T.; Ahmed-Richards, M.; Bhanderi, M.B.; Kachhadiya, R.; Allemann, S.S.; Elger, B.S. Generative AI/LLMs for Plain Language Medical Information for Patients, Caregivers and General Public: Opportunities, Risks and Ethics. Patient Prefer. Adherence 2025, 19, 2227–2249. https://doi.org/10.2147/PPA.S527922.
266. Alessandro, G.; Dimitri, O.; Cristina, B.; Anna, M. The Emotional Impact of Generative AI: Negative Emotions and Perception of Threat. Behav. Inf. Technol. 2025, 44, 676–693. https://doi.org/10.1080/0144929X.2024.2333933.
267. Wester, J.; de Jong, S.; Pohl, H.; van Berkel, N. Exploring People's Perceptions of LLM-Generated Advice. Comput. Hum. Behav. Artif. Hum. 2024, 2, 100072. https://doi.org/10.1016/j.chbah.2024.100072.

268. Huang, J.-T.; Lam, M.H.; Li, E.J.; Ren, S.; Wang, W.; Jiao, W.; Tu, Z.; Lyu, M.R. Apathetic or Empathetic? Evaluating LLMss' Emotional Alignments with Humans. In Advances in Neural Information Processing Systems, NeurIPS 2024; Curran Associates Inc.: Red Hook, NY, USA, 2024. https://doi.org/10.52202/079017-3077.
269. Li, Y.; Huang, Y.; Wang, H.; Cheng, Y.; Zhang, X.; Zou, J.; Sun, L. Evaluating Large Language Models with Psychometrics. arXiv 2024, arXiv:2406.17675. https://doi.org/10.48550/arXiv.2406.17675.
270. Moreira, J.I.; Zhang, J. ChatGPT as a Fourth Arbitrator? The Ethics and Risks of Using Large Language Models in Arbitration. Arbitr. Int. 2025, 41, 71–84. https://doi.org/10.1093/arbint/aiae031.
271. Puig, X.; Undersander, E.; Szot, A.; Dallaire Cote, M.; Yang, T.-Y.; Partsey, R.; Desai, R.; Clegg, A.W.; Hlavac, M.; Min, S.Y.; Vondruš, V.; Gervet, T.; Berges, V.-P.; Turner, J.M.; Maksymets, O.; Kira, Z.; Kalakrishnan, M.; Malik, J.; Chaplot, D.S.; Jain, U.; Batra, D.; Rai, A.; Mottaghi, R. Habitat 3.0: A Co-Habitat for Humans, Avatars and Robots. arXiv 2023, arXiv:2310.13724. https://doi.org/10.48550/arXiv.2310.13724.
272. Ranasinghe, N.; Mohammed, W.M.; Stefanidis, K.; Martinez Lastra, J.L. Large Language Models in Human-Robot Collaboration with Cognitive Validation Against Context-Induced Hallucinations. IEEE Access 2025, 13, 77418–77430. https://doi.org/10.1109/ACCESS.2025.3565918.
273. Hamid, O.H. Beyond probabilities: Unveiling the Delicate Dance of Large Language Models (LLMs) and AI Hallucination. In Proceedings of the 2024 IEEE Conference on Cognitive and Computational Aspects of Situation Management (CogSIMA), 7–10 May 2024; IEEE: New York, NY, USA, 2024; pp. 85–90. https://doi.org/10.1109/CogSIMA61085.2024.10553755.
274. Dong, Q.; Zeng, P.; He, Y.; Wan, G.; Dong, X. Mitigating Catastrophic Forgetting in Robot Continual Learning: A Guided Policy Search Approach Enhanced with Memory-Aware Synapses. IEEE Robot. Autom. Lett. 2024, 9, 11242–11249. https://doi.org/10.1109/LRA.2024.3487484.
275. Taie, W.; ElGeneidy, K.; Al-Yacoub, A.; Sun, R. Addressing Catastrophic Forgetting in Payload Parameter Identification Using Incremental Ensemble Learning. Front. Robot. AI 2024, 11. https://doi.org/10.3389/frobt.2024.1470163.
276. O'Neill, A.; Rehman, A.; Maddukuri, A.; Gupta, A.; Padalkar, A.; Lee, A.; Pooley, A.; Gupta, A.; Mandlekar, A.; Jain, A.; et al. Open X-Embodiment: Robotic Learning Datasets And RT-X Models. In Proceedings of the 2024 IEEE International Conference on Robotics and Automation (ICRA), Yokohama, Japan, 13–17 May 2024; pp. 6892–6903. https://doi.org/10.1109/ICRA57147.2024.10611477.
277. Xu, M.; Cai, D.; Yin, W.; Wang, S.; Jin, X.; Liu, X. Resource-Efficient Algorithms and Systems of Foundation Models: A Survey. ACM Comput. Surv. 2025, 57, 1–39. https://doi.org/10.1145/3706418.
278. Merlo, E.; Lagomarsino, M.; Ajoudani, A. A Human-in-The-Loop Approach to Robot Action Replanning Through LLM Common-Sense Reasoning. IEEE Robot. Autom. Lett. 2025, 10, 10767–10774. https://doi.org/10.1109/LRA.2025.3604702.
279. Wang, H.; Tan, A.H.; Fung, A.; Nejat, G. X-Nav: Learning End-to-End Cross-Embodiment Navigation for Mobile Robots. IEEE Robot. Autom. Lett. 2026, 11, 698–705. https://doi.org/10.1109/LRA.2025.3632119.